\documentclass{article} 
\usepackage[preprint]{Template-2026/colm2026_conference}

\usepackage{microtype}
\usepackage{hyperref}
\usepackage{url}
\usepackage{booktabs}
\usepackage{amsmath}

\usepackage{graphicx}
\usepackage{multirow}
\usepackage[table]{xcolor}
\usepackage{arydshln}

\usepackage{placeins}

\usepackage{lineno}

\definecolor{darkblue}{rgb}{0, 0, 0.5}
\hypersetup{colorlinks=true, citecolor=darkblue, linkcolor=darkblue, urlcolor=darkblue}

\title{MF-QAT: Multi-Format Quantization-Aware Training for \\Elastic Inference}

\author{Zifei Xu\\
d-Matrix\\
Santa Clara, CA 95054\\
\texttt{xuzifei@d-matrix.ai} 
\And
Sayeh Sharify\\
d-Matrix\\
Santa Clara, CA 95054\\
\texttt{sayehs@d-matrix.ai} 
\And
Hesham Mostafa \\
d-Matrix\\
Santa Clara, CA 95054 \\
\texttt{hmostafa@d-matrix.ai}
}

\begin{document}

\ifcolmsubmission
\linenumbers
\fi

\maketitle

\begin{abstract}
  Quantization-aware training (QAT) is typically performed for a single target numeric format, while practical deployments often need to choose numerical precision at inference time based on hardware support or runtime constraints. We study \emph{multi-format QAT}, where a single model is trained to be robust across multiple quantization formats. We find that multi-format QAT can match single-format QAT at each target precision, yielding one model that performs well overall across different formats, even formats that were not seen during training. To enable practical deployment, we propose the \emph{Slice-and-Scale} conversion procedure for both MXINT and MXFP that converts a high-precision representation into lower-precision formats without re-training. Building on this, we introduce a pipeline that (i) trains a model with multi-format QAT, (ii) stores a single anchor format checkpoint (MXINT8/MXFP8), and (iii) allows on-the-fly conversion to lower MXINT or MXFP formats at runtime with negligible-or no-additional accuracy degradation. Together, these components provide a practical path to elastic precision scaling and allow selecting the runtime format at inference time across diverse deployment targets.

\end{abstract}

\section{Introduction}
Deep neural networks increasingly rely on quantization to meet the latency,
throughput, and memory constraints of modern deployment \citep{jacob_quantization_2017,esser_learned_2020,dettmers_llmint8_2022,xiao_smoothquant_2024,lin_awq_2024}. In many production
settings, however, the target numeric format is not fixed: different
accelerators support different formats, and the same device might want to serve at different precisions for different batches based on the current load of the system \citep{cai_once-for-all_2020,lee_nestedfp_2025,jin_adabits_2020,yi_one_2025,chen_otaro_2025}. Standard quantization-aware training (QAT) commonly optimizes a model for a \emph{single} format \citep{chen_efficientqat_2025,ma_era_2024}, which
creates a brittle dependency between training and deployment.

This paper asks a simple question: can we train one model that is robust across
multiple quantization formats, and deploy it in a way that supports
further precision reduction at inference time with minimal loss?

We answer both affirmatively. First, we discover that models trained with
multi-format QAT perform on par with single-format QAT at the target precisions, so a single model can serve multiple deployment formats effectively.

Second, we design Slice-and-Scale format transformation methods for both MXINT
and MXFP that produce lower-precision representations from higher-precision
ones. These transformations enable a practical workflow in which a single stored
checkpoint can be converted to different precisions efficiently without
re-expanding to FP32 model weights.

Third, we propose an inference scheme MF-QAT that combines these ideas: we train a model
with multi-format QAT, store the resulting model in a single anchor microscaling format (MXINT8/MXFP8), and allow it to be further quantized directly to an even lower precision
at inference time. Empirically, this final step incurs only minimal additional
accuracy degradation, enabling elastic precision scaling depending on
runtime constraints.

\section{Background and related work}
\label{sec:background}

\paragraph{Quantization and QAT.}
Post-training quantization and quantization-aware training (QAT) are standard
approaches to reduce memory footprint and improve throughput \citep{jacob_quantization_2017,esser_learned_2020,dettmers_llmint8_2022,xiao_smoothquant_2024,lin_awq_2024,xu_understanding_2025,saxena_resq_2025}. QAT simulates the
effects of quantization during training, typically via
straight-through estimators \cite{yin_understanding_2019}, allowing the model to adapt to the induced noise.
While effective, most QAT pipelines optimize for a \emph{single} target format
(e.g., a fixed bitwidth and quantization scheme) \citep{chen_efficientqat_2025,ma_era_2024}, which can lead to accuracy
drops when the deployment format differs.

\paragraph{Microscaling formats.}
Many modern accelerators support integer microscaling (MXINT) and floating-point microscaling (MXFP), which employ shared scaling factors over blocks of values to achieve efficient computation while retaining sufficient dynamic range. In this framework, a tensor is represented by low-precision elements together with a shared scale for each block. Accordingly, a microscaling format is defined by (i) the scale-factor data type, (ii) the element data type and precision, and (iii) the scaling block size \citep{rouhani_ocp_nodate}. From this perspective, MXINT and MXFP share the same block-wise scaling abstraction and differ only in whether the elements are encoded as integers or floating-point numbers. Due to their ability to reduce memory usage and inference latency with negligible accuracy loss, microscaling formats are becoming an increasingly important quantization primitive in modern hardware \citep{sharify_post_2024,rouhani_microscaling_2023,mxstudy,lee2025mxpushinglimitsmicroscaling}.

\paragraph{Elastic inference serving.}
Recent work on elastic inference goes beyond single-format quantization by explicitly enabling conversion from a higher-precision checkpoint to lower-precision variants, reducing or eliminating the need to store multiple full model copies. Representative examples include overlay-style and nested parameterizations that expose multiple precision levels from a largely shared memory footprint
\citep{park_any-precision_2024,lee_nestedfp_2025,chen_otaro_2025}. This prior work considered clustering-based quantization formats ~\citep{park_any-precision_2024} or FP16/FP8 formats~\cite{lee_nestedfp_2025}, while in the current work we mainly consider MXINT and MXFP formats. 
Other work focuses on training a single model that can tolerate dynamic precision selection under varying runtime constraints, such as latency or throughput requirements. These approaches include training with nested data formats, precision-specific batch normalization, token/layer-level precision switching, and lightweight LoRA adapters tailored to different precisions.
\citep{yi_one_2025,chen_otaro_2025,liu_flexquant_2025,jin_adabits_2020,bulat_bit-mixer_2021}. Unlike this prior work, we do not change the network architecture  through extra normalization layers or LoRA adapters. Instead, we directly quantize the weights to obtain a single high-precision checkpoint that can be efficiently converted to many different low-precision formats during deployment. Our multi-format QAT scheme leads to minimal or no accuracy degradation across the many possible deployment formats, even the formats that were not seen during QAT.  

\section{Methods}
\label{sec:methods}
\subsection{Models, tasks, and datasets}
\label{sec:methods_models_data}
We conduct experiments on the following pretrained LLMs:
\texttt{Llama-2-7b-hf}, \texttt{Llama-3.2-1B}, and \texttt{Llama-3.2-3B} from the Llama2~\citep{touvron_llama_2023} and the Llama-3.2~\citep{grattafiori2024llama3herdmodels} families and \texttt{Qwen3-0.6B-Base}, \texttt{Qwen3-1.7B-Base}, and \texttt{Qwen3-4B-Base}, from the recently released Qwen3 family~\citep{yang2025qwen3technicalreport}. We also include multimodal language models belonging to Qwen3-VL family: \texttt{Qwen3-VL-2B-Instruct} and \texttt{Qwen3-VL-4B-Instruct}~\citep{bai_qwen3-vl_2025} for our evaluations.

For quantization-aware finetuning, we use 128 examples from the WikiText-2 train split~\citep{merity_pointer_2016}, and we evaluate the multiformat QAT approach on a range of tasks which measure the \textit{language modeling ability}: perplexity on WikiText-2 validation split~\citep{merity_pointer_2016}, \textit{common sense reasoning ability}: normalized 0-shot accuracy on HellaSwag~\citep{zellers_hellaswag_2019}, \textit{language understanding}: 0-shot accuracy on MMLU~\citep{hendrycks_measuring_2021}, and \textit{mathematical understanding}: normalized 0-shot accuracy on MathQA~\citep{amini_mathqa_2019}. For multimodal vision language models, we use ChartQA \citep{masry_chartqa_2022} for \textit{visual question answering}. 

\subsection{Multi-format training scheme}
\label{sec:methods_multiformat_qat}
\paragraph{Weight-only quantization.}
We perform weight-only quantization in the text decoder stack (excluding \emph{lm\_head}). During QAT, only the quantized weight
parameters are updated, all other parameters are frozen. Similarly, for the full precision finetuning baseline, only these weight parameters are updated. 

\paragraph{Training formats and schedule.}
We train sequentially on MXINT and MXFP weight formats with bitwidths
$b \in \{2,4,6,8\}$ and $\{4 (E2M1),6(E3M2),8(E4M3)\}$ respectively. We train in \emph{increasing} bit order (2$\rightarrow$4$\rightarrow$6$\rightarrow$8),
because lower-precision weights typically require larger updates to jump out of quantization bin, training in the opposite direction can destabilize the higher-precision quantization settings learned earlier. Each bitwidth is trained for 1 epoch, resulting in 4 epochs total for MXINT multi-format QAT and 3 epochs total for MXFP multi-format QAT. Each epoch contains the same 128 data examples.

\paragraph{Baselines.}
For comparison, we (i) finetune the model in full precision, and
(ii) run single-format QAT for each target format. We use the same number of epochs as the multi-format QAT runs for fair comparison. For models larger than 2B, we observed severe overfitting for single-format QAT models, hence, the number of epochs are reduced to 1 for all runs for these models (each format in multi-format QAT is assigned equal number of steps within the epoch).

\paragraph{Evaluation.}
We evaluate on MXINT formats with bitwidths $b \in \{2,3,4,5,6,7,8\}$ and MXFP formats with bitwidths $b \in \{4 (E2M1),5 (E2M2),6(E3M2),7(E3M3),8(E4M3)\}$.
For each trained variant (full-precision finetune, single-format QAT, and
multi-format QAT), we apply post-training quantization (PTQ) to convert the resulting checkpoint to the target evaluation format before measuring
performance on WikiText-2 validation. This isolates the effect of the training procedure on robustness to format changes and ensures that all compared variants are evaluated in the same target format.

\paragraph{Optimization details.}
For each model and training variant, we sweep learning rates in
$\{10^{-4}, 10^{-5}, 10^{-6}\}$ for MXINT and $\{10^{-5}, 10^{-6},10^{-7}\}$ for MXFP, and report results for the best-performing
learning rate. We use \texttt{torch.optim.AdamW} with default hyperparameters.

\subsection{Slice-and-Scale MXINT (SSMXINT) format conversion}
\label{sec:ssmxint}
To enable fast conversion between higher-precision and lower-precision
microscaling integer (MXINT) formats, we propose \emph{Slice-and-Scale MXINT} (SSMXINT). SSMXINT converts a tensor represented in a higher-precision MXINT format (e.g., MXINT8) into a lower-precision MXINT format  by (i) \emph{slicing} the least-significant bits of the
elements with rounding and (ii) \emph{scaling} the shared block scale
so that the represented real range is preserved.

\paragraph{Notation (used in Sec.~\ref{sec:ssmxint}--\ref{sec:ssmxfp}).}
For a block of scalar floats $V=\{V_i\}_{i=1}^{k}$ with block size of $k$, MX conversion returns a shared scale $X$ and elements $\{P_i\}_{i=1}^{k}$. The reconstruction is $\hat{V}_i = X P_i$.
Let $e_{\max}(f)$ denote the exponent of the largest normal number in
element format $f$, and let $\operatorname{quantize}_{f}(\cdot)$ denote conversion to $f$ (rounding plus clamping).
We use $\operatorname{Round}(\cdot)$ for element-wise round-to-nearest.
Following the MX format conversion algorithm in \citep{rouhani_microscaling_2023},
{\small
\begin{equation}
\texttt{shared\_exp}
= \left\lfloor \log_2\!\left(\max_i |V_i|\right) \right\rfloor
- e_{\max}(f),
\qquad
X = 2^{\texttt{shared\_exp}},
\label{eq:scale}
\end{equation}
\begin{equation}
P_i = \operatorname{quantize}_{f}\!\left(\frac{V_i}{X}\right).
\label{eq:element}
\end{equation}
}

\paragraph{MXINT representation.}
For MXINT, the shared scale follows Eq.\ref{eq:scale} with
$f=\mathrm{MXINT}(b)$, where $b$ is the number of bits for the element format:
{\small
\begin{equation}
\texttt{shared\_exp}
= \left\lfloor \log_2\!\left(\max_i |V_i|\right) \right\rfloor
- e_{\max}(b),
\qquad
X = 2^{\texttt{shared\_exp}}.
\end{equation}
}
Elements are then computed by
$P_i = \operatorname{clip}_{b}(\operatorname{Round}(V_i/X))$\label{element}.

\paragraph{High-to-low conversion via slice-and-scale.}
Let $(X_h,P_h,b_h)$ be a high-precision MXINT block and
$(X_\ell,P_\ell,b_\ell)$ the target one, with $b_h>b_\ell$. Since $V$ is fixed, the computation of scale for different MXINT precisions only differs in the value of $e_{\max}(b)$. Hence, it is theoretically equivalent to obtain the low-precision MXINT scale from the high-precision MXINT scale accounting for the difference in $e_{\max}(b)$ versus obtaining the low-precision MXINT scale from the original full precision tensor.\\

Define
$\Delta e=e_{\max}(b_h)-e_{\max}(b_\ell)$.
For signed MXINT, this is equivalent to $\Delta e=b_h-b_\ell$.
SSMXINT converts by
{\small
\begin{equation}
P_{\ell,i}
= \operatorname{clip}_{b_\ell}\!\left(
\operatorname{Round}\!\left(\frac{P_{h,i}}{2^{\Delta e}}\right)
\right),
\qquad
X_\ell = X_h\,2^{\Delta e}.
\end{equation}
}
Hence $\hat{V}_{\ell,i}=X_\ell P_{\ell,i}\approx X_h P_{h,i}=\hat{V}_{h,i}$, where the residual error is due to rounding in the low-precision cast. Since the element value is in integer format and is divided by powers of two, this conversion is equivalent to a right shift and round on integer elements:
for each $P_{h,i}$, divide by $2^{\Delta e}$ (right shift by $\Delta e$), drop the
$\Delta e$ least-significant bits, round using the most-significant dropped bit,
and keep the rounded value in the target element representation. The scale update
$X_\ell=X_h 2^{\Delta e}$ compensates for the element down-scaling so the
reconstructed block remains numerically close to the high-precision one.

\subsection{Slice-and-Scale MXFP (SSMXFP) format conversion}
\label{sec:ssmxfp}
The same idea can be applied to MXFP.
\paragraph{MXFP representation.}
Let the MXFP element format be parameterized by exponent bits $\eta$ and
mantissa bits $\mu$ (total element bits $b=1+\eta+\mu$), with $\eta>=\mu$.
$\operatorname{quantize}_{\eta,\mu}(\cdot)$ denotes quantizing (including rounding and clamping) to the target MXFP element format.
Let $e_{\max}(\eta)$ denote the exponent of the largest normal
number in this MXFP format.
Following the same Eq. \ref{eq:scale},
{\small
\begin{equation}
\texttt{shared\_exp}
= \left\lfloor \log_2\!\left(\max_i |V_i|\right) \right\rfloor
- e_{\max}(\eta),
\qquad
X = 2^{\texttt{shared\_exp}},
\end{equation}
}
followed by
$P_i=\operatorname{quantize}_{\eta,\mu}(V_i/X)$.

\paragraph{High-to-low conversion via slice-and-scale.}
Let high precision use $(\eta_h,\mu_h)$ and low precision use $(\eta_\ell,\mu_\ell)$, with $\eta_h>\eta_\ell$ and $\Delta e=e_{\max}(\eta_h)-e_{\max}(\eta_\ell)$.
Following the same slice-and-scale rule,
{\small
\begin{equation}
P_{\ell,i} = \operatorname{quantize}_{\eta_\ell,\mu_\ell}\!\left(\frac{P_{h,i}}{2^{\Delta e}}\right),
\qquad
X_\ell = X_h\,2^{\Delta e}.
\end{equation}
}
Unlike MXINT, bit shifting cannot be applied for MXFP elements since it is not an integer format, instead, we do explicit division and quantization to the low-precision element format. Depending on hardware support, this cast can be implemented either directly from the high-precision anchor format to the lower-precision format or through an FP32 intermediate.

\subsection{Training and inference with anchor format storage}
\label{sec:ss_storage}
Let $A$ denote the anchor format (MXINT8 or MXFP8), and $t$ a runtime target format derived from $A$.

\paragraph{Training.}
We maintain full-precision master weights $W_{\mathrm{fp}}$. For each forward pass,
{\small
\begin{equation}
W_A = Q_A(W_{\mathrm{fp}}),
\qquad
W_t = Q_{A\rightarrow t}(W_A),
\end{equation}
}
where $Q_A$ quantizes to the anchor format and $Q_{A\rightarrow t}$ denotes SSMXINT or SSMXFP conversion. The model runs with $W_t$, and gradients are propagated through both operators using straight-through estimators \citep{yin_understanding_2019} to update $W_{\mathrm{fp}}$.

\paragraph{Inference.}
We store only the anchor checkpoint $W_A$. At runtime, we generate $W_t = Q_{A\rightarrow t}(W_A)$ on demand for any target format $t$. This enables elastic precision selection without retraining.

\section{Results}
\label{sec:results}
\subsection{Multi-format QAT across MXINT and MXFP}
\label{sec:results_multiformat_qat}
\begin{figure}[!t]
  \centering
  \includegraphics[width=0.49\linewidth]{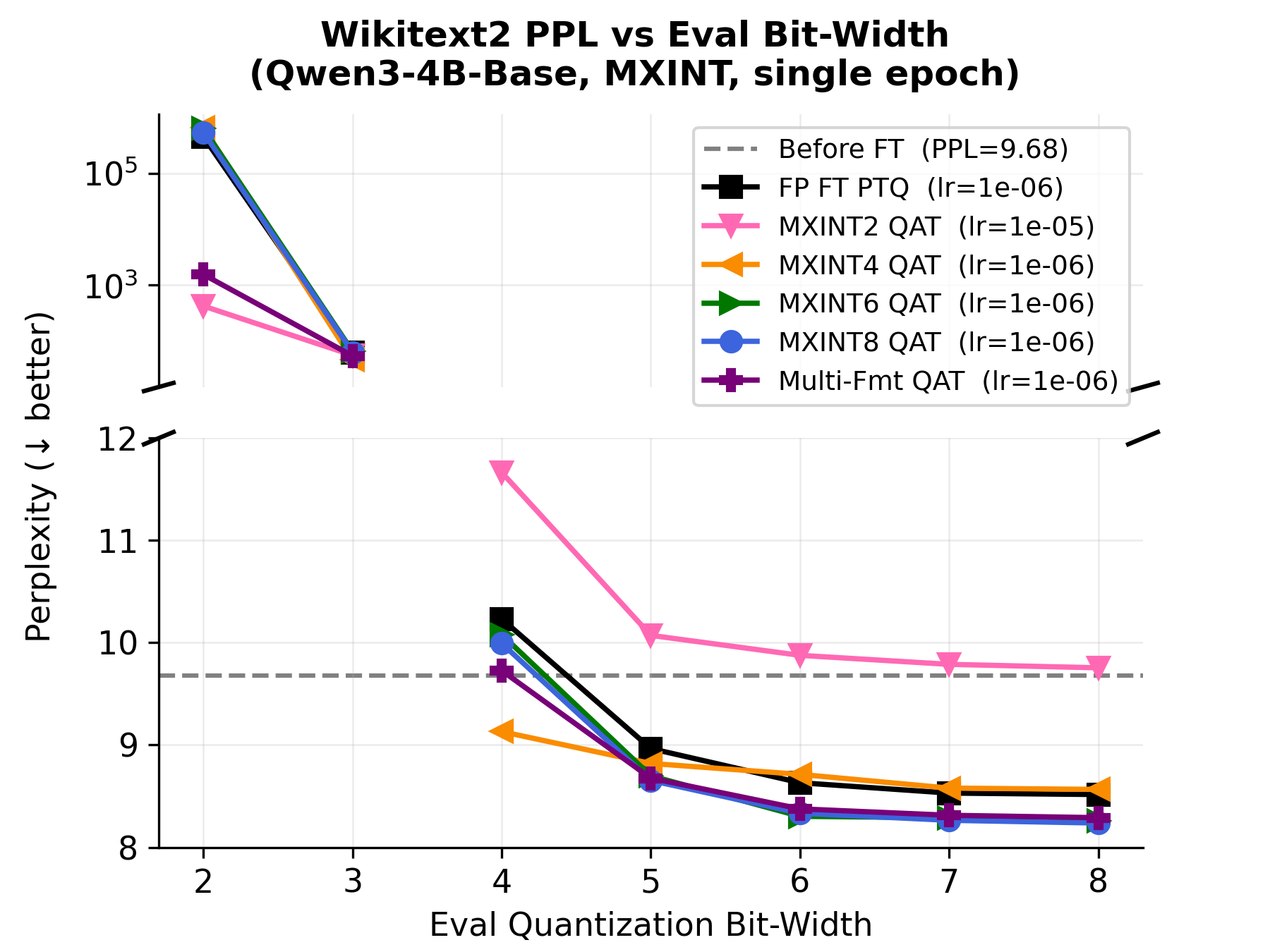}
  \hfill
  \includegraphics[width=0.49\linewidth]{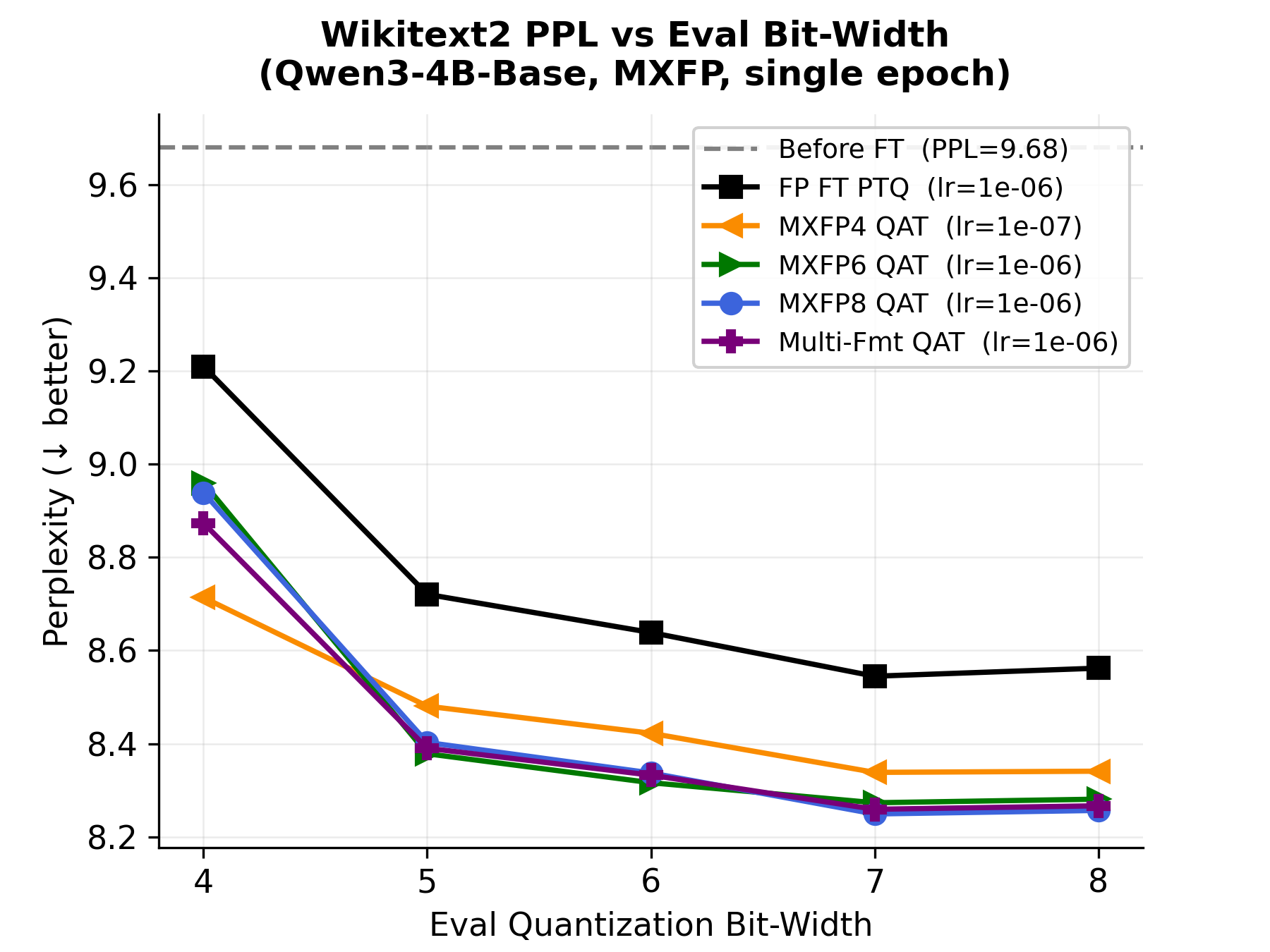}
  \caption{Multi-format QAT on \texttt{Qwen3-4B-Base}, trained for one epoch on 128 WikiText-2 examples. Left: MXINT formats. Right: MXFP formats. For each training variant, we show the result at the best learning rate, chosen as the one with the lowest evaluation loss among the three learning rates in Section~\ref{sec:methods_multiformat_qat}. The x-axis denotes evaluation quantization bit-width, and the y-axis denotes WikiText-2 validation perplexity after post-training quantization to the corresponding bit-width. The horizontal dashed line marks the original high-precision model (FP16/BF16/FP32, depending on pretraining). The black solid line corresponds to full-precision fine-tuning, the purple solid line to multi-format QAT, and the other colored solid lines to single-format QAT.}
  \label{fig:mfqat_main}
\end{figure}

We compare multi-format QAT against precision-specific single-format QAT and full-precision fine-tuning followed by post-training quantization across both MXINT and MXFP formats. Figure~\ref{fig:mfqat_main} reports zero-shot WikiText-2 perplexity for \texttt{Qwen3-4B-Base} as a function of evaluation bit-width. Results for all models are provided in Appendix~\ref{app:extra_multi_format_qat}.

\textbf{Brittleness of single-format QAT.}
As expected, single-format QAT performs best at its target bit-width. However, these models are highly brittle under format mismatch. In the MXINT setting (Figure~\ref{fig:mfqat_main}, left), models trained at 4, 6, or 8 bits degrade sharply when evaluated at 2 or 3 bits. Conversely, the 2-bit model performs poorly at higher precisions. This shows that precision-specific QAT tends to over-specialize to its training format.

\textbf{Robustness of multi-format QAT.}
In contrast, multi-format QAT (purple) closely tracks the best achievable performance across the full precision range. At the explicitly trained bit-widths ($\{2,4,6,8\}$), it remains consistently close to the corresponding single-format optimum, while also generalizing well to unseen intermediate bit-widths ($\{3,5,7\}$). Overall, multi-format QAT removes the need to commit to a single deployment precision, enabling robust performance under dynamic inference-time format selection.

\subsection{Multi-format QAT with downstream task}
Tables~\ref{tab:main_results_mxint_downstream} and~\ref{tab:main_results_mxfp_downstream} present the average 0-shot accuracy across MMLU~\citep{hendrycks_measuring_2021}, MathQA~\citep{amini_mathqa_2019}, and HellaSwag~\citep{zellers_hellaswag_2019} for six models from the Llama and Qwen families when quantized to MXINT and MXFP formats, respectively. 
Across all six models and both formats, multi-format QAT either outperforms the corresponding single-format QAT baselines or remains within $1\%$ accuracy of the best-performing model. The only exception is the MXINT2 format, where accuracy stays within $3\%$ for all models except \texttt{Llama-2-7B}. This is achieved despite training a single model to handle all precisions simultaneously.

For instance, on \texttt{Qwen3-1.7B} under MXINT (Table~\ref{tab:main_results_mxint_downstream}), multi-format QAT achieves the best or second-best accuracy at six out of seven evaluation precisions, including at unseen precisions such as MXINT5 ($53.7\%$) and MXINT7 ($54.7\%$). A similar trend holds under the MXFP formats (Table~\ref{tab:main_results_mxfp_downstream}), where multi-format QAT on \texttt{Qwen3-1.7B} reaches $54.3\%$ and $54.8\%$ at MXFP6 and MXFP7-matching or exceeding all single-format baselines. These results confirm that the downstream task performance mirrors the perplexity trends observed in Section~\ref{sec:results_multiformat_qat}: a single multi-format QAT model can robustly serve diverse deployment precisions with negligible-or no-additional accuracy degradation at individual formats, eliminating the need to train and store separate models for each target precision.

We further evaluate multi-format QAT on the multimodal Qwen3-VL models using ChartQA (Table~\ref{tab:main_results_multimodal_downstream}). Despite higher variance inherent to visual reasoning tasks, multi-format QAT remains competitive across precisions-for example, achieving the best accuracy at MXINT6 ($30.0\%$) on \texttt{Qwen3-VL-4B} and matching the top single-format result at MXFP7 ($39.0\%$) on \texttt{Qwen3-VL-2B}. These results confirm that the benefits of multi-format QAT extend to multimodal workloads.

\begin{table*}[t!]
\centering
\resizebox{0.97\textwidth}{!}{
\begin{tabular}{c|c|ccccccc}
\hline
 \multirow{2}{*}{\textbf{Model}} & \multirow{2}{*}{\textbf{QAT/FT Precision}} & \multicolumn{7}{c}{\textbf{PTQ Precision}}  \\ \cline{3-9} 
  &  & \textbf{MXINT2} & \textbf{MXINT3$^*$} & \textbf{MXINT4} & \textbf{MXINT5$^*$} & \textbf{MXINT6} & \textbf{MXINT7$^*$} & \textbf{MXINT8} \\ \hline 
 & Full Precision FT & 22.7 & 37.5 & 44.4 & 44.7 & \textbf{45.8} & 46.0 & 46.0 \\ \cdashline{2-9} 
 & MXINT2 QAT        & \textbf{29.3} & \textbf{39.6} & 44.4 & 44.6 & 45.1 & 45.6 & 45.1 \\
 & MXINT4 QAT        & 22.9 & 38.7 & \textbf{44.6} & 44.8 & 45.6 & 45.9 & 46.0 \\
 & MXINT6 QAT        & 22.1 & 38.4 & 43.7 & 44.6 & 45.2 & 46.1 & 46.0 \\
 & MXINT8 QAT        & 21.9 & 37.7 & 44.2 & 44.8 & 45.2 & 46.0 & 46.0 \\
\multirow{-6}{*}{Llama-2-7B} & \cellcolor[HTML]{CCFACC}Multi-format QAT & \cellcolor[HTML]{CCFACC}23.1 & \cellcolor[HTML]{CCFACC}38.7 & \cellcolor[HTML]{CCFACC}43.9 & \cellcolor[HTML]{CCFACC}\textbf{44.9} & \cellcolor[HTML]{CCFACC}45.5 & \cellcolor[HTML]{CCFACC}\textbf{46.2} & \cellcolor[HTML]{CCFACC}\textbf{46.1} \\ \hline 
 & Full Precision FT & 21.9 & 25.6 & 34.9 & 40.9 & 41.3 & \textbf{41.3} & 41.2 \\ \cdashline{2-9} 
 & MXINT2 QAT & \textbf{23.6} & 26.2 & 34.7 & 40.2 & 41.2 & 40.6 & 40.8 \\
 & MXINT4 QAT & 22.8 & \textbf{28.3} & \textbf{36.1} & 41.1 & \textbf{41.4} & 40.9 & 41.5 \\
 & MXINT6 QAT & 22.4 & 25.8 & 35.1 & 41.3 & \textbf{41.4} & 40.7 & 41.5 \\
 & MXINT8 QAT & 22.1 & 25.6 & 35.2 & \textbf{41.4} & \textbf{41.4} & \textbf{41.3} & \textbf{41.6} \\
\multirow{-6}{*}{Llama-3.2-1B} & \cellcolor[HTML]{CCFACC}Multi-format QAT & \cellcolor[HTML]{CCFACC}23.3 & \cellcolor[HTML]{CCFACC}28.2 & \cellcolor[HTML]{CCFACC}35.2 & \cellcolor[HTML]{CCFACC}40.8 & \cellcolor[HTML]{CCFACC}41.3 & \cellcolor[HTML]{CCFACC}40.8 & \cellcolor[HTML]{CCFACC}41.4 \\ \hline 
 & Full Precision FT & 25.2 & 35.6 & 49.5 & 51.1 & 52.1 & 51.7 & 51.8 \\
 & MXINT2 QAT & \textbf{25.5} & 36.6 & 49.2 & 51.4 & 52.3 & \textbf{52.3} & 52.0 \\
 & MXINT4 QAT & 24.5 & 36.6 & 49.3 & 51.4 & 52.5 & 51.7 & 51.6 \\
 & MXINT6 QAT & 24.1 & 35.4 & 49.4 & \textbf{51.5} & \textbf{52.6} & 51.8 & 51.9 \\
 & MXINT8 QAT & 24.9 & 35.5 & 49.4 & 51.4 & \textbf{52.6} & 51.9 & 51.8 \\
\multirow{-6}{*}{Llama-3.2-3B} & \cellcolor[HTML]{CCFACC}Multi-format QAT & \cellcolor[HTML]{CCFACC}23.9 & \cellcolor[HTML]{CCFACC}\textbf{36.7} & \cellcolor[HTML]{CCFACC}\textbf{49.7} & \cellcolor[HTML]{CCFACC}51.2 & \cellcolor[HTML]{CCFACC}52.4 & \cellcolor[HTML]{CCFACC}52.0 & \cellcolor[HTML]{CCFACC}\textbf{52.1} \\ \hline 
 & Full Precision FT & 24.0 & 25.0 & 38.4 & 43.5 & 45.9 & 46.7 & \textbf{46.9} \\ \cdashline{2-9} 
 & MXINT2 QAT & \textbf{24.4} & 24.9 & 38.1 & 42.3 & 45.6 & 45.8 & 45.9 \\
 & MXINT4 QAT & 23.3 & 25.3 & \textbf{39.6} & \textbf{44.1} & \textbf{47.2} & 46.1 & 46.5 \\
 & MXINT6 QAT & 23.6 & 24.7 & 38.3 & 43.8 & 46.4 & \textbf{47.2} & 46.8 \\
 & MXINT8 QAT & 22.6 & \textbf{25.4} & 38.0 & 43.7 & 46.5 & 47.0 & 46.8 \\
\multirow{-6}{*}{Qwen3-0.6B} & \cellcolor[HTML]{CCFACC}Multi-format QAT & \cellcolor[HTML]{CCFACC}22.0 & \cellcolor[HTML]{CCFACC}25.2 & \cellcolor[HTML]{CCFACC}\textbf{39.6} & \cellcolor[HTML]{CCFACC}43.7 & \cellcolor[HTML]{CCFACC}46.1 & \cellcolor[HTML]{CCFACC}46.9 & \cellcolor[HTML]{CCFACC}46.2 \\ \hline  
 & Full Precision FT & \textbf{24.1} & 29.2 & 48.6 & 53.1 & 54.2 & 54.6 & 55.2 \\ \cdashline{2-9} 
 & MXINT2 QAT & 22.7 & 28.8 & 48.7 & 53.4 & 54.3 & 54.1 & \textbf{55.5} \\
 & MXINT4 QAT & 22.7 & \textbf{29.5} & 48.6 & 52.9 & \textbf{55.1} & \textbf{54.7} & 55.4 \\
 & MXINT6 QAT & 22.5 & 29.1 & 48.9 & \textbf{53.7} & 54.5 & 54.6 & 55.1 \\
 & MXINT8 QAT & 23.9 & 28.6 & 48.8 & 53.2 & 54.7 & 54.3 & 55.1 \\
\multirow{-6}{*}{Qwen3-1.7B} & \cellcolor[HTML]{CCFACC}Multi-format QAT & \cellcolor[HTML]{CCFACC}22.7 & \cellcolor[HTML]{CCFACC}29.3 & \cellcolor[HTML]{CCFACC}\textbf{49.4} & \cellcolor[HTML]{CCFACC}\textbf{53.7} & \cellcolor[HTML]{CCFACC}54.7 & \cellcolor[HTML]{CCFACC}\textbf{54.7} & \cellcolor[HTML]{CCFACC}\textbf{55.5} \\ \hline   
 & Full Precision FT & \textbf{24.6} & 31.1 & 60.1 & 60.5 & 60.9 & 61.7 & 61.7 \\ \cdashline{2-9} 
 & MXINT2 QAT & 22.7 & 31.6 & 59.4 & 60.2 & 60.9 & 61.9 & 62.0 \\
 & MXINT4 QAT & 23.3 & \textbf{32.8} & \textbf{60.8} & 60.5 & 61.6 & 62.2 & 62.0 \\
 & MXINT6 QAT & 22.2 & 31.9 & 60.1 & \textbf{60.6} & \textbf{61.8} & \textbf{62.7} & \textbf{62.9} \\
 & MXINT8 QAT & 23.0 & 31.4 & 60.1 & 60.5 & 61.6 & 62.4 & 62.8 \\
\multirow{-6}{*}{Qwen3-4B} & \cellcolor[HTML]{CCFACC}Multi-format QAT & \cellcolor[HTML]{CCFACC}23.9 & \cellcolor[HTML]{CCFACC}31.8 & \cellcolor[HTML]{CCFACC}60.0 & \cellcolor[HTML]{CCFACC}60.4 & \cellcolor[HTML]{CCFACC}61.5 & \cellcolor[HTML]{CCFACC}61.9 & \cellcolor[HTML]{CCFACC}62.4 \\ \bottomrule 
\end{tabular}
}

\caption{Average 0-shot accuracy (higher is better) on MMLU~\citep{hendrycks_measuring_2021}, MathQA~\citep{amini_mathqa_2019}, and HellaSwag~\citep{zellers_hellaswag_2019} for \textit{the MXINT formats}. Each row corresponds to a different QAT/FT training precision, and each column shows the PTQ evaluation precision. Columns marked with $^*$ denote precisions not seen during multi-format/QAT training. QAT, PTQ, and FT denote Quantization Aware Training, Post Training Quantization, and Finetuning, respectively. Detailed per-individual task results in Appendix~\ref{sec:appendix_complete_results}, Tables~\ref{tab:main_results_mxint_downstream_MMLU}-\ref{tab:main_results_mxint_downstream_HellaSwag}.
}
\label{tab:main_results_mxint_downstream}
\end{table*}

\begin{table*}[t!]
\centering
\resizebox{0.9\textwidth}{!}{
\begin{tabular}{c|c|ccccc}
\hline
 \multirow{2}{*}{\textbf{Model}} & \multirow{2}{*}{\textbf{QAT/FT Precision}} & \multicolumn{5}{c}{\textbf{PTQ Precision}}  \\ \cline{3-7} 
  &  & \textbf{MXFP4} & \textbf{MXFP5$^*$} & \textbf{MXFP6} & \textbf{MXFP7$^*$} & \textbf{MXFP8} \\ \hline 
 & Full Precision FT & 43.3 & 45.5 & 45.7 & \textbf{46.0} & \textbf{46.2} \\ \cdashline{2-7} 
 & MXFP4 QAT        & \textbf{43.7} & \textbf{46.0} & \textbf{46.1} & 45.9 & 45.6 \\
 & MXFP6 QAT        & 43.3 & 45.5 & 45.8 & 45.7 & 45.8 \\
 & MXFP8 QAT        & 43.6 & 45.4 & 45.7 & \textbf{46.0} & 45.8 \\
\multirow{-5}{*}{Llama-2-7B} & \cellcolor[HTML]{CCFACC}Multi-format QAT & \cellcolor[HTML]{CCFACC}43.5 & \cellcolor[HTML]{CCFACC}45.5 & \cellcolor[HTML]{CCFACC}45.8 & \cellcolor[HTML]{CCFACC}45.8 & \cellcolor[HTML]{CCFACC}45.8 \\ \hline
 & Full Precision FT & 37.4 & 41.1 & \textbf{41.5} & 41.5 & 41.7 \\ \cdashline{2-7} 
 & MXFP4 QAT & \textbf{38.2} & 41.4 & 41.2 & \textbf{42.0} & \textbf{41.8} \\
 & MXFP6 QAT & 37.3 & 41.1 & 41.1 & 41.8 & 41.4 \\
 & MXFP8 QAT & 37.8 & 41.0 & 41.1 & 41.8 & \textbf{41.8} \\
\multirow{-5}{*}{Llama-3.2-1B} & \cellcolor[HTML]{CCFACC}Multi-format QAT & \cellcolor[HTML]{CCFACC}37.6 & \cellcolor[HTML]{CCFACC}\textbf{41.5} & \cellcolor[HTML]{CCFACC}41.2 & \cellcolor[HTML]{CCFACC}41.4 & \cellcolor[HTML]{CCFACC}41.5 \\ \hline
& Full Precision FT & \textbf{50.6} & \textbf{52.5} & 51.6 & 51.8 & 51.6 \\ \cdashline{2-7} 
 & MXFP4 QAT & 50.2 & 51.9 & 51.8 & 51.9 & \textbf{51.8} \\
 & MXFP6 QAT & 50.0 & 52.3 & 51.9 & \textbf{52.0} & 51.7 \\
 & MXFP8 QAT & 50.2 & \textbf{52.5} & \textbf{52.3} & 51.8 & 51.7 \\
\multirow{-5}{*}{Llama-3.2-3B} & \cellcolor[HTML]{CCFACC}Multi-format QAT & \cellcolor[HTML]{CCFACC}49.9 & \cellcolor[HTML]{CCFACC}\textbf{52.5} & \cellcolor[HTML]{CCFACC}52.2 & \cellcolor[HTML]{CCFACC}51.8 & \cellcolor[HTML]{CCFACC}51.7 \\ \hline 
 & Full Precision FT & \textbf{44.0} & 46.1 & 46.3 & 45.7 & 45.5 \\ \cdashline{2-7} 
 & MXFP4 QAT & 43.7 & 45.9 & 46.7 & \textbf{46.1} & \textbf{46.2} \\
 & MXFP6 QAT & 43.7 & \textbf{46.5} & 46.7 & 45.6 & 45.4 \\
 & MXFP8 QAT & 43.6 & \textbf{46.5} & \textbf{46.8} & 45.6 & 45.2 \\
\multirow{-5}{*}{Qwen3-0.6B} & \cellcolor[HTML]{CCFACC}Multi-format QAT & \cellcolor[HTML]{CCFACC}43.5 & \cellcolor[HTML]{CCFACC}46.1 & \cellcolor[HTML]{CCFACC}46.4 & \cellcolor[HTML]{CCFACC}45.6 & \cellcolor[HTML]{CCFACC}45.4 \\ \hline  
 & Full Precision FT & 51.8 & 53.7 & 53.7 & 54.3 & 54.7 \\ \cdashline{2-7} 
 & MXFP4 QAT & \textbf{52.8} & 53.7 & 53.9 & \textbf{54.8} & \textbf{55.2} \\
 & MXFP6 QAT & 51.7 & 54.0 & \textbf{54.3} & 54.3 & 54.7 \\
 & MXFP8 QAT & 52.3 & \textbf{54.3} & 54.0 & 54.4 & 54.4 \\
\multirow{-5}{*}{Qwen3-1.7B} & \cellcolor[HTML]{CCFACC}Multi-format QAT & \cellcolor[HTML]{CCFACC}52.2 & \cellcolor[HTML]{CCFACC}54.1 & \cellcolor[HTML]{CCFACC}\textbf{54.3} & \cellcolor[HTML]{CCFACC}\textbf{54.8} & \cellcolor[HTML]{CCFACC}54.7 \\ \hline
 & Full Precision FT & 60.1 & 61.2 & 61.1 & 61.4 & 61.4 \\ \cdashline{2-7} 
 & MXFP4 QAT & \textbf{60.3} & \textbf{62.2} & 61.9 & 62.3 & \textbf{62.4} \\
 & MXFP6 QAT & \textbf{60.3} & 61.6 & 61.5 & \textbf{62.4} & 62.0 \\
 & MXFP8 QAT & 60.2 & 61.9 & 62.1 & 62.3 & 62.2 \\
\multirow{-5}{*}{Qwen3-4B} & \cellcolor[HTML]{CCFACC}Multi-format QAT & \cellcolor[HTML]{CCFACC}\textbf{60.3} & \cellcolor[HTML]{CCFACC}61.9 & \cellcolor[HTML]{CCFACC}\textbf{62.2} & \cellcolor[HTML]{CCFACC}\textbf{62.4} & \cellcolor[HTML]{CCFACC}62.3 \\ \bottomrule 
\end{tabular}
}

\caption{Average 0-shot accuracy (higher is better) on MMLU~\citep{hendrycks_measuring_2021}, MathQA~\citep{amini_mathqa_2019}, and HellaSwag~\citep{zellers_hellaswag_2019} for \textit{the MXFP formats}. Each row corresponds to a different QAT/FT training precision, and each column shows the PTQ evaluation precision. Columns marked with $^*$ denote precisions not seen during multi-format/QAT training. QAT, PTQ, and FT denote Quantization Aware Training, Post Training Quantization, and Finetuning, respectively. Detailed per-individual task results in Appendix~\ref{sec:appendix_complete_results}, Table~\ref{tab:main_results_mxfp_downstream_detailed}.
}
\label{tab:main_results_mxfp_downstream}
\end{table*}

\begin{table*}[t!]
\centering
\resizebox{0.9\textwidth}{!}{
\begin{tabular}{c|c|ccccc}
\hline
 \multirow{2}{*}{\textbf{Model}} & \multirow{2}{*}{\textbf{QAT/FT Precision}} & \multicolumn{5}{c}{\textbf{PTQ Precision}}  \\ \cline{3-7} 
  &  & \textbf{MXINT4} & \textbf{MXINT5$^*$} & \textbf{MXINT6} & \textbf{MXINT7$^*$} & \textbf{MXINT8} \\ \hline 
 & Full Precision FT & 15.0 & 52.0 & 36.0 & \textbf{44.0} & 35.0 \\ \cdashline{2-7} 
 & MXINT4 QAT        & \textbf{23.0} & \textbf{55.0} & 32.0 & 39.0 & 38.0 \\
 & MXINT6 QAT        & 17.0 & 50.0 & 37.0 & 41.0 & \textbf{40.0} \\
 & MXINT8 QAT        & 20.0 & 47.0 & \textbf{40.0} & 40.0 & 32.0 \\
\multirow{-5}{*}{Qwen3-VL-2B} & \cellcolor[HTML]{CCFACC}Multi-format QAT & \cellcolor[HTML]{CCFACC}15.0 & \cellcolor[HTML]{CCFACC}48.0 & \cellcolor[HTML]{CCFACC}31.0 & \cellcolor[HTML]{CCFACC}34.0 & \cellcolor[HTML]{CCFACC}34.0 \\ \hline 
 & Full Precision FT & 23.0 & \textbf{25.0} & 28.0 & \textbf{23.0} & 25.0 \\ \cdashline{2-7} 
 & MXINT4 QAT & 24.0 & 24.0 & \textbf{30.0} & \textbf{23.0} & \textbf{28.0} \\
 & MXINT6 QAT & \textbf{28.0} & 21.0 & 27.0 & 22.0 & 24.0 \\
 & MXINT8 QAT & 27.0 & 22.0 & 28.0 & 22.0 & 26.0 \\
\multirow{-5}{*}{Qwen3-VL-4B} & \cellcolor[HTML]{CCFACC}Multi-format QAT & \cellcolor[HTML]{CCFACC}25.0 & \cellcolor[HTML]{CCFACC}23.0 & \cellcolor[HTML]{CCFACC}\textbf{30.0} & \cellcolor[HTML]{CCFACC}22.0 & \cellcolor[HTML]{CCFACC}23.0 \\ \hline \hline 
\multirow{2}{*}{\textbf{Model}} & \multirow{2}{*}{\textbf{QAT/FT Precision}} & \multicolumn{5}{c}{\textbf{PTQ Precision}}  \\ \cline{3-7} 
  &  & \textbf{MXFP4} & \textbf{MXFP5$^*$} & \textbf{MXFP6} & \textbf{MXFP7$^*$} & \textbf{MXFP8} \\ \hline 
 & Full Precision FT & 30.0 & 33.0 & 29.0 & 30.0 & 27.0 \\ \cdashline{2-7}
 & MXFP4 QAT & \textbf{42.0} & \textbf{37.0} & \textbf{34.0} & \textbf{39.0} & \textbf{43.0} \\
 & MXFP6 QAT & 33.0 & 34.0 & 31.0 & 38.0 & 37.0 \\
 & MXFP8 QAT & 35.0 & 35.0 & \textbf{34.0} & 32.0 & 28.0 \\
\multirow{-5}{*}{Qwen3-VL-2B} & \cellcolor[HTML]{CCFACC}Multi-format QAT & \cellcolor[HTML]{CCFACC}32.0 & \cellcolor[HTML]{CCFACC}34.0 & \cellcolor[HTML]{CCFACC}33.0 & \cellcolor[HTML]{CCFACC}\textbf{39.0} & \cellcolor[HTML]{CCFACC}36.0 \\ \hline 
 & Full Precision FT & \textbf{18.0} & \textbf{29.0} & \textbf{27.0} & 23.0 & \textbf{24.0} \\ \cdashline{2-7} 
 & MXFP4 QAT & 16.0 & 27.0 & 24.0 & 24.0 & \textbf{24.0} \\
 & MXFP6 QAT & \textbf{18.0} & 27.0 & 25.0 & \textbf{27.0} & 20.0 \\
 & MXFP8 QAT & \textbf{18.0} & \textbf{29.0} & 25.0 & 24.0 & 23.0 \\
\multirow{-5}{*}{Qwen3-VL-4B} & \cellcolor[HTML]{CCFACC}Multi-format QAT & \cellcolor[HTML]{CCFACC}16.0 & \cellcolor[HTML]{CCFACC}26.0 & \cellcolor[HTML]{CCFACC}24.0 & \cellcolor[HTML]{CCFACC}24.0 & \cellcolor[HTML]{CCFACC}23.0 \\  \bottomrule 
\end{tabular}
}

\caption{ChartQA~\citep{masry_chartqa_2022} accuracy for Qwen3-VL~\citep{bai_qwen3-vl_2025} multi-modal models when quantized to the MXINT and MXFP formats with different precisions. Rows indicate the QAT/FT training precision; columns indicate the PTQ inference precision. Columns marked with $^*$ denote precisions not seen during multi-format/QAT training. QAT, PTQ, and FT denote Quantization Aware Training, Post Training Quantization, and Finetuning, respectively.}
\label{tab:main_results_multimodal_downstream}
\end{table*}

\subsection{Effectiveness of Slice-and-Scale conversion}
\label{sec:results_ss_conversion}
The previous subsection shows that multi-format QAT is robust across target precisions while requiring only a single full-precision model during training. We now show that storage can be reduced further by replacing this with an 8-bit anchor checkpoint, which is converted on demand to lower-precision MX formats using Slice-and-Scale (SS). The details of the SS algorithm are described in Sections~\ref{sec:ssmxint} and \ref{sec:ssmxfp}.

We evaluate SS conversion using two metrics: average layer-wise MSE on 100 random tensors of shape $(1,1024)$, and zero-shot WikiText-2 perplexity for \texttt{Llama-3.2-1B}. Here, MXINT/MXFP denote direct quantization into target OCP MX formats \citep{rouhani_ocp_nodate}, while SSMXINT/SSMXFP denote our conversion from an 8-bit anchor (MXINT8/MXFP8) to lower precisions without access to full-precision weights.

SS closely matches direct quantization at the tensor level across bit precisions and block sizes for both MXINT and MXFP. Full MSE results are provided in Appendix~\ref{app:ss_tensor_level}.

\begin{figure*}[ht]
  \centering
  \includegraphics[width=0.49\textwidth]{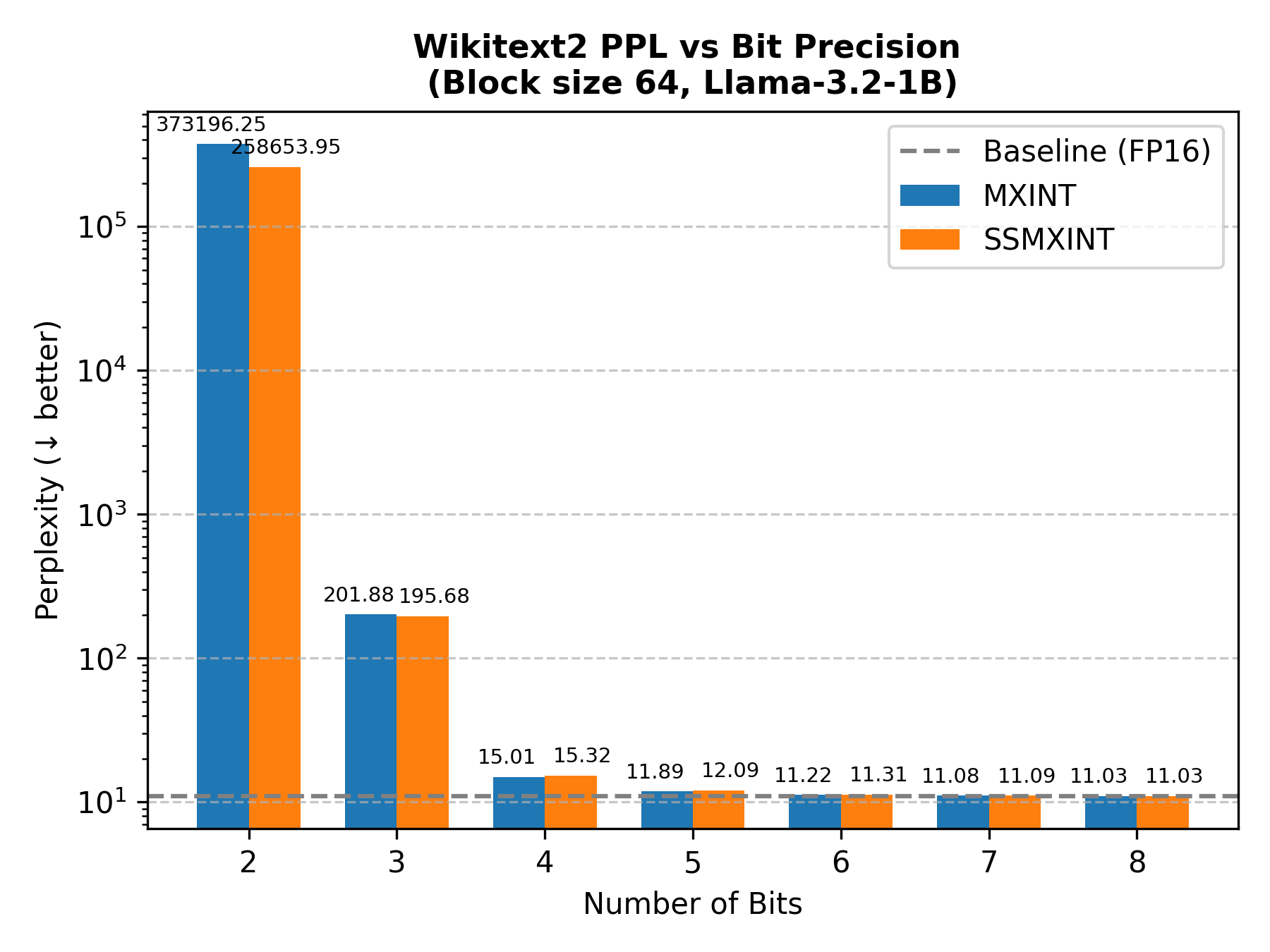}
  \hfill
  \includegraphics[width=0.49\textwidth]{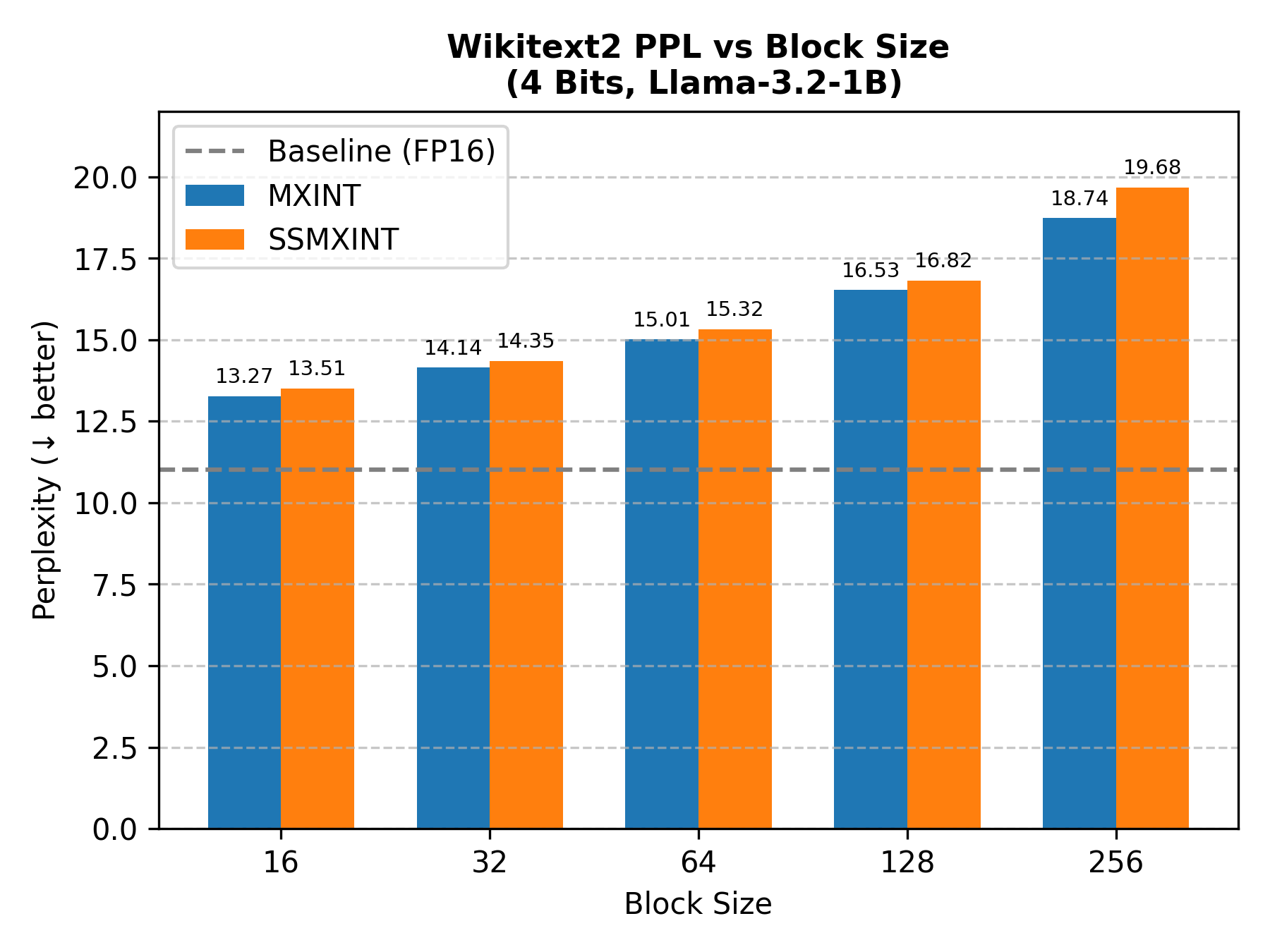}
  \caption{WikiText-2 perplexity for \texttt{Llama-3.2-1B}, comparing direct MXINT quantization and SSMXINT. \textbf{Left:} Varying bit precision at block size 64. \textbf{Right:} Varying block size at 4-bit precision. Horizontal dashed line denotes the baseline model.}
  \label{fig:ssmxint_study_ppl}
\end{figure*}

\begin{figure*}[!t]
  \centering
  \includegraphics[width=0.49\textwidth]{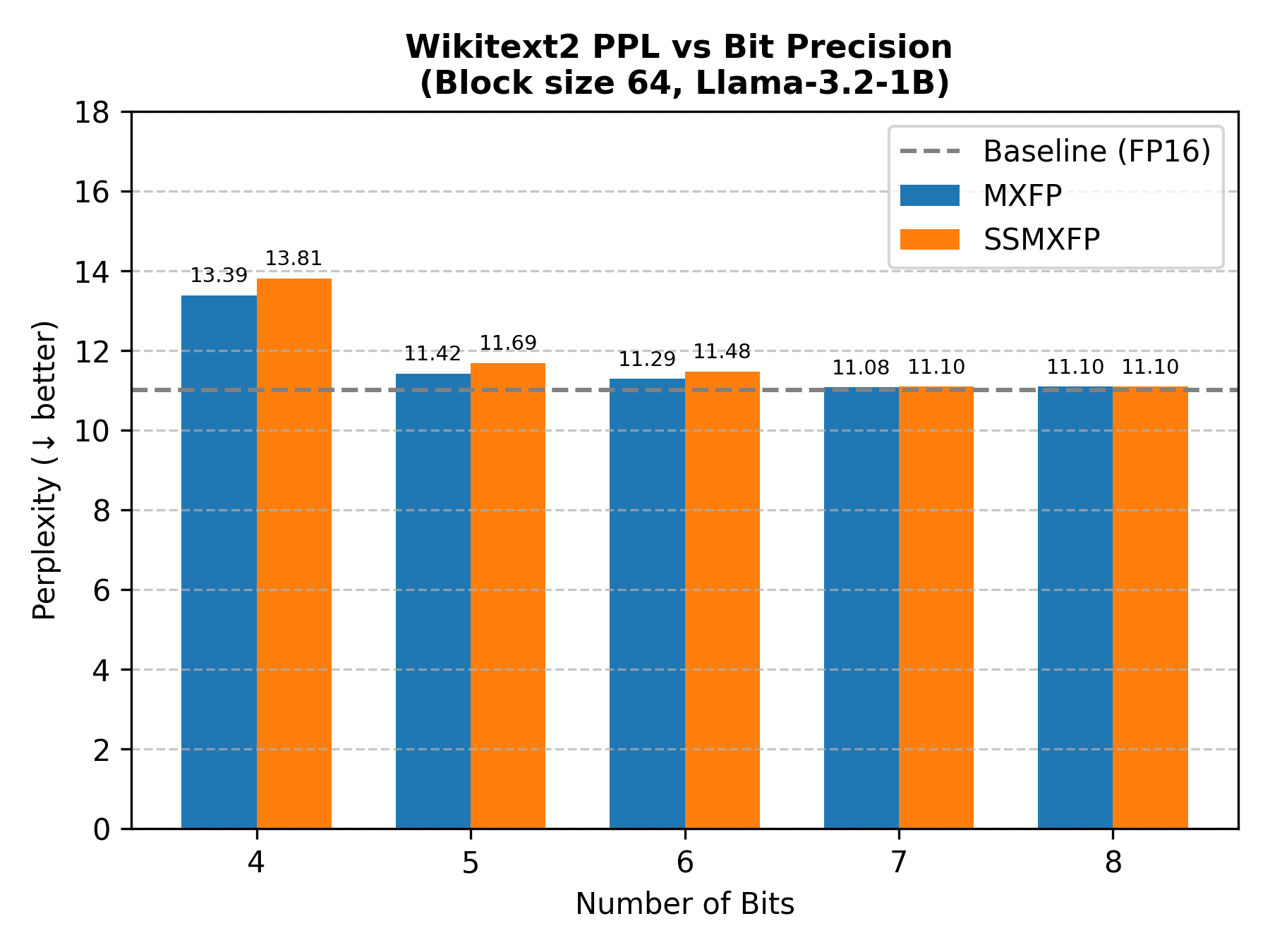}
  \hfill
  \includegraphics[width=0.49\textwidth]{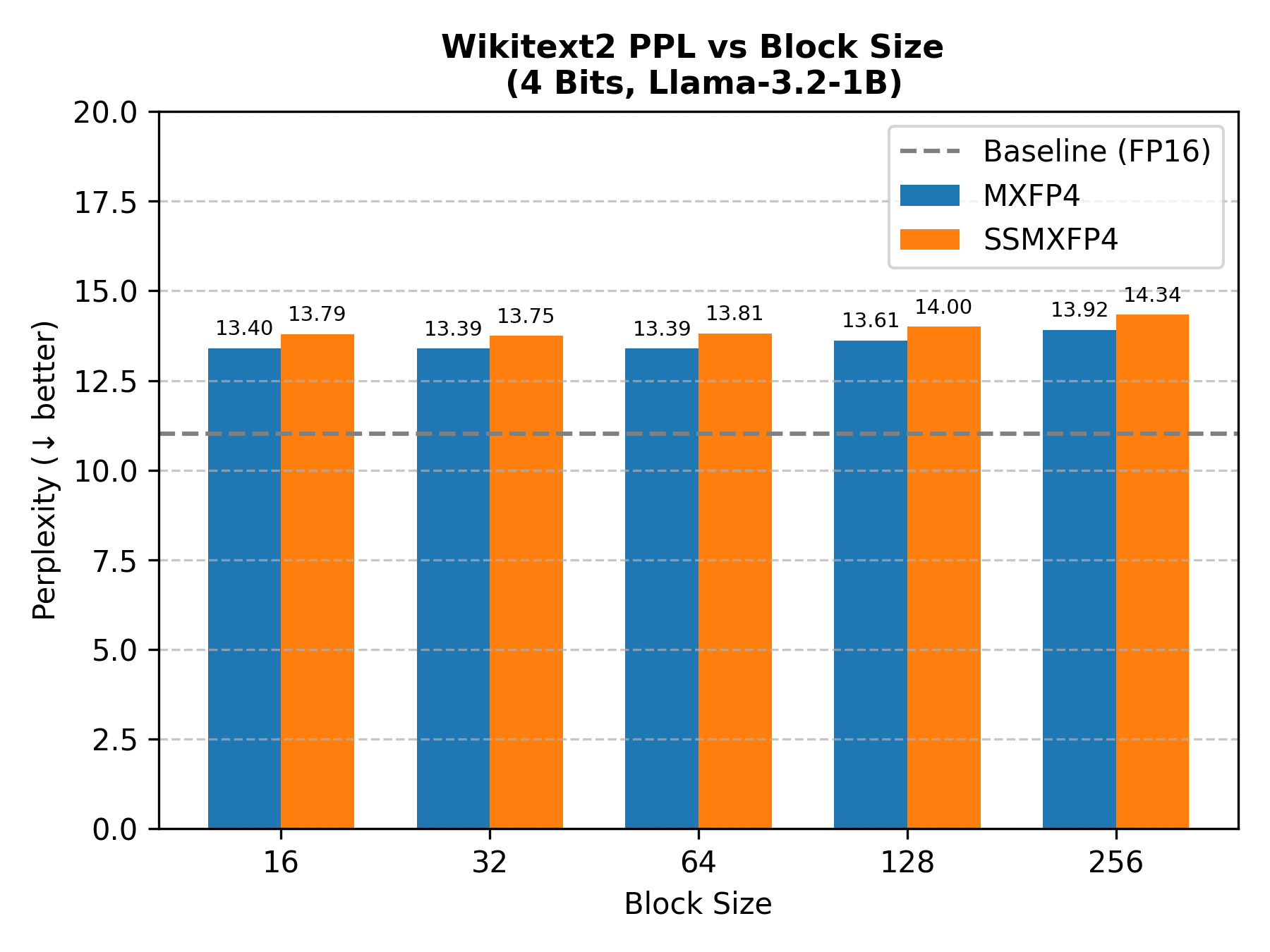}
  \caption{WikiText-2 perplexity for \texttt{Llama-3.2-1B}, comparing direct MXFP quantization and SSMXFP. \textbf{Left:} Varying bit precision at block size 64. \textbf{Right:} Varying block size at 4-bit precision. Horizontal dashed line denotes the baseline model.}
  \label{fig:ssmxfp_study_ppl}
\end{figure*}

\textbf{End-to-end language modeling performance.}
Figures~\ref{fig:ssmxint_study_ppl} and~\ref{fig:ssmxfp_study_ppl} show that SS achieves WikiText-2 perplexity nearly identical to direct target-format quantization across all evaluated settings. These results show that a single higher-precision anchor checkpoint can be down-converted at runtime to lower-precision MX formats with minimal loss.

\subsection{Multi-format QAT with Slice and Scale}
\label{sec:results_multiformat_ss}
\begin{figure}[!t]
  \centering
  \includegraphics[width=0.49\linewidth]{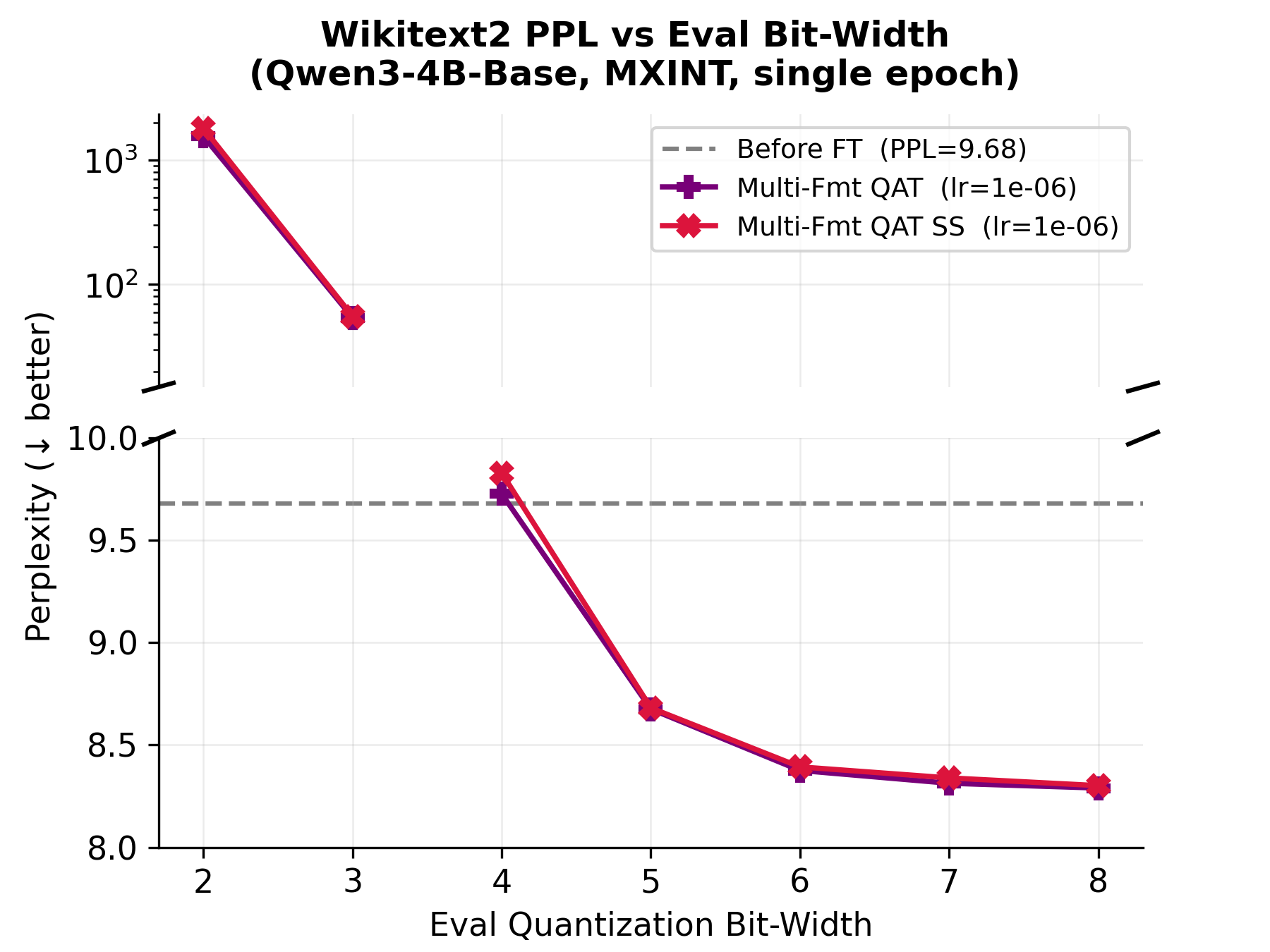}
  \hfill
  \includegraphics[width=0.49\linewidth]{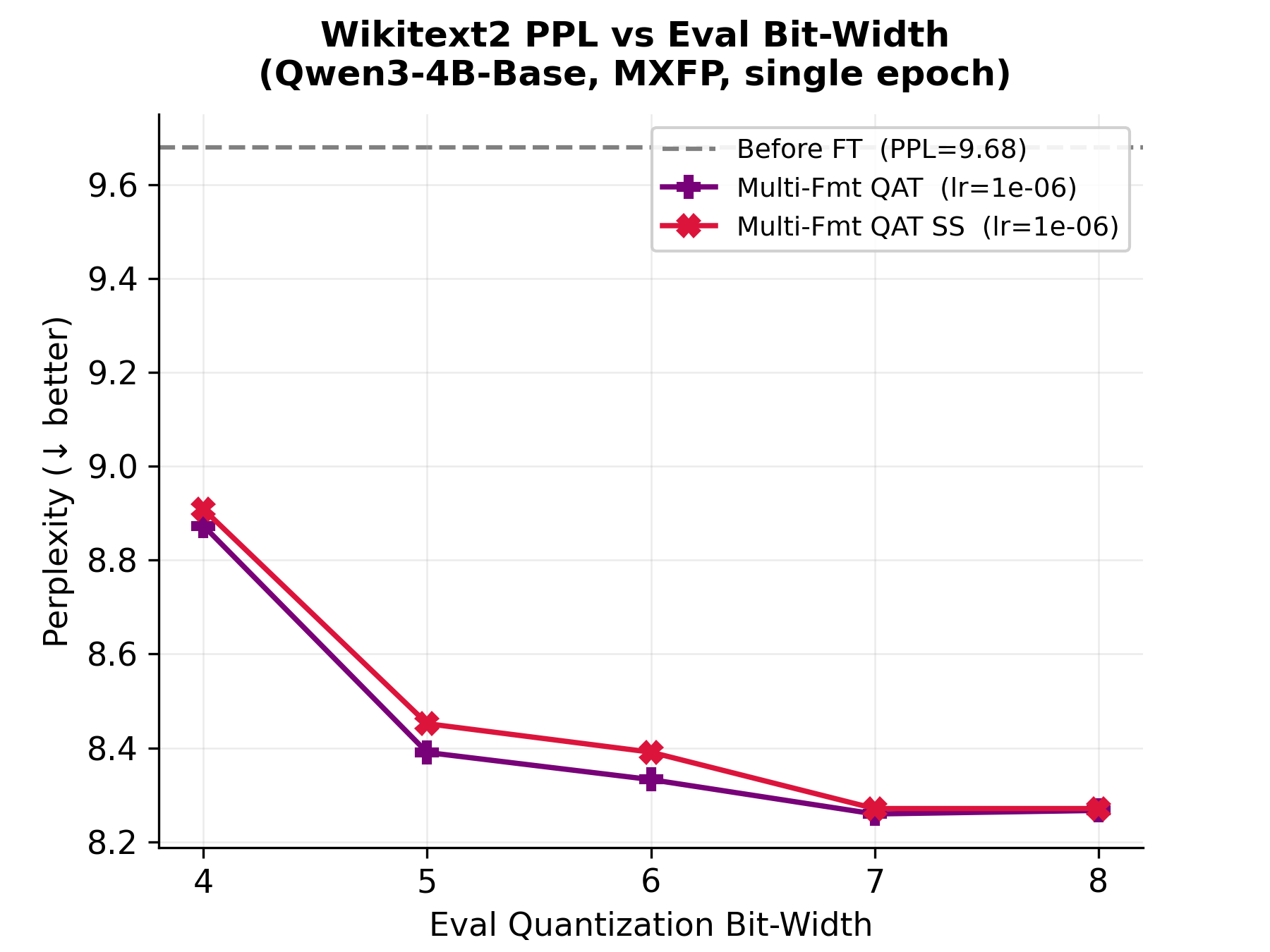}
  \caption{Multi-format QAT with Slice-and-Scale on \texttt{Qwen3-4B-Base}. \textbf{Left:} MXINT formats. \textbf{Right:} MXFP formats. Plotting conventions match Figure~\ref{fig:mfqat_main}.}
  \label{fig:mfqat_ss_main}
\end{figure}

We next evaluate the full pipeline combining multi-format QAT with Slice-and-Scale (SS) conversion. Figure~\ref{fig:mfqat_ss_main} shows results for \texttt{Qwen3-4B-Base}. Results for all models are deferred to Appendix~\ref{app:extra_multi_format_ss}. We train this variant using the anchor-storage procedure in Section~\ref{sec:ss_storage}, cycling through target formats uniformly during training.

Figure~\ref{fig:mfqat_ss_main} shows that the Slice-and-Scale variant (red) closely matches standard multi-format QAT (purple) across the full precision range. For MXINT, the two curves are nearly indistinguishable; for MXFP, only a small gap appears at intermediate bit-widths, consistent with Section~\ref{sec:results_ss_conversion}.

These results show that a single low-precision anchor checkpoint, together with on-the-fly Slice-and-Scale conversion, recovers nearly the same performance as standard multi-format QAT while reducing checkpoint storage overhead.

\section{Conclusion}
\label{sec:conclusion}

We presented a study of multi-format quantization-aware training (MF-QAT) and showed that training a single model across multiple quantization formats can match the performance of single-format QAT at each target precision, providing a single model that remains effective across formats. To make these benefits deployable, we introduced a Microscaling format transformation method that derives a lower-precision MXINT/MXFP representation from a higher-precision one without retraining. Finally, we proposed an elastic inference scheme that stores a single low-precision checkpoint and enables further on-the-fly quantization to even lower precision at runtime with minimal loss degradation.
These results suggest a practical path to decoupling training from rigid
deployment-format choices, improving robustness across heterogeneous hardware, and enabling dynamic precision scaling under varying runtime constraints.

\clearpage
\bibliography{references}
\bibliographystyle{Template-2026/colm2026_conference}

\clearpage
\appendix

\section{Wikitext2 PPL Eval results for all models}

\subsection{Additional plots for Multi-format QAT across MXINT and MXFP}
\label{app:extra_multi_format_qat}
This appendix contains plots for
Section~\ref{sec:results_multiformat_qat} on all models.  Each figure below contains results for one model with left figure for MXINT and right figure for MXFP. Plotting conventions follow Figure \ref{fig:mfqat_main}.

\begin{figure*}[h]
  \centering
  \includegraphics[width=0.49\textwidth]{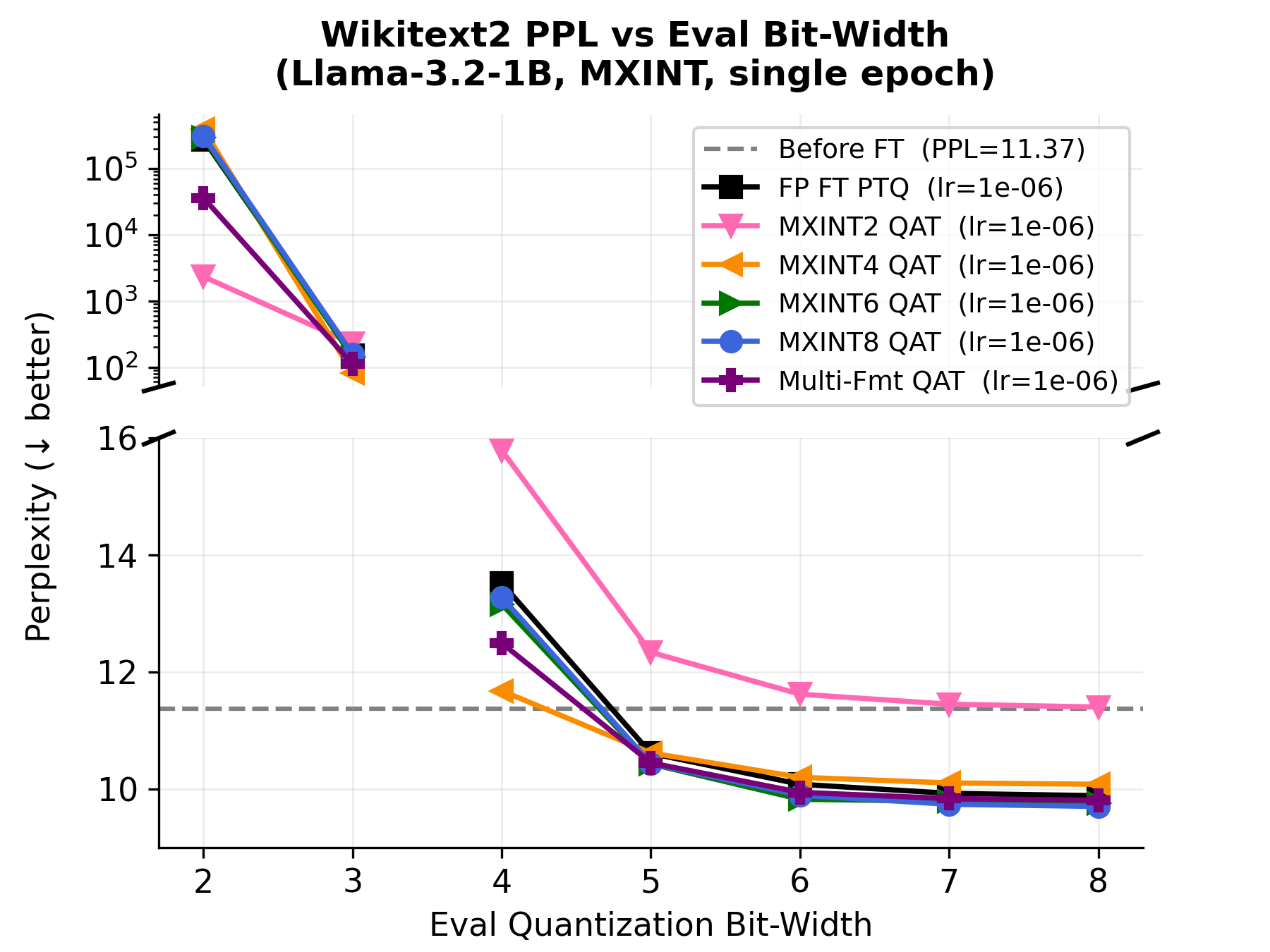}
  \hfill
  \includegraphics[width=0.49\textwidth]{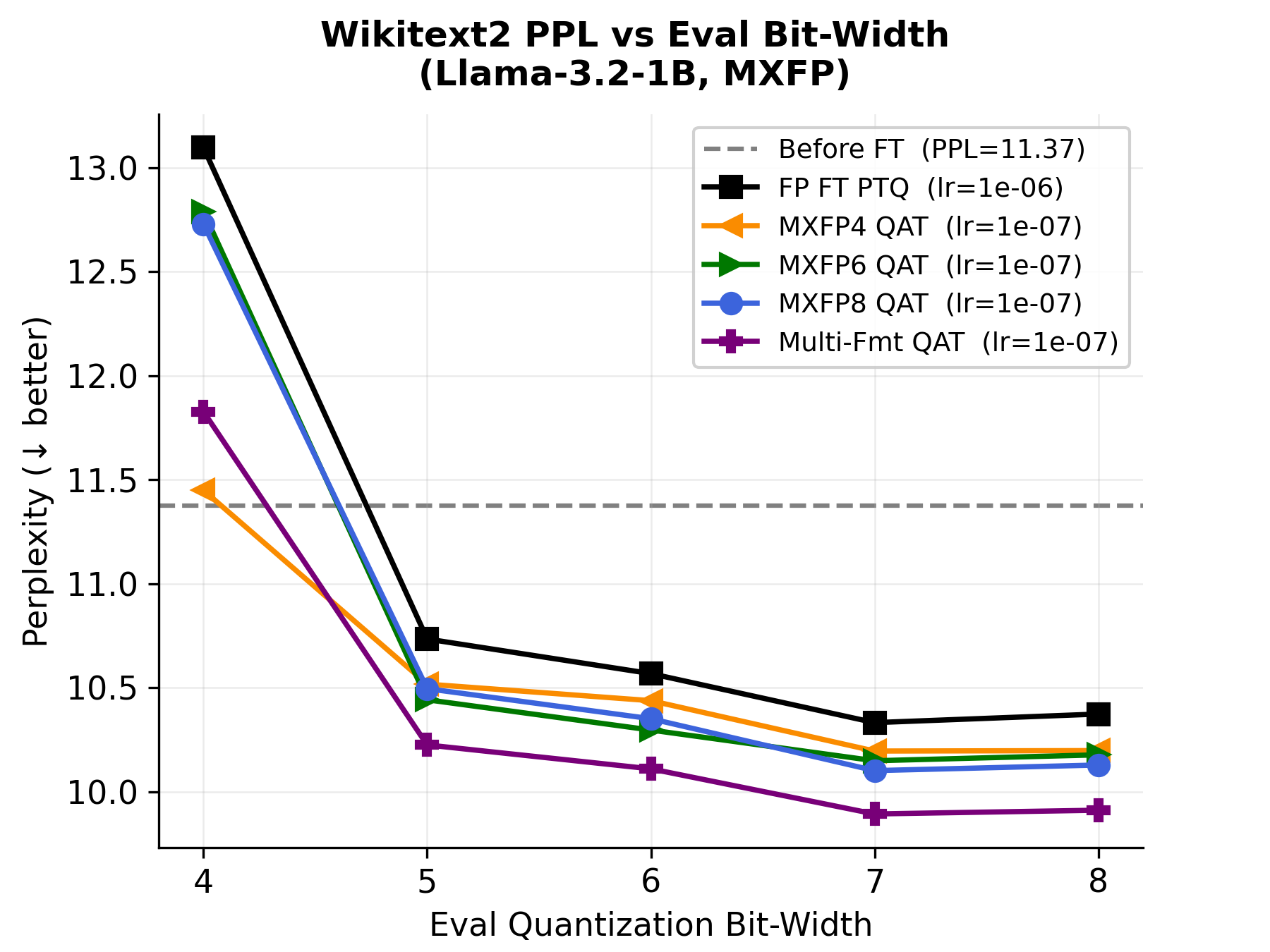}
  \caption{Multi-format QAT results for \texttt{Llama-3.2-1B}.}
  \label{fig:app_mfqat_llama32_1b}
\end{figure*}

\begin{figure*}[h]
  \centering
  \includegraphics[width=0.49\textwidth]{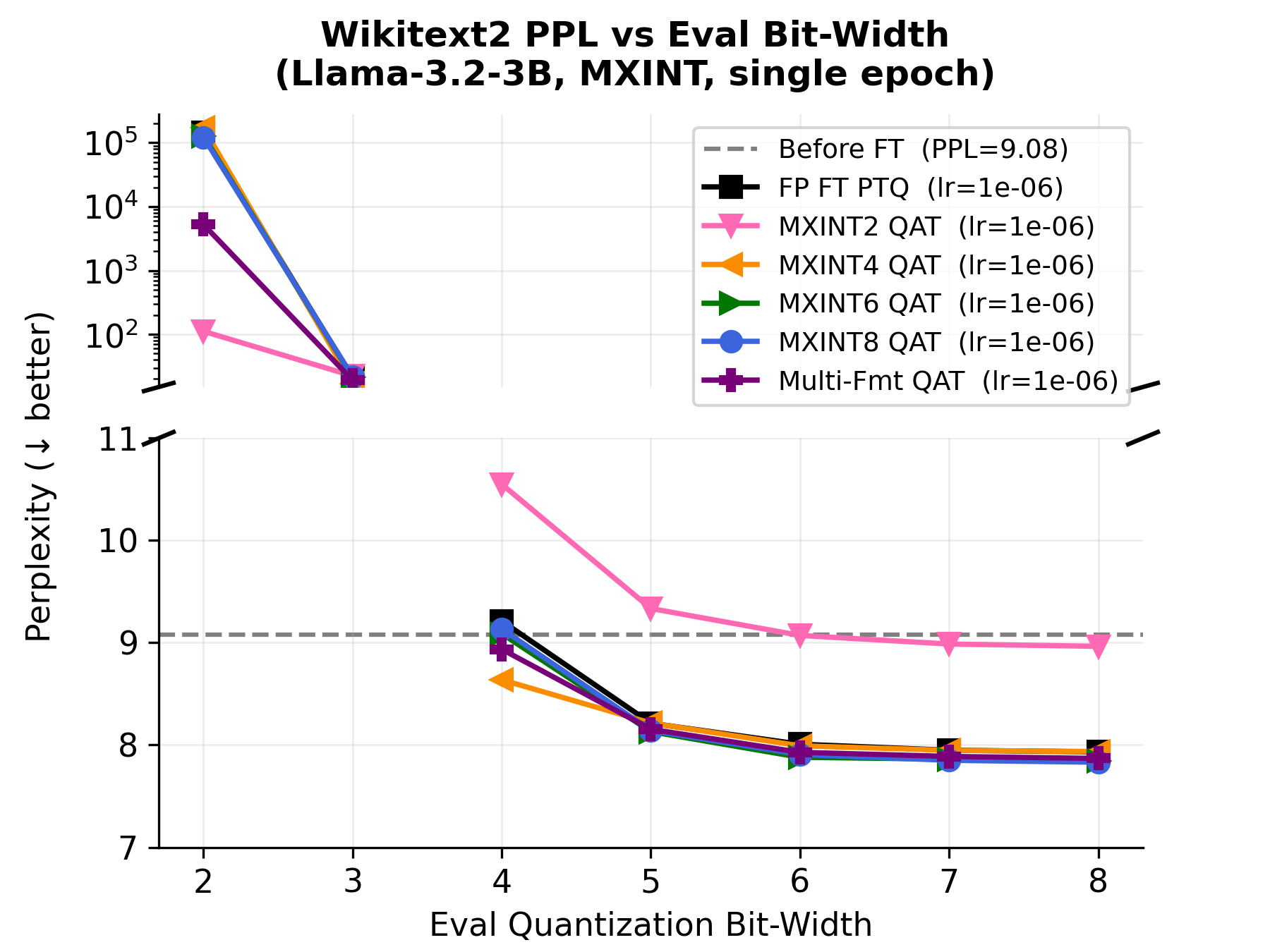}
  \hfill
  \includegraphics[width=0.49\textwidth]{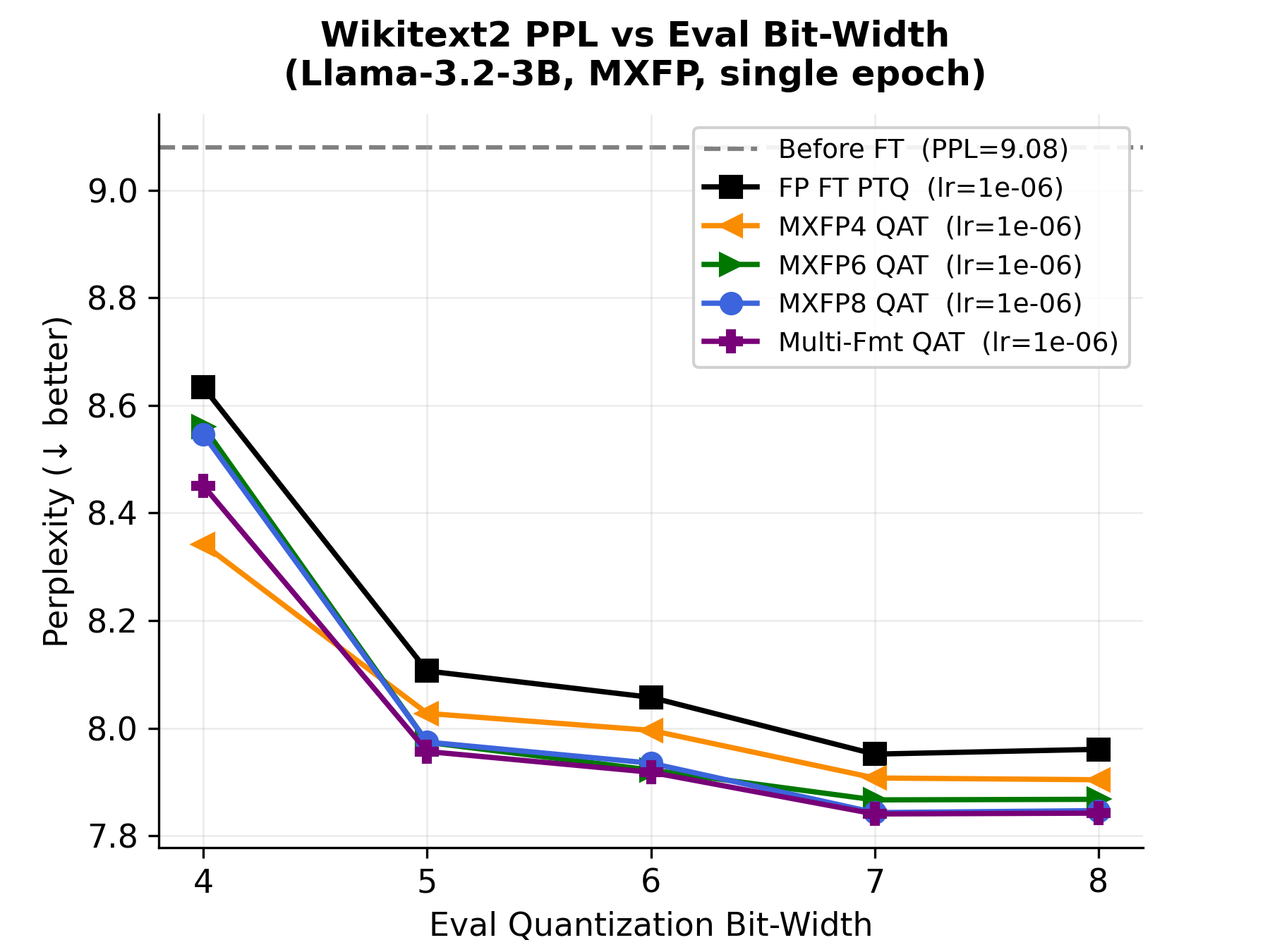}
  \caption{Multi-format QAT results for \texttt{Llama-3.2-3B}.}
  \label{fig:app_mfqat_llama32_3b}
\end{figure*}

\begin{figure*}[h]
  \centering
  \includegraphics[width=0.49\textwidth]{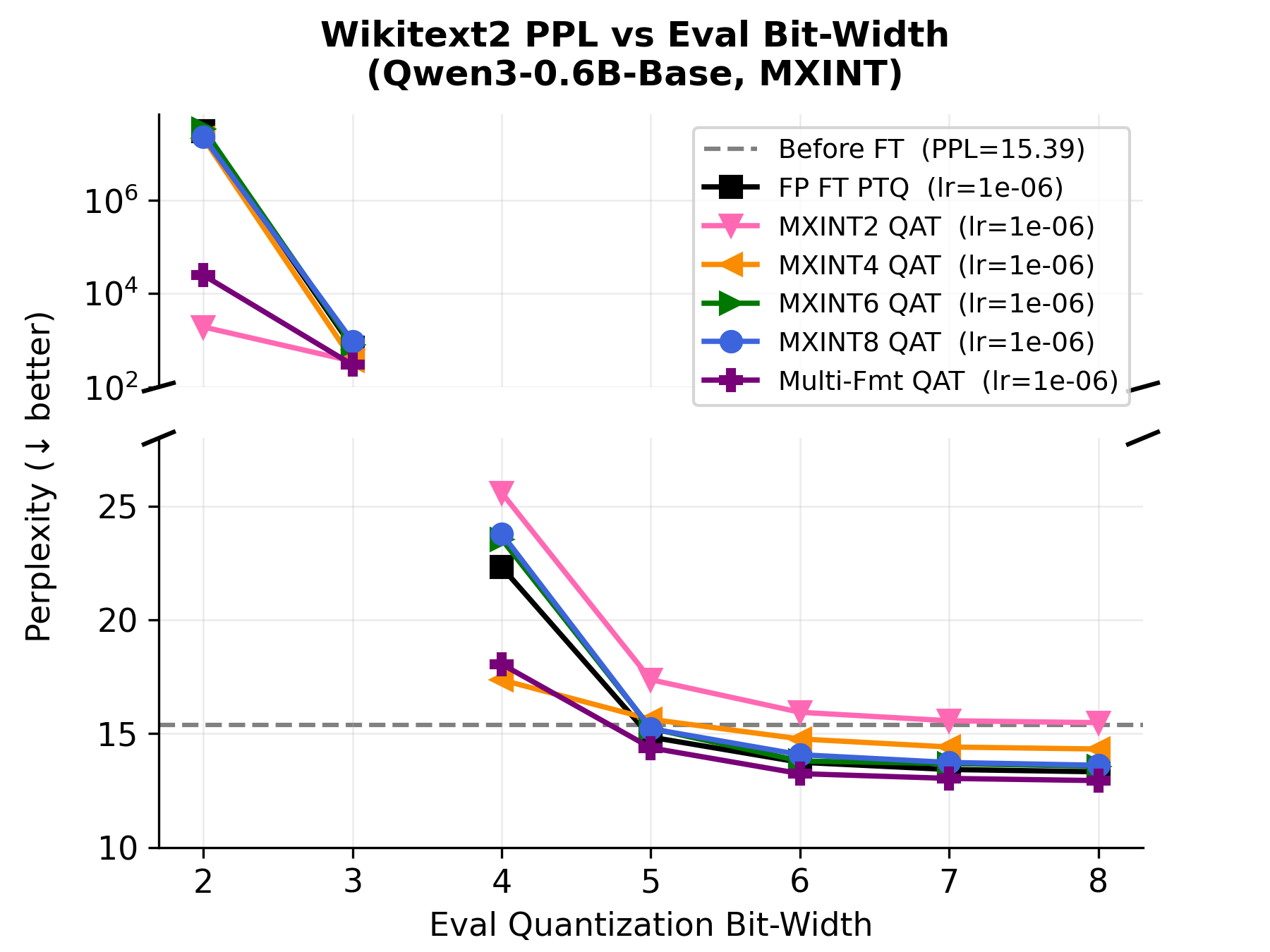}
  \hfill
  \includegraphics[width=0.49\textwidth]{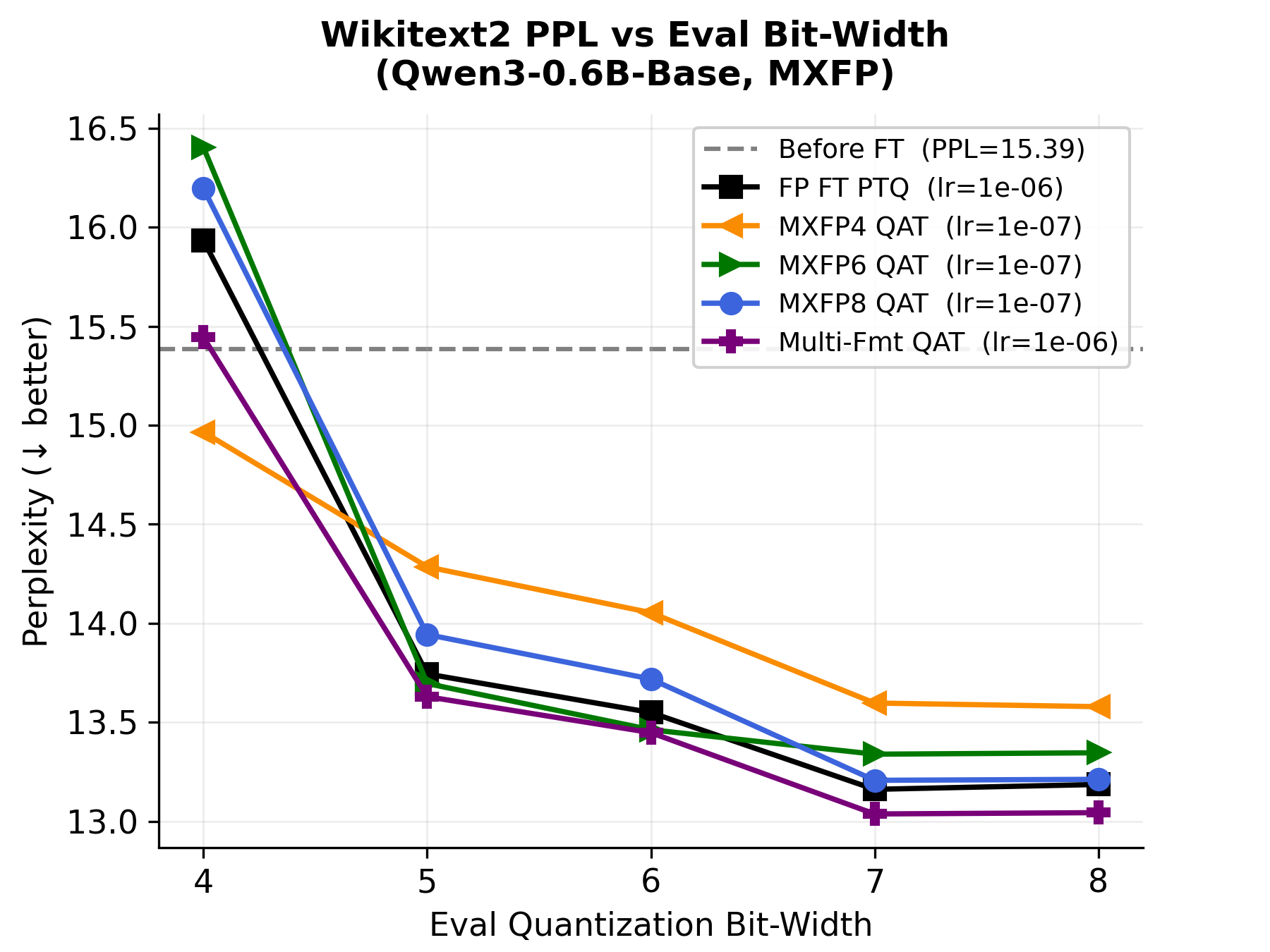}
  \caption{Multi-format QAT results for \texttt{Qwen3-0.6B-Base}.}
  \label{fig:app_mfqat_qwen3_06b}
\end{figure*}

\begin{figure*}[t]
  \centering
  \includegraphics[width=0.49\textwidth]{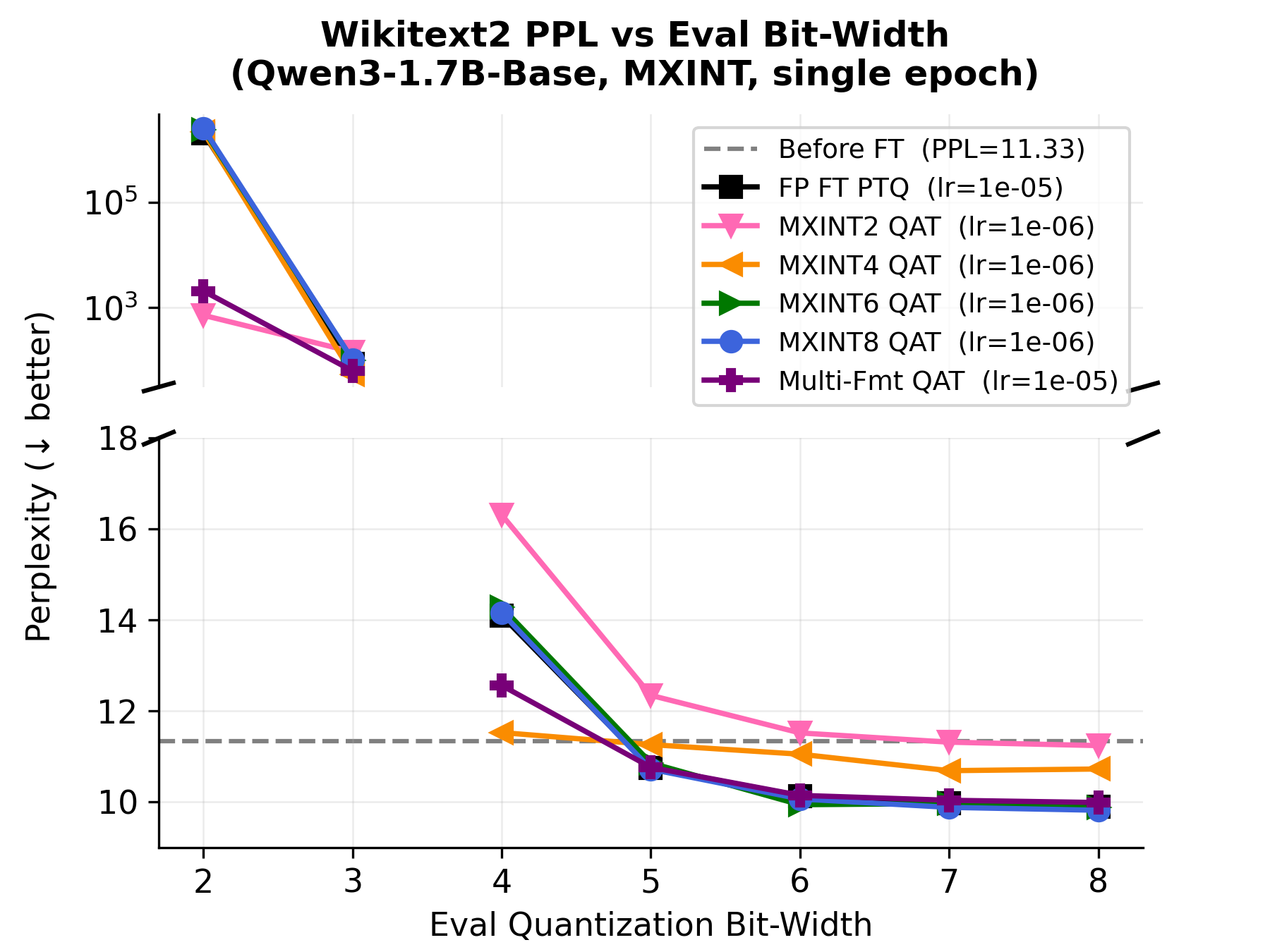}
  \hfill
  \includegraphics[width=0.49\textwidth]{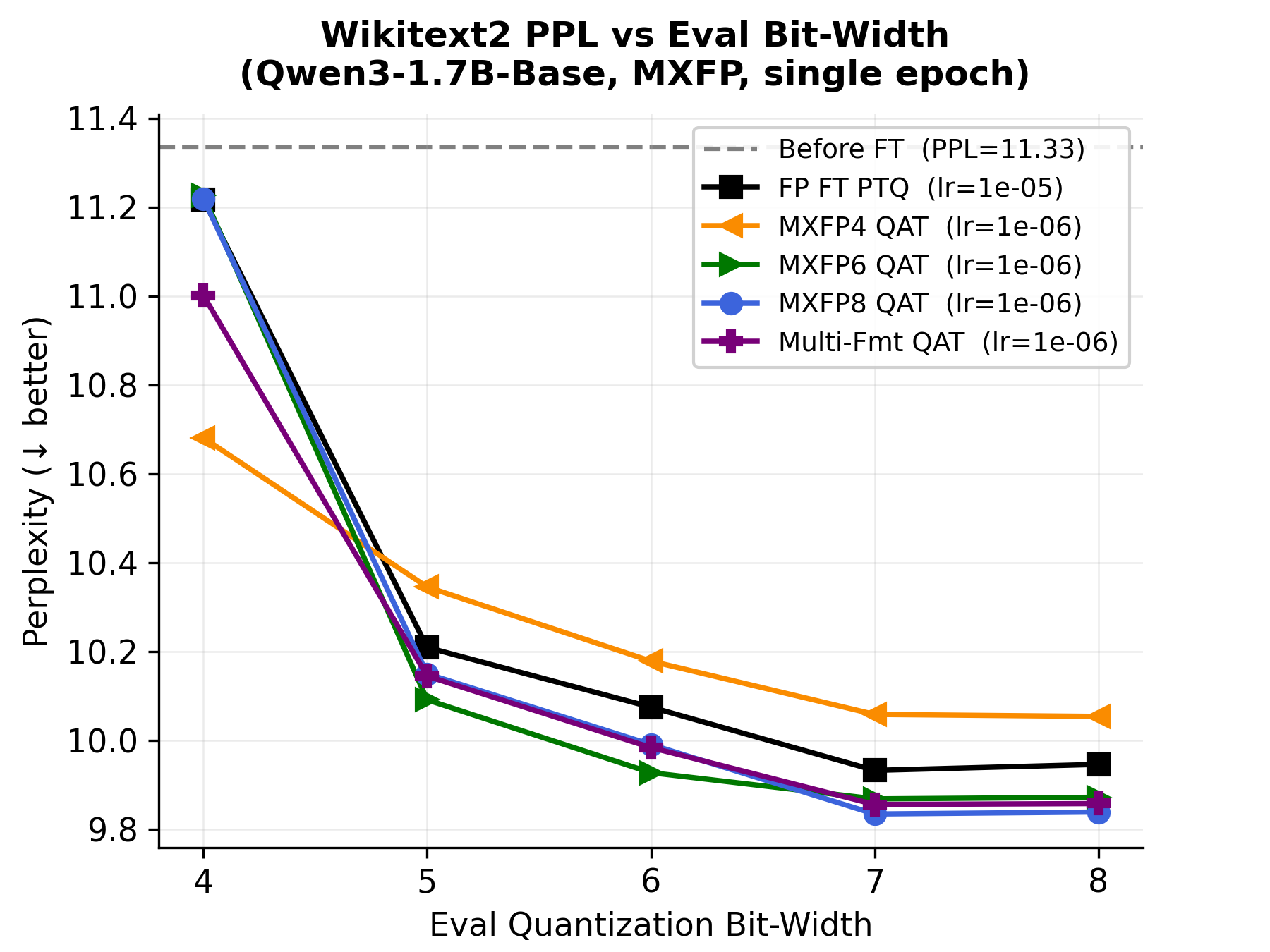}
  \caption{Multi-format QAT results for \texttt{Qwen3-1.7B-Base}.}
  \label{fig:app_mfqat_qwen3_17b}
\end{figure*}

\begin{figure*}[t]
  \centering
  \includegraphics[width=0.49\textwidth]{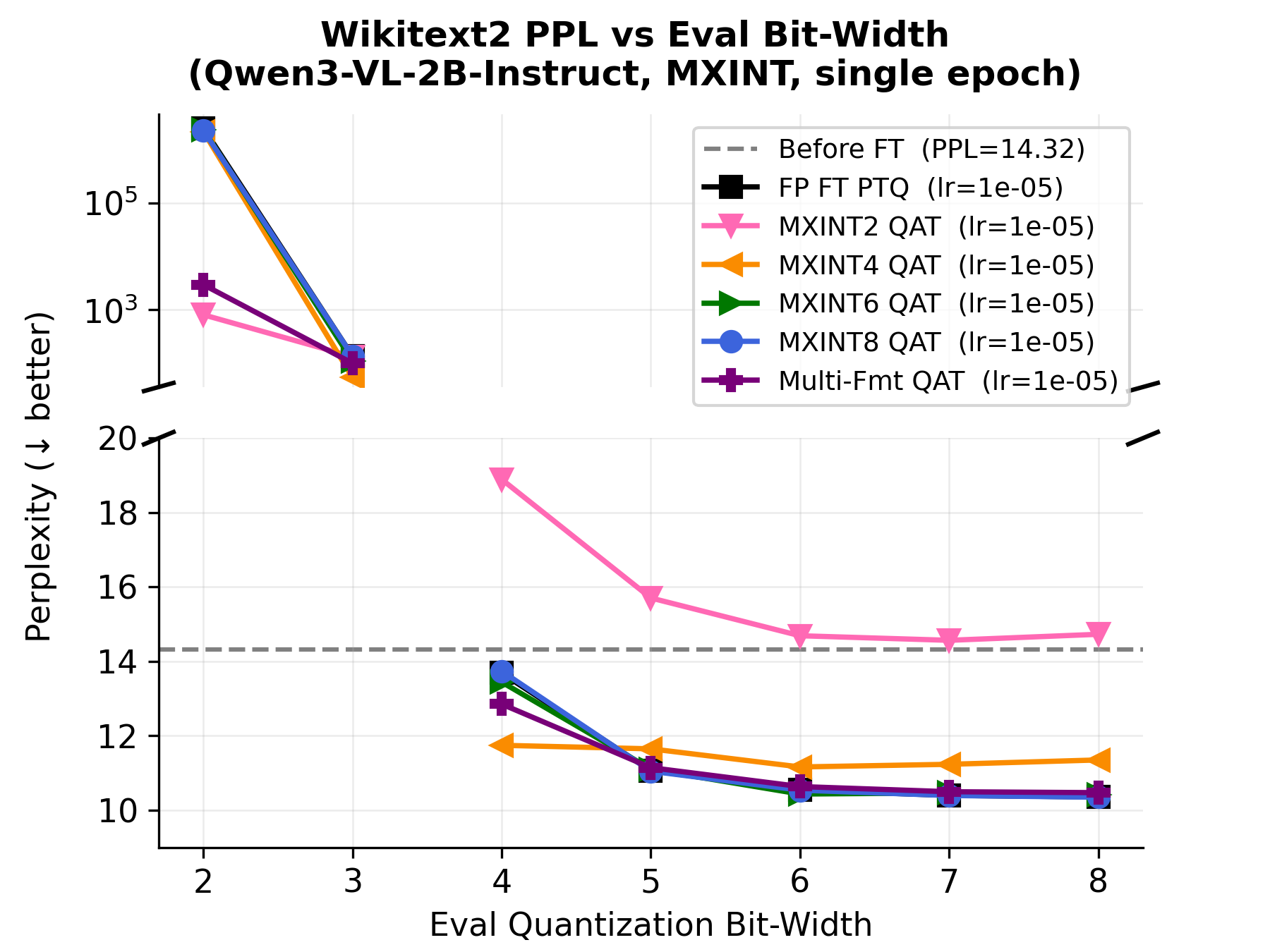}
  \hfill
  \includegraphics[width=0.49\textwidth]{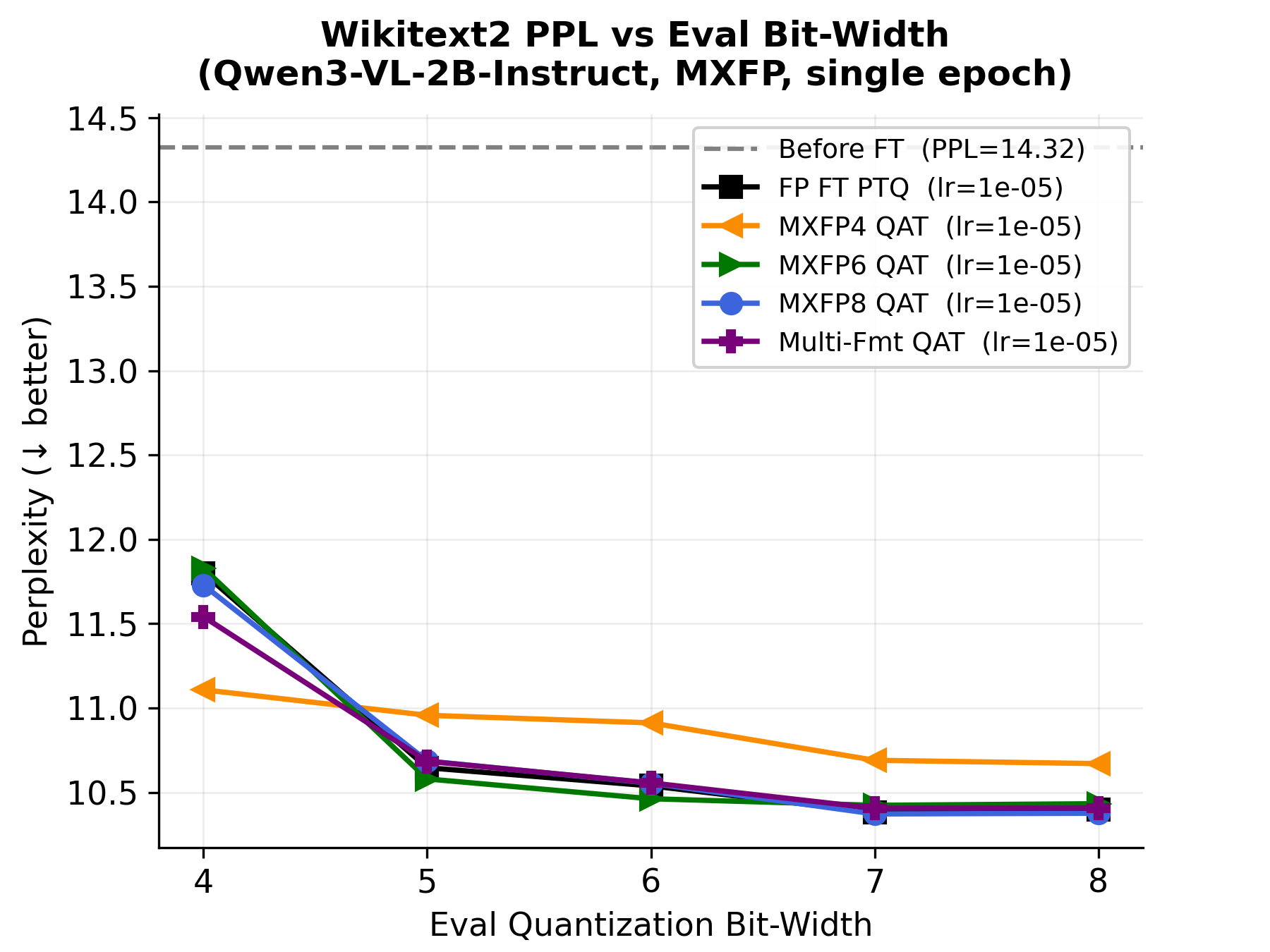}
  \caption{Multi-format QAT results for
  \texttt{Qwen3-VL-2B-Instruct}.}
  \label{fig:app_mfqat_qwen3vl_2b}
\end{figure*}

\begin{figure*}[t]
  \centering
  \includegraphics[width=0.49\textwidth]{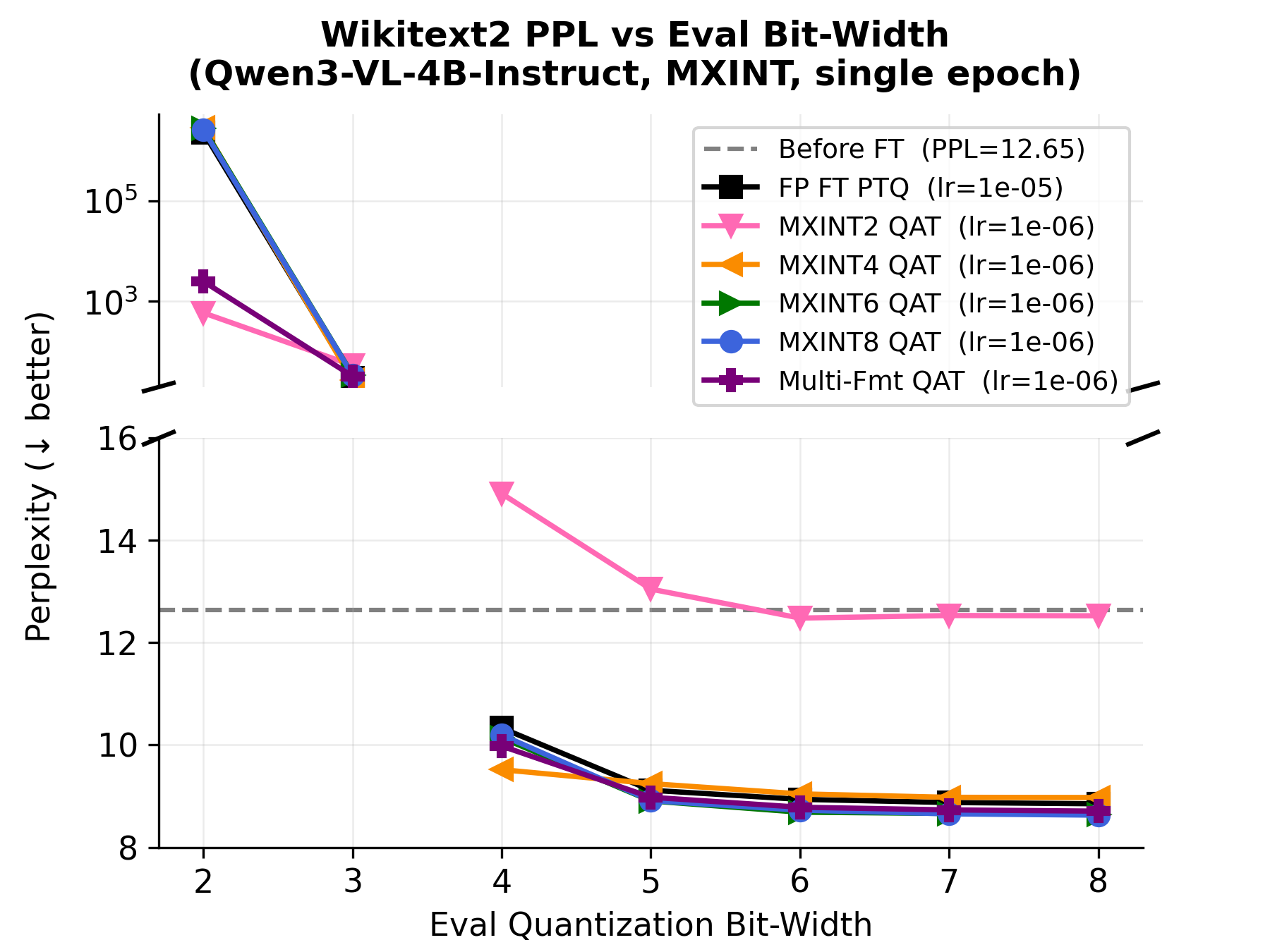}
  \hfill
  \includegraphics[width=0.49\textwidth]{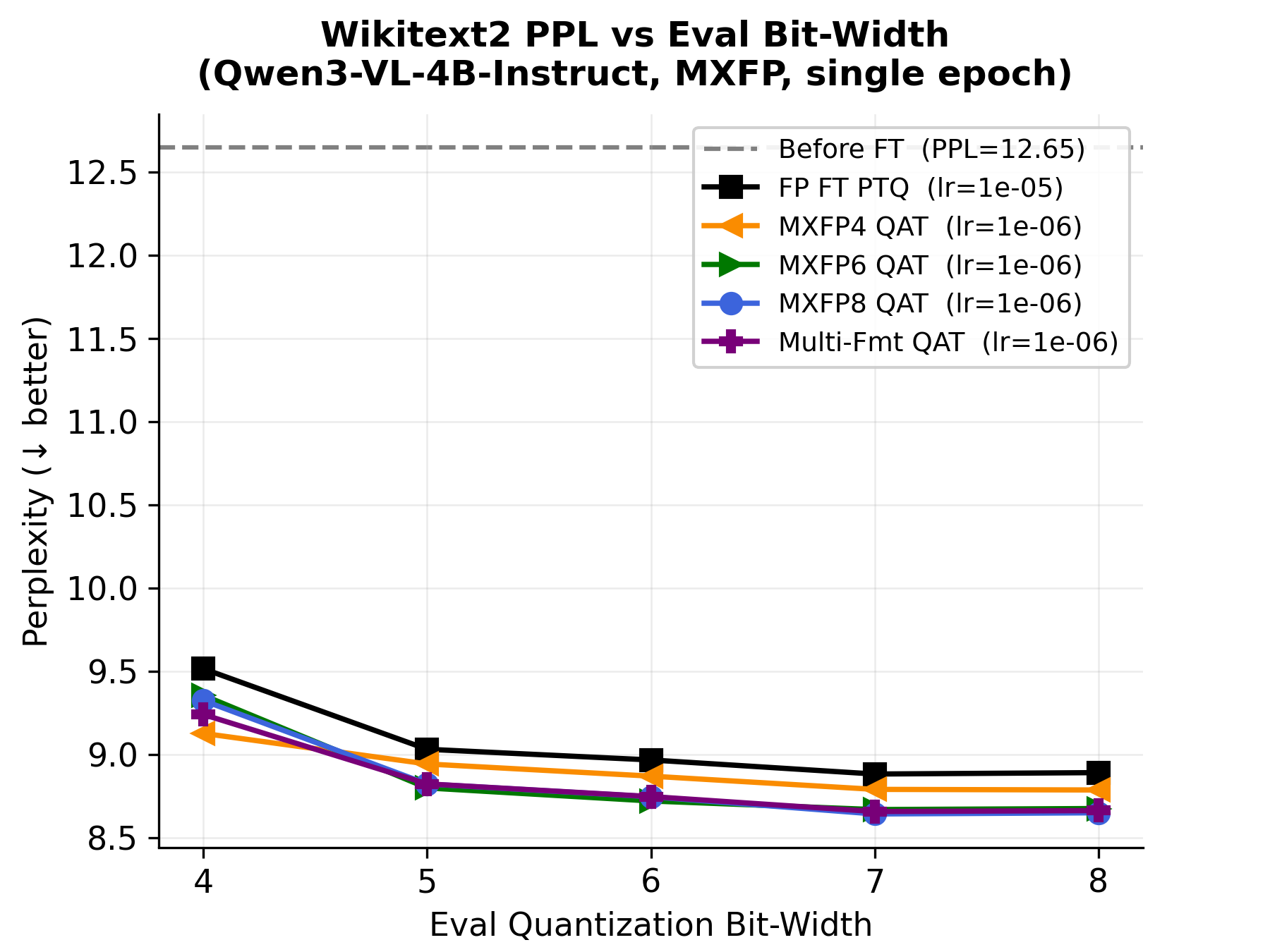}
  \caption{Multi-format QAT results for
  \texttt{Qwen3-VL-4B-Instruct}.}
  \label{fig:app_mfqat_qwen3vl_4b}
\end{figure*}

\begin{figure*}[t]
  \centering
  \includegraphics[width=0.49\textwidth]{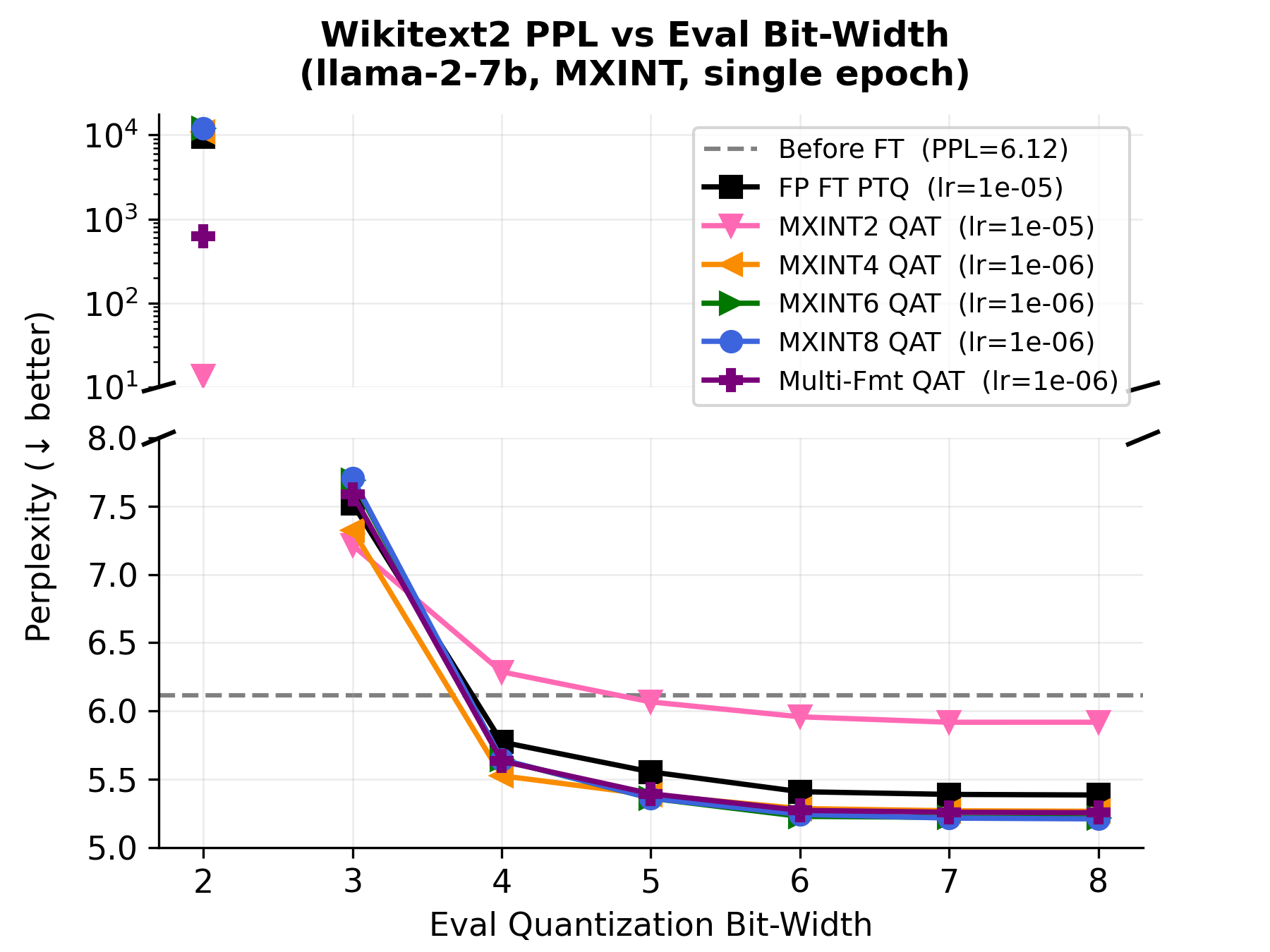}
  \hfill
  \includegraphics[width=0.49\textwidth]{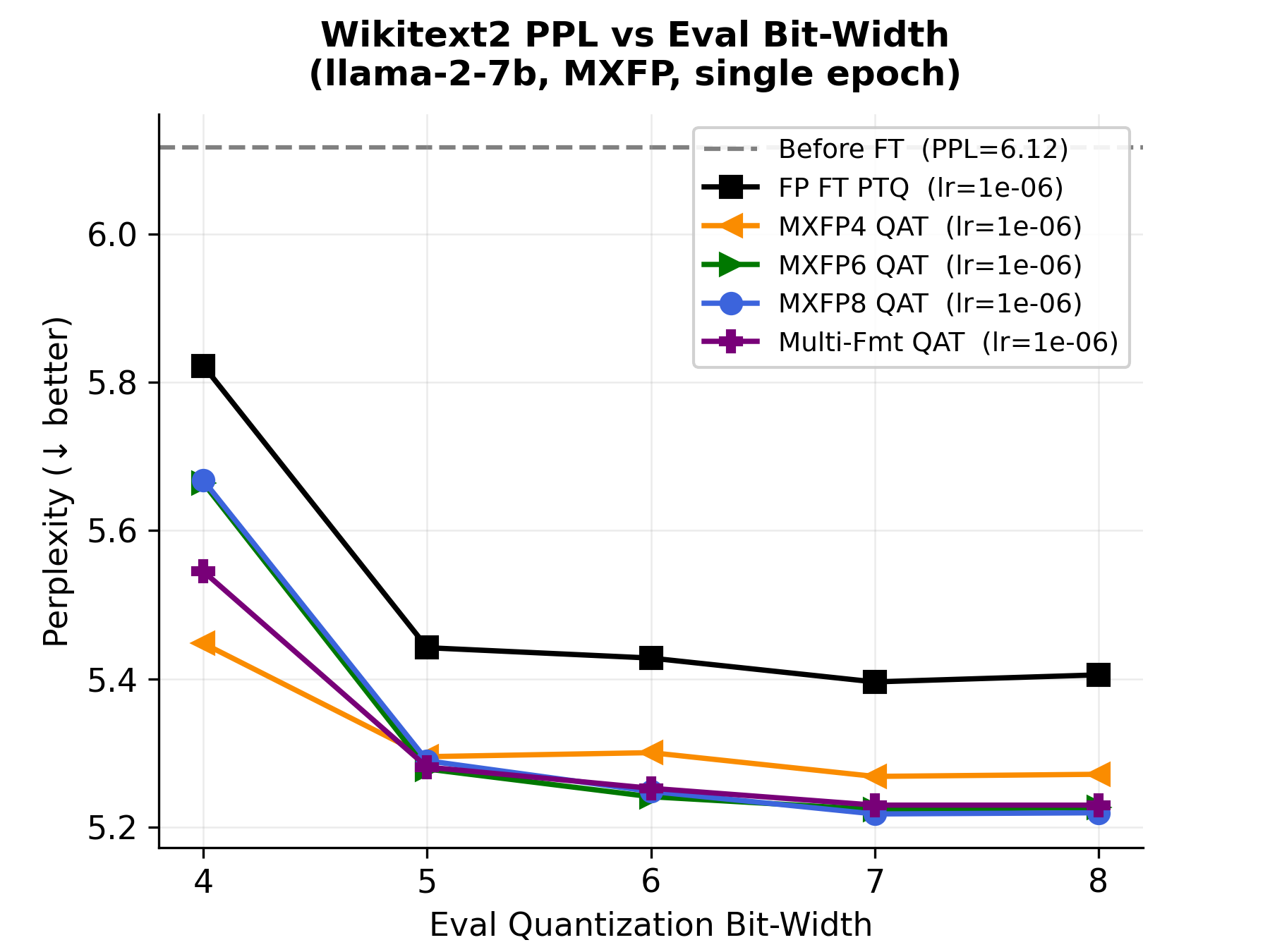}
  \caption{Multi-format QAT results for \texttt{llama-2-7b}.}
  \label{fig:app_mfqat_llama2_7b}
\end{figure*}

\subsection{Additional plots for Multi-format QAT with Slice and Scale}
\label{app:extra_multi_format_ss}
This appendix contains plots for
Section~\ref{sec:results_multiformat_ss} on all models. Each figure below contains results for one model with left figure for MXINT and right figure for MXFP. Plotting conventions follow Figure \ref{fig:mfqat_ss_main}.

\begin{figure*}[t]
  \centering
  \includegraphics[width=0.49\textwidth]{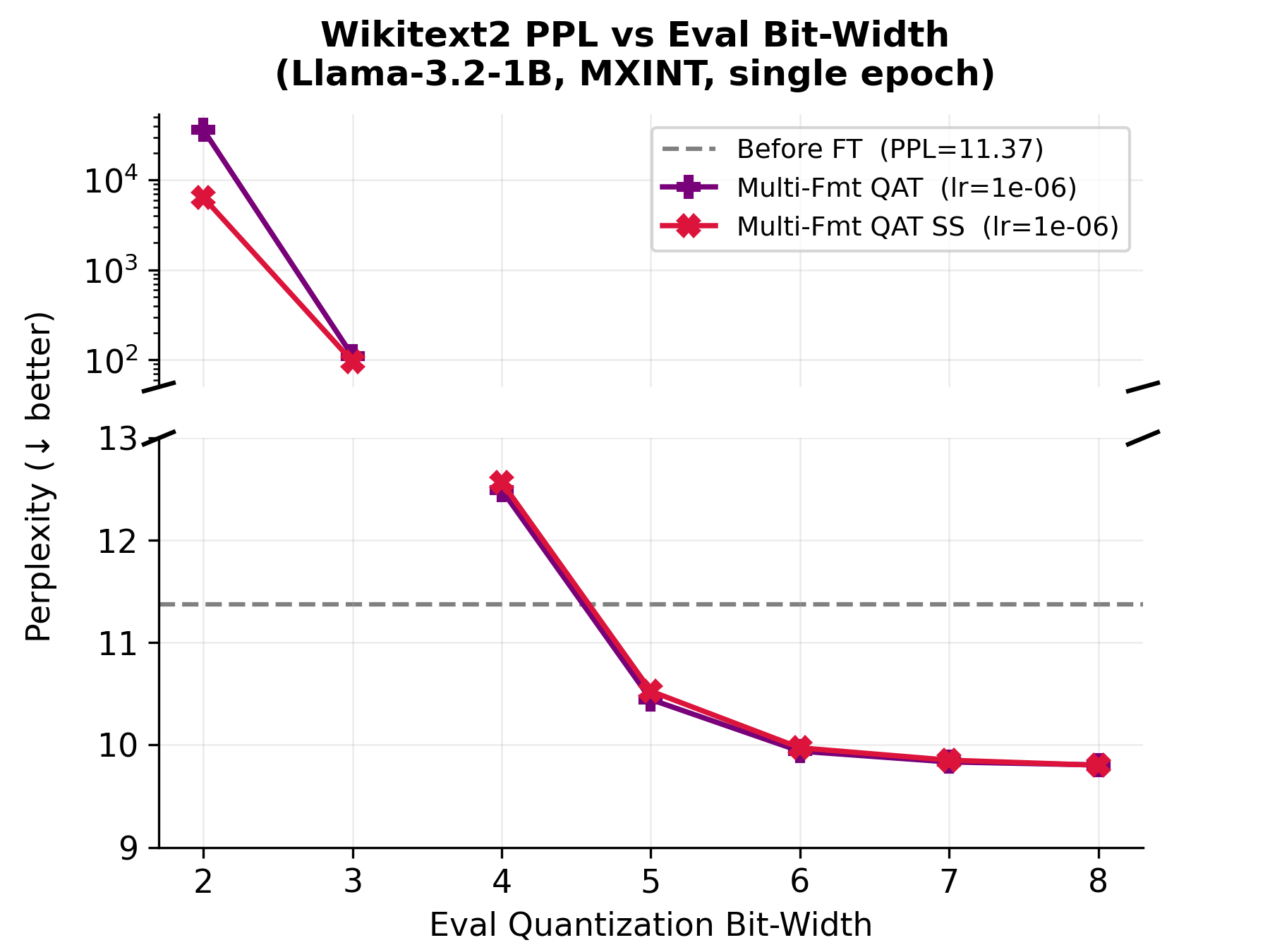}
  \hfill
  \includegraphics[width=0.49\textwidth]{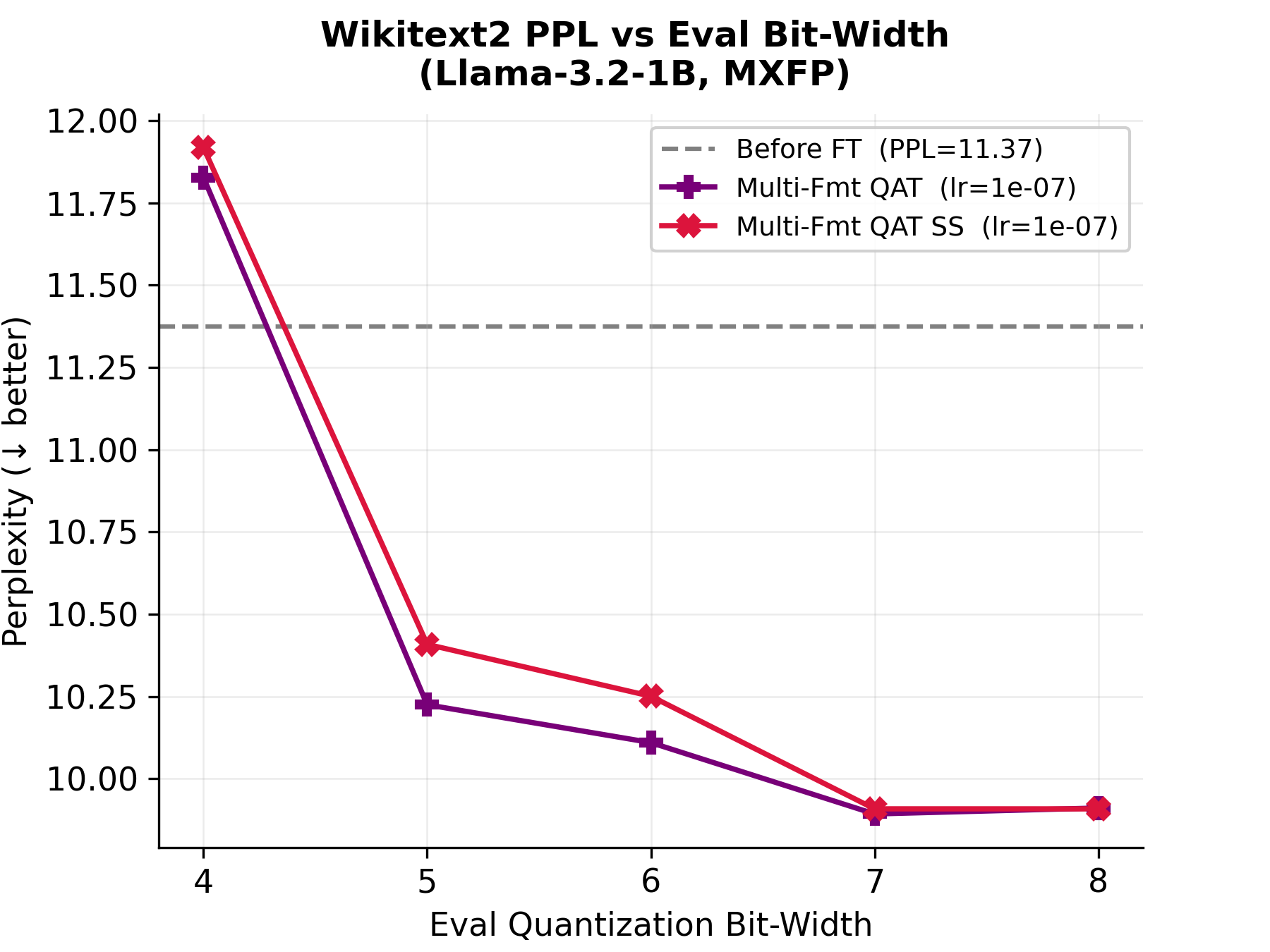}
  \caption{Multi-format QAT with Slice-and-Scale results for
  \texttt{Llama-3.2-1B}.}
  \label{fig:app_mfss_llama32_1b}
\end{figure*}

\begin{figure*}[t]
  \centering
  \includegraphics[width=0.49\textwidth]{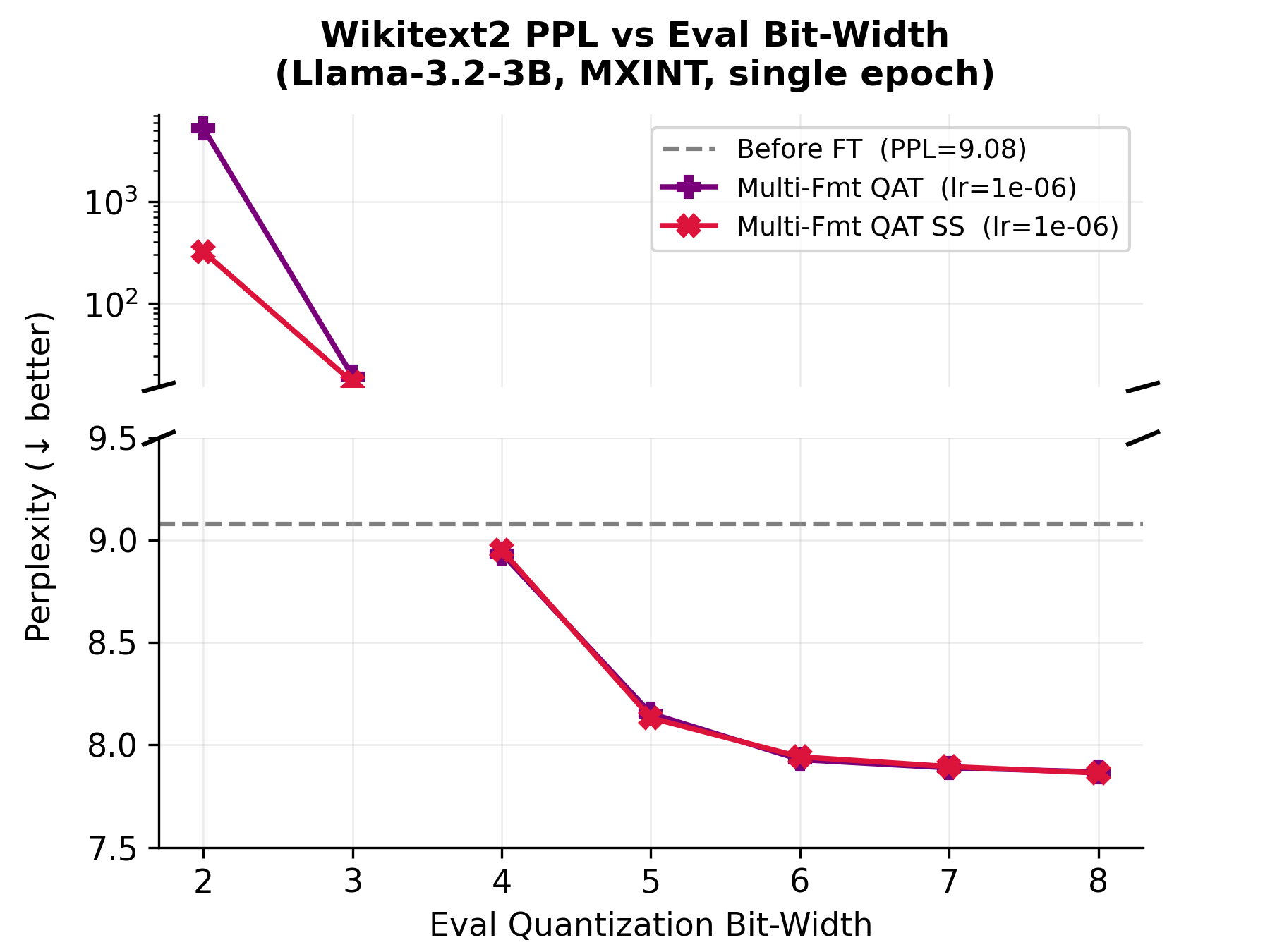}
  \hfill
  \includegraphics[width=0.49\textwidth]{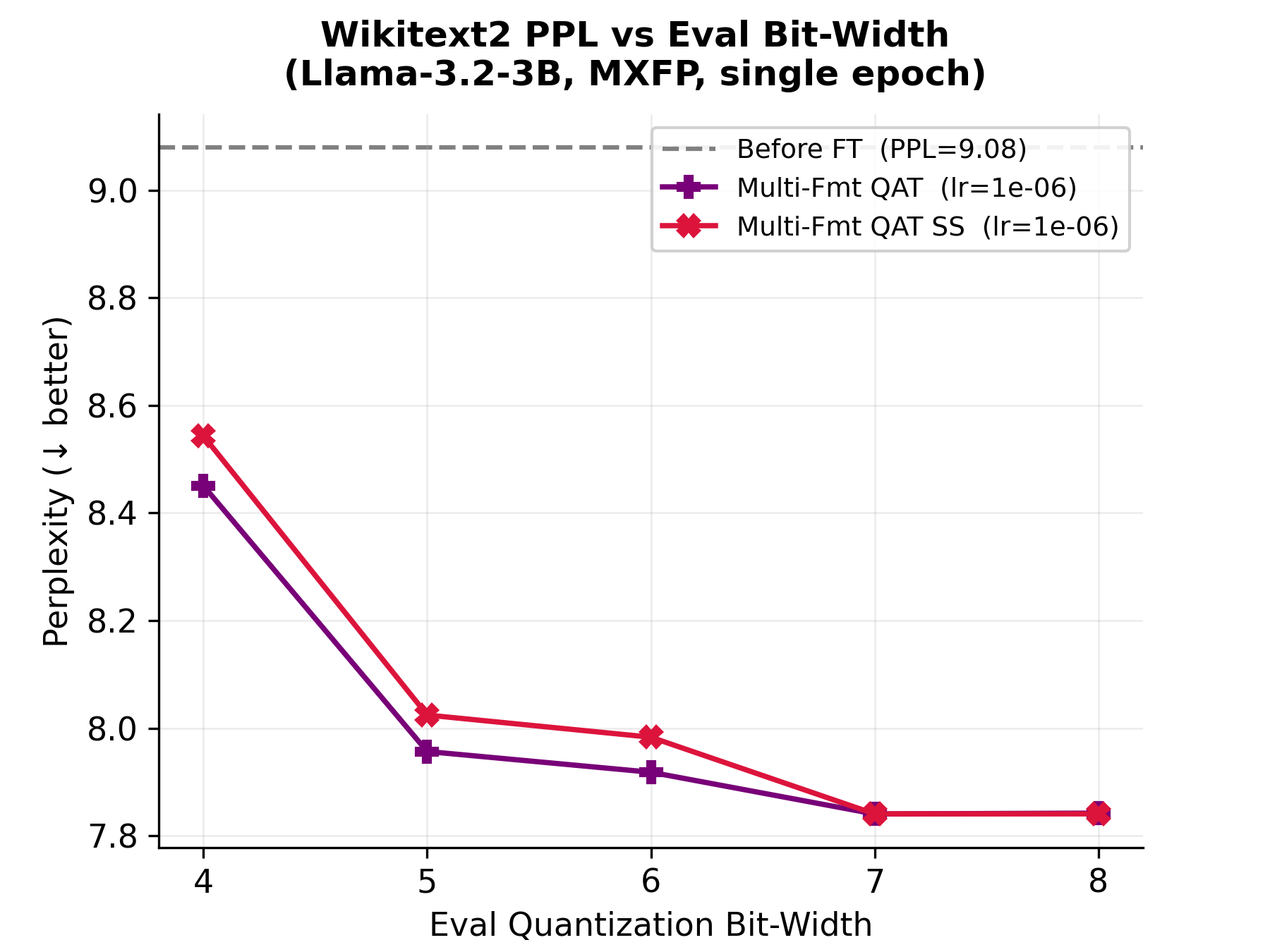}
  \caption{Multi-format QAT with Slice-and-Scale results for
  \texttt{Llama-3.2-3B}.}
  \label{fig:app_mfss_llama32_3b}
\end{figure*}

\begin{figure*}[t]
  \centering
  \includegraphics[width=0.49\textwidth]{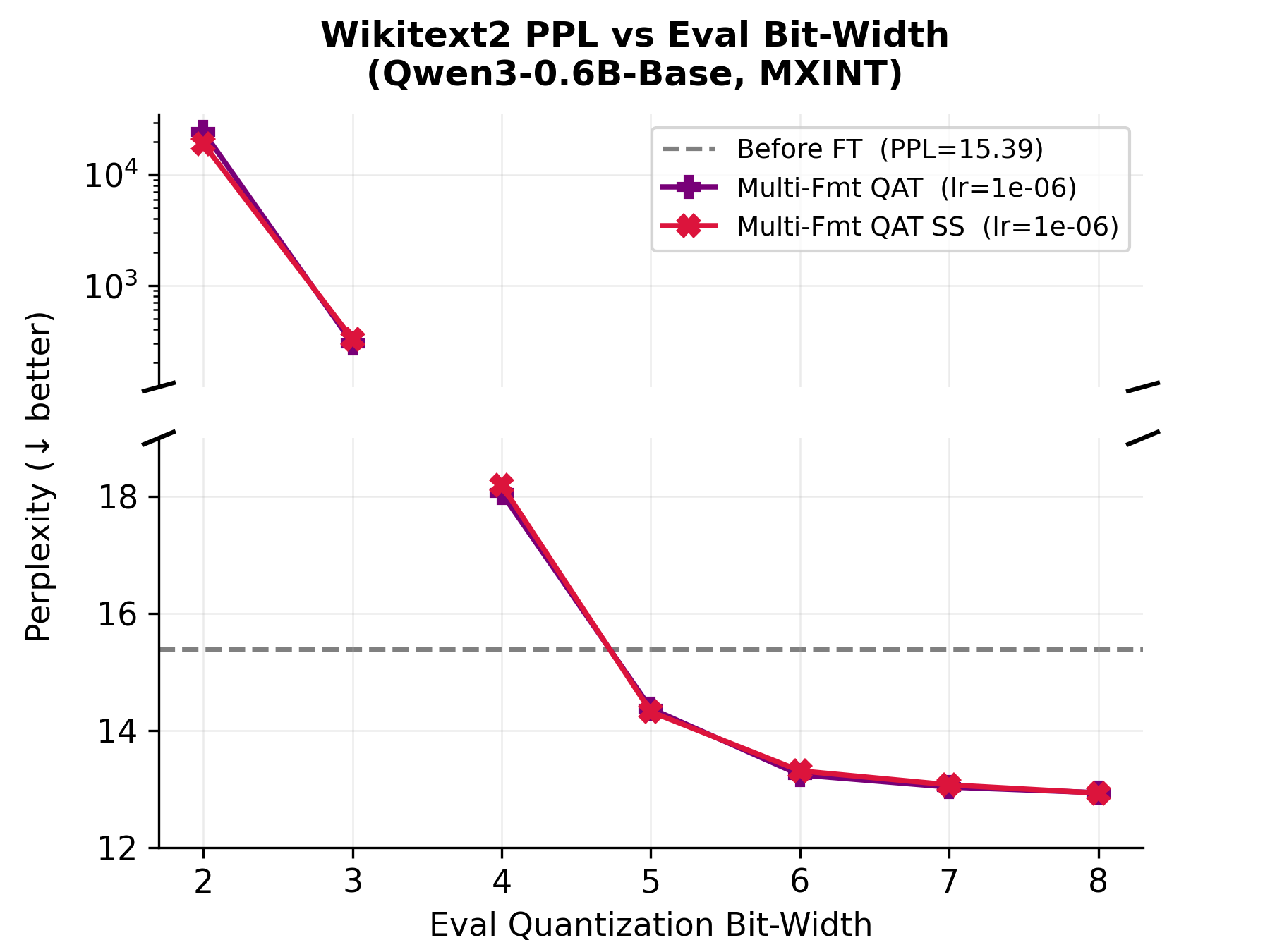}
  \hfill
  \includegraphics[width=0.49\textwidth]{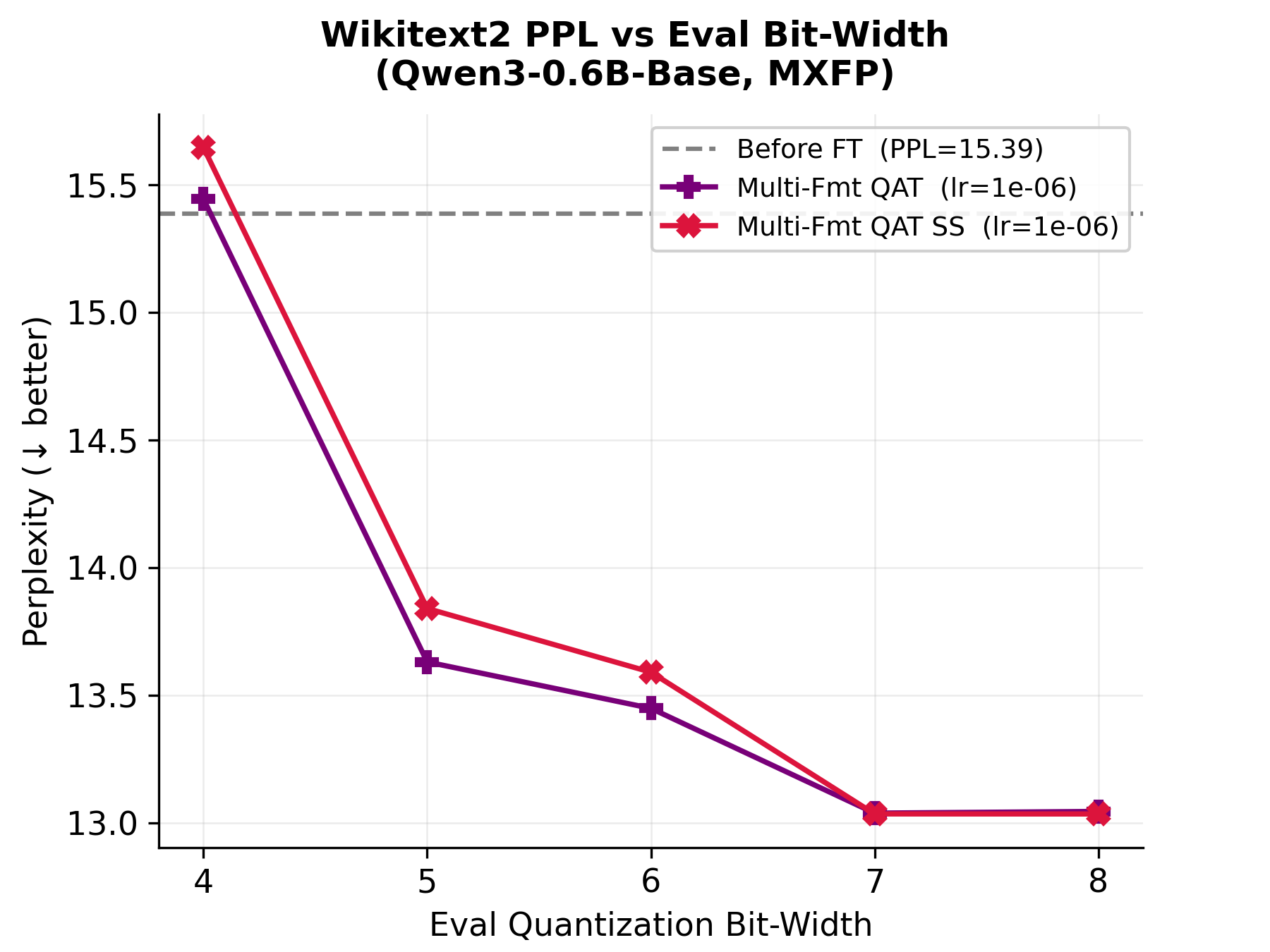}
  \caption{Multi-format QAT with Slice-and-Scale results for
  \texttt{Qwen3-0.6B-Base}.}
  \label{fig:app_mfss_qwen3_06b}
\end{figure*}

\begin{figure*}[t]
  \centering
  \includegraphics[width=0.49\textwidth]{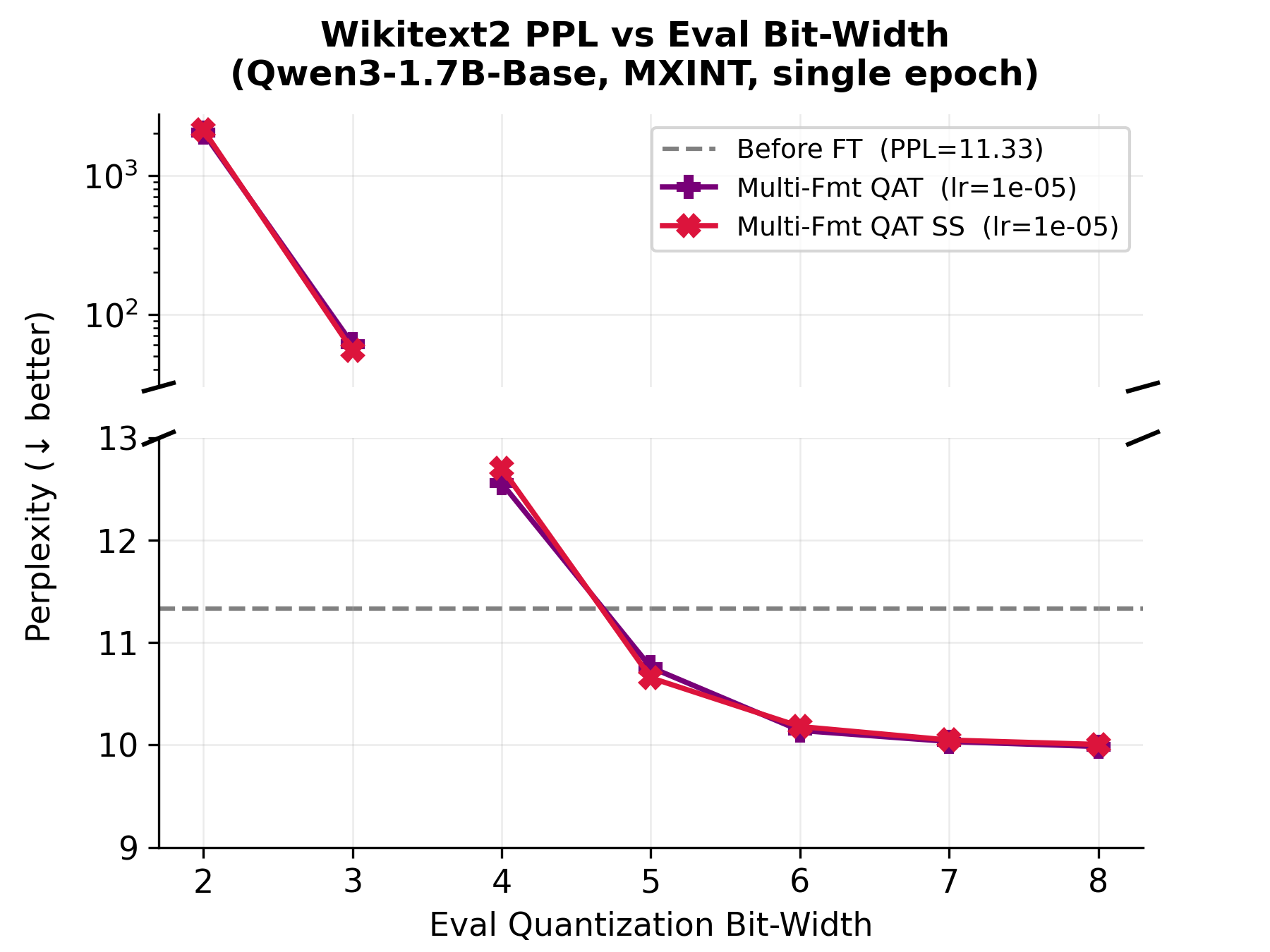}
  \hfill
  \includegraphics[width=0.49\textwidth]{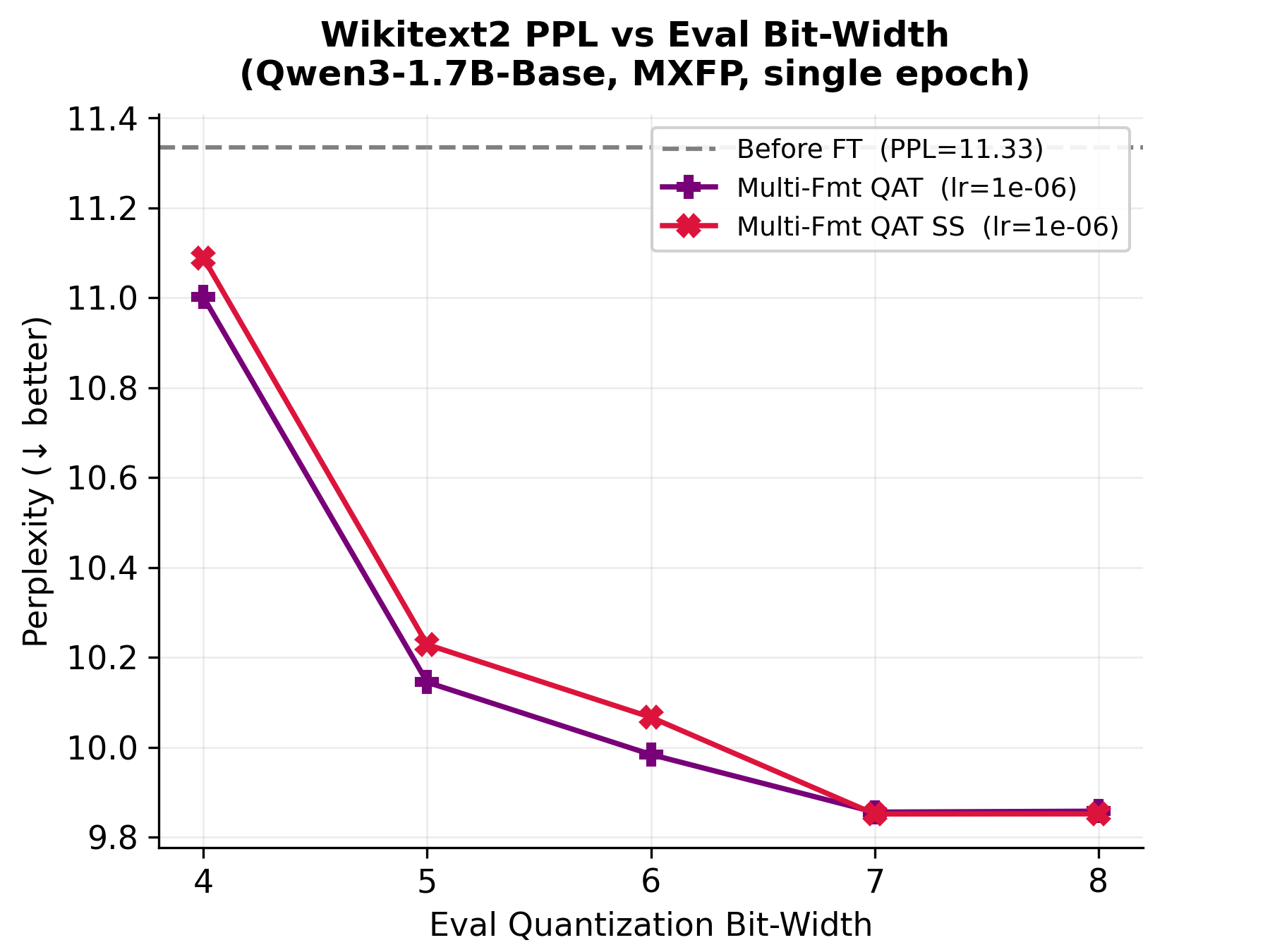}
  \caption{Multi-format QAT with Slice-and-Scale results for
  \texttt{Qwen3-1.7B-Base}.}
  \label{fig:app_mfss_qwen3_17b}
\end{figure*}

\begin{figure*}[t]
  \centering
  \includegraphics[width=0.49\textwidth]{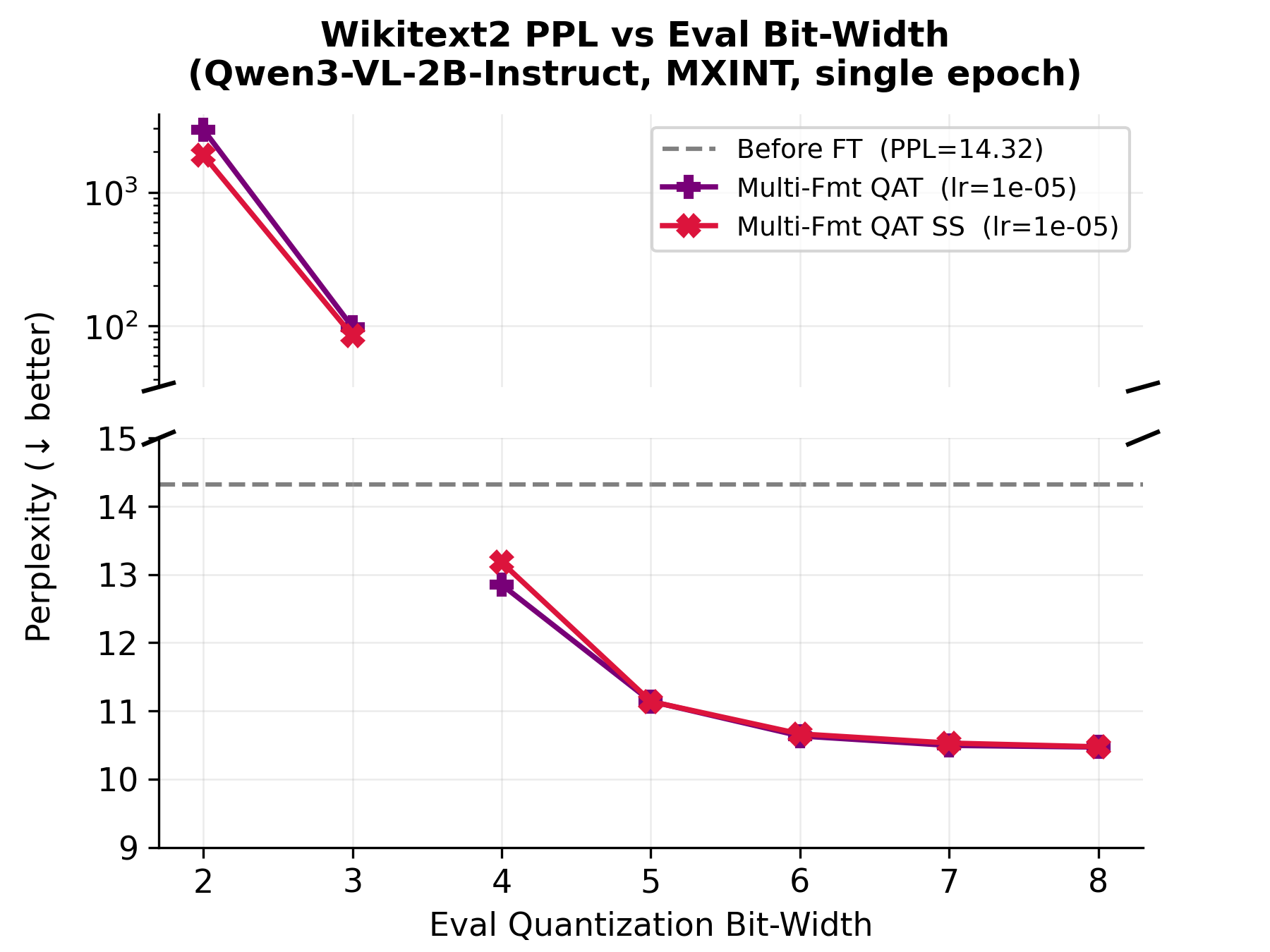}
  \hfill
  \includegraphics[width=0.49\textwidth]{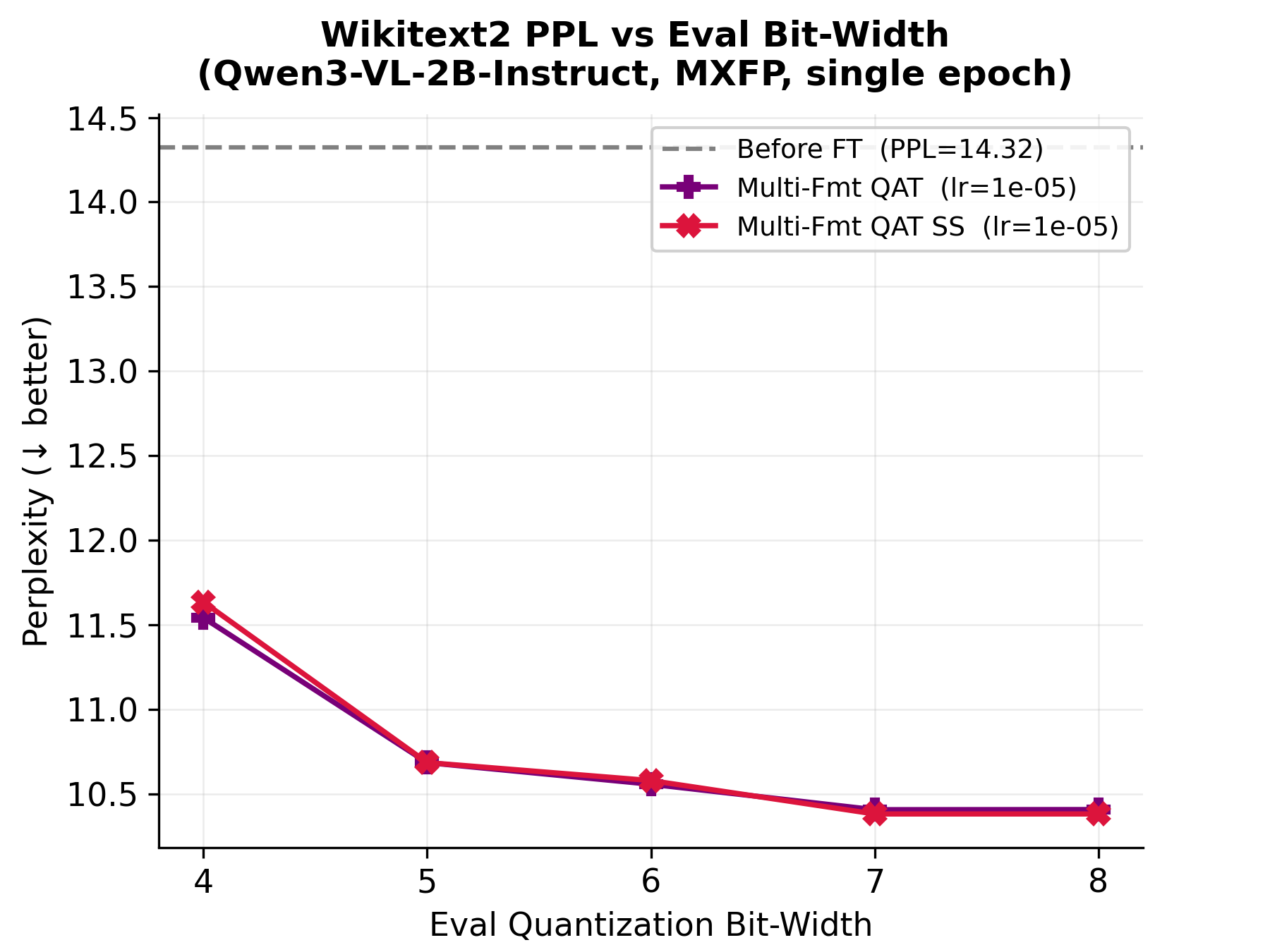}
  \caption{Multi-format QAT with Slice-and-Scale results for
  \texttt{Qwen3-VL-2B-Instruct}.}
  \label{fig:app_mfss_qwen3vl_2b}
\end{figure*}

\begin{figure*}[t]
  \centering
  \includegraphics[width=0.49\textwidth]{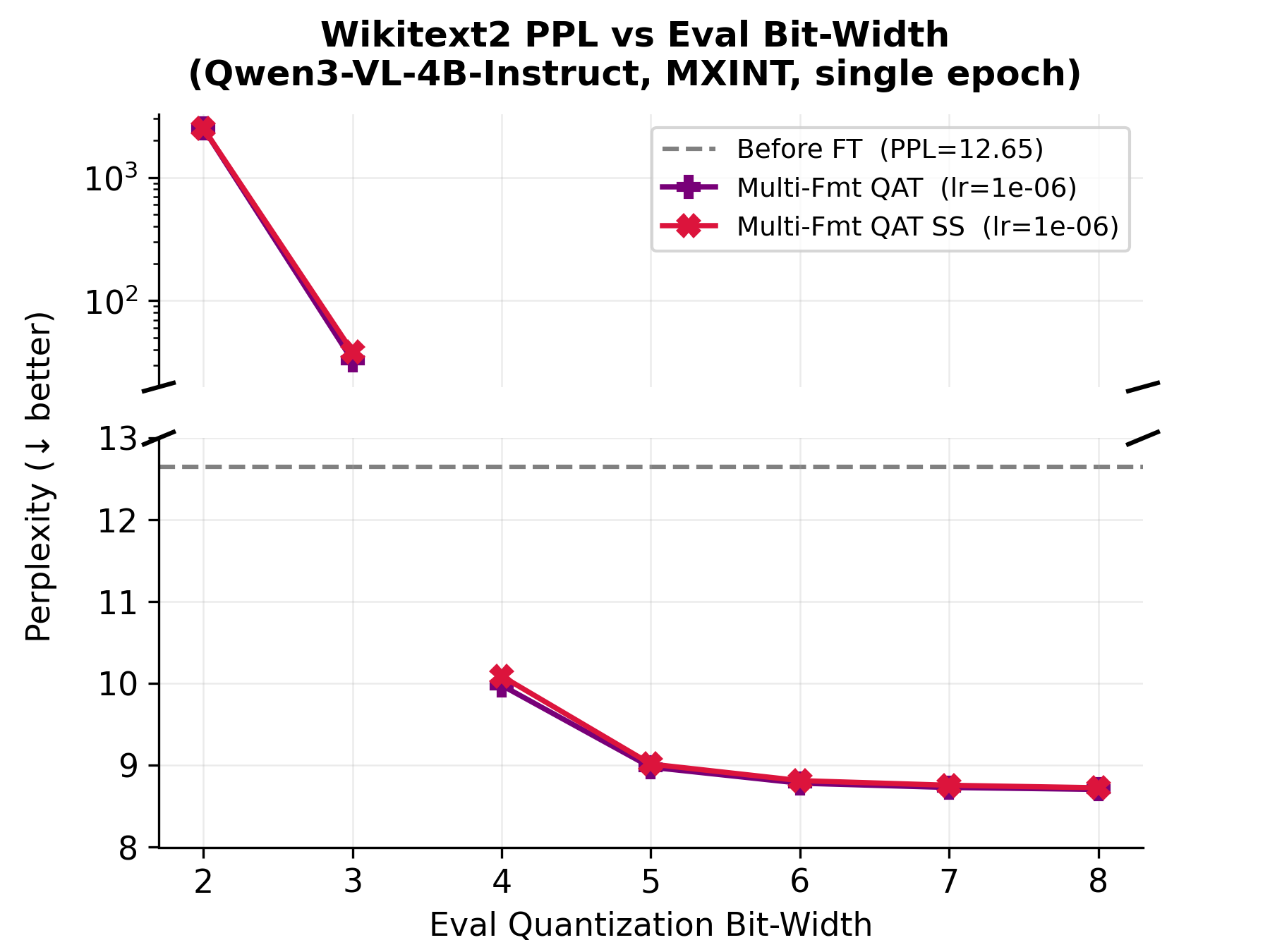}
  \hfill
  \includegraphics[width=0.49\textwidth]{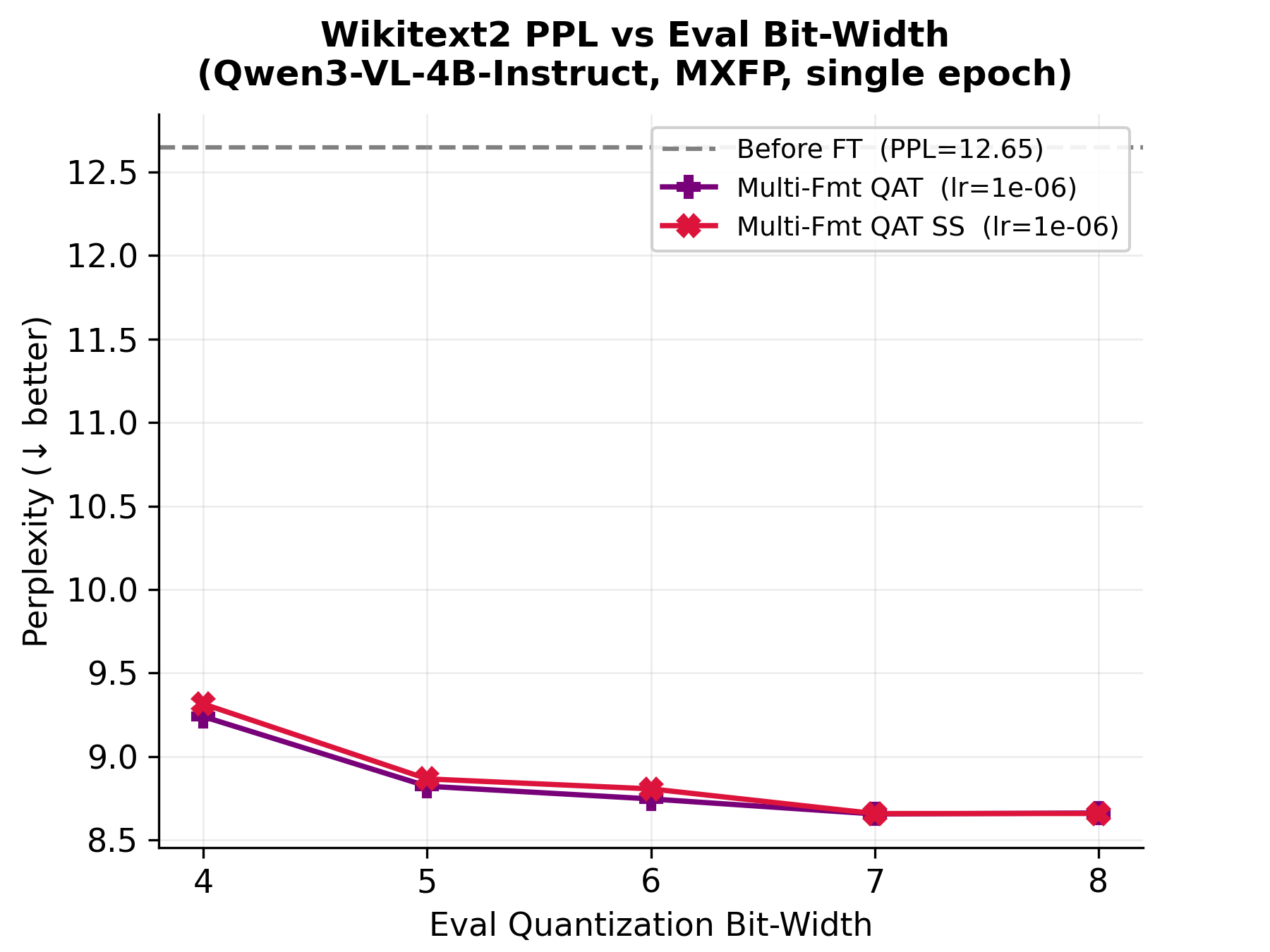}
  \caption{Multi-format QAT with Slice-and-Scale results for
  \texttt{Qwen3-VL-4B-Instruct}.}
  \label{fig:app_mfss_qwen3vl_4b}
\end{figure*}

\begin{figure*}[t]
  \centering
  \includegraphics[width=0.49\textwidth]{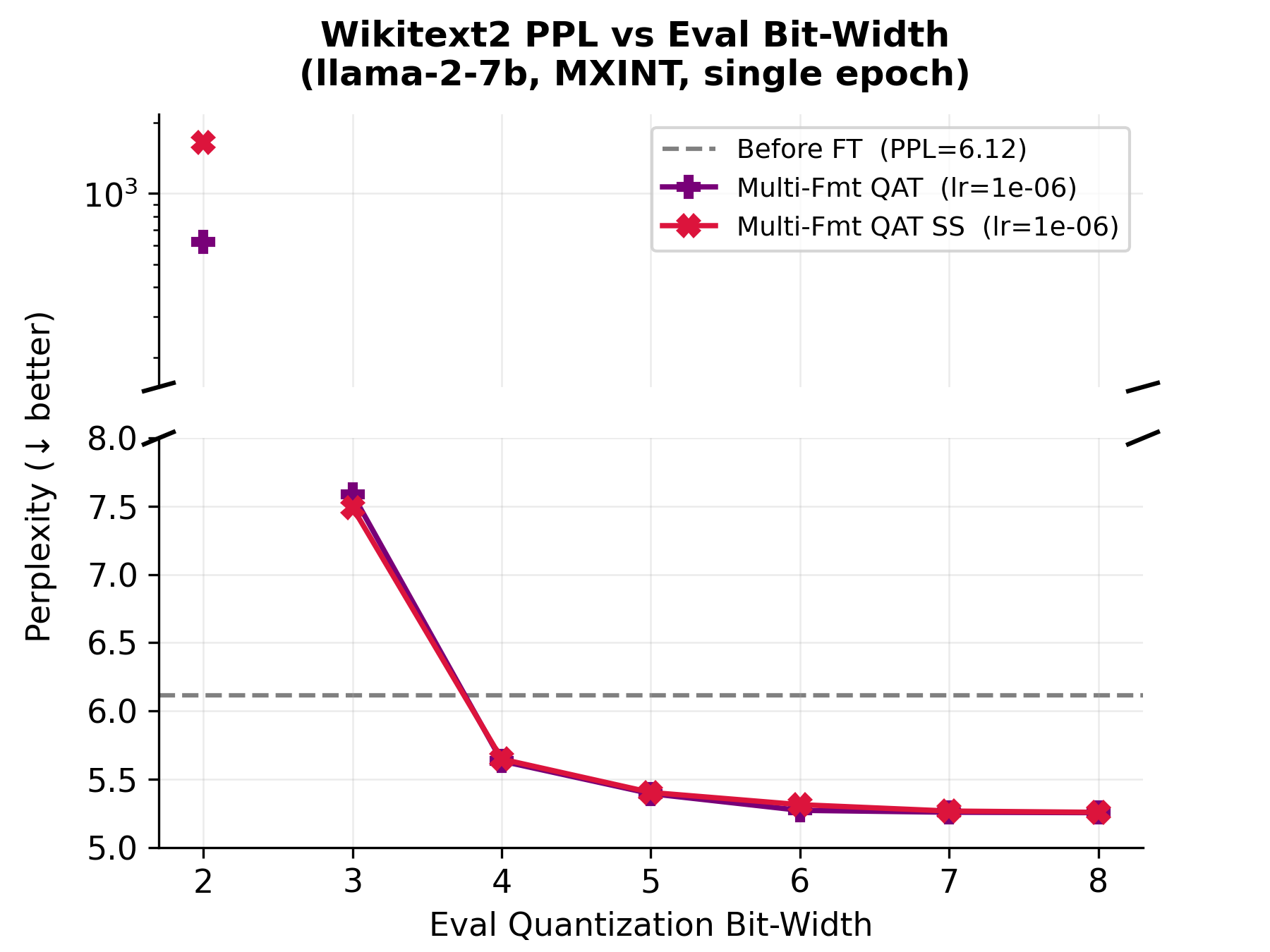}
  \hfill
  \includegraphics[width=0.49\textwidth]{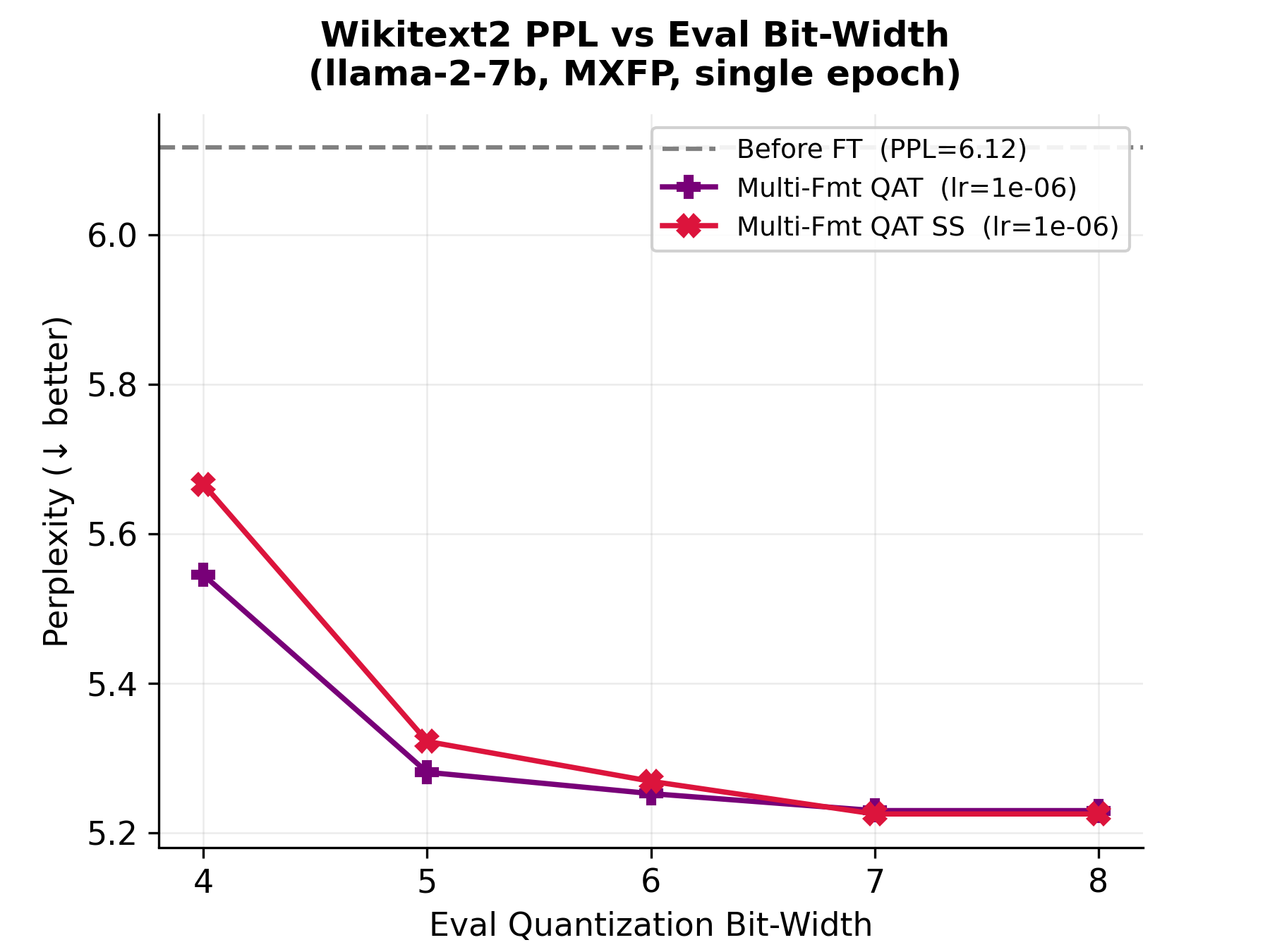}
  \caption{Multi-format QAT with Slice-and-Scale results for
  \texttt{llama-2-7b}.}
  \label{fig:app_mfss_llama2_7b}
\end{figure*}

\FloatBarrier
\section{Complete results of main result tables} 
\label{sec:appendix_complete_results}
Detailed results of Table~\ref{tab:main_results_mxint_downstream} in the main paper, including task-by-task performance for the MXINT formats, are shown in Table~\ref{tab:main_results_mxint_downstream_MMLU} (MMLU accuracy~\citep{hendrycks_measuring_2021}), Table~\ref{tab:main_results_mxint_downstream_MathQA} (MathQA accuracy~\citep{amini_mathqa_2019}), and Table~\ref{tab:main_results_mxint_downstream_HellaSwag} (HellaSwag accuracy~\citep{zellers_hellaswag_2019}). Similarly, per-task accuracy for the MXFP formats is indicated in Table~\ref{tab:main_results_mxfp_downstream_detailed}. Our results show that, across diverse downstream tasks, a single multi-format QAT model can effectively generalize to multiple deployment precisions while maintaining accuracy for each format, avoiding the need to train and store separate models for every target precision.

\begin{table*}[t!]
\centering
\resizebox{0.97\textwidth}{!}{
\begin{tabular}{c|c|ccccccc}
\hline
 \multirow{2}{*}{\textbf{Model}} & \multirow{2}{*}{\textbf{QAT/FT Precision}} & \multicolumn{7}{c}{\textbf{PTQ Precision}}  \\ \cline{3-9} 
  &  & \textbf{MXINT2} & \textbf{MXINT3$^*$} & \textbf{MXINT4} & \textbf{MXINT5$^*$} & \textbf{MXINT6} & \textbf{MXINT7$^*$} & \textbf{MXINT8} \\ \hline 
 & Full Precision FT & 25.0 & 28.1 & 39.3 & 41.1 & 42.3 & 42.5 & 42.7 \\ \cdashline{2-9} 
 & MXINT2 QAT        & 24.5 & \textbf{29.4} & 39.2 & 40.1 & 41.2 & 41.7 & 41.0 \\
 & MXINT4 QAT        & \textbf{25.3} & 28.6 & \textbf{39.7} & \textbf{41.8} & 42.7 & 42.6 & \textbf{42.9} \\
 & MXINT6 QAT        & 24.5 & 28.0 & 38.5 & 41.1 & 41.9 & 42.6 & 42.5 \\
 & MXINT8 QAT        & 24.3 & 27.1 & 39.1 & 40.7 & 42.1 & 42.4 & 42.7 \\
\multirow{-6}{*}{Llama-2-7B} & \cellcolor[HTML]{CCFACC}Multi-format QAT & \cellcolor[HTML]{CCFACC}23.8 & \cellcolor[HTML]{CCFACC}28.3 & \cellcolor[HTML]{CCFACC}39.0 & \cellcolor[HTML]{CCFACC}41.4 & \cellcolor[HTML]{CCFACC}\textbf{42.9} & \cellcolor[HTML]{CCFACC}\textbf{43.2} & \cellcolor[HTML]{CCFACC}\textbf{42.9} \\ \hline 
 & Full Precision FT & 23.8 & 23.3 & 28.8 & 37.4 & 39.2 & 39.6 & 39.6 \\ \cdashline{2-9} 
 & MXINT2 QAT & \textbf{25.7} & 23.9 & 28.2 & 37.4 & 39.3 & 39.5 & 39.8 \\
 & MXINT4 QAT & 24.0 & \textbf{24.6} & 28.4 & \textbf{38.6} & 39.2 & 39.0 & 39.4 \\
 & MXINT6 QAT & 23.5 & 23.8 & 28.2 & 37.8 & 39.3 & 39.0 & 39.7 \\
 & MXINT8 QAT & 22.2 & 23.7 & \textbf{29.1} & 37.3 & 39.7 & \textbf{39.8} & \textbf{40.4} \\
\multirow{-6}{*}{Llama-3.2-1B} & \cellcolor[HTML]{CCFACC}Multi-format QAT & \cellcolor[HTML]{CCFACC}24.4 & \cellcolor[HTML]{CCFACC}23.9 & \cellcolor[HTML]{CCFACC}27.8 & \cellcolor[HTML]{CCFACC}37.5 & \cellcolor[HTML]{CCFACC}\textbf{40.1} & \cellcolor[HTML]{CCFACC}39.1 & \cellcolor[HTML]{CCFACC}39.8 \\ \hline 
 & Full Precision FT & 26.5 & 31.1 & 50.2 & 55.8 & 56.1 & 55.7 & 55.7 \\ \cdashline{2-9} 
 & MXINT2 QAT & 25.3 & 30.3 & 50.5 & 55.9 & 56.0 & 55.9 & 56.0 \\
 & MXINT4 QAT & 26.7 & \textbf{31.7} & \textbf{51.4} & \textbf{56.0} & 56.4 & 55.7 & \textbf{56.1} \\
 & MXINT6 QAT & \textbf{26.8} & 29.5 & 50.6 & 55.2 & 56.0 & 55.6 & 55.8 \\
 & MXINT8 QAT & \textbf{26.8} & 31.0 & 50.3 & 55.6 & 56.0 & \textbf{56.0} & 55.8 \\
\multirow{-6}{*}{Llama-3.2-3B} & \cellcolor[HTML]{CCFACC}Multi-format QAT & \cellcolor[HTML]{CCFACC}23.8 & \cellcolor[HTML]{CCFACC}31.4 & \cellcolor[HTML]{CCFACC}50.6 & \cellcolor[HTML]{CCFACC}55.3 & \cellcolor[HTML]{CCFACC}\textbf{56.5} & \cellcolor[HTML]{CCFACC}55.7 & \cellcolor[HTML]{CCFACC}55.8 \\ \hline 
 & Full Precision FT & 25.4 & \textbf{25.5} & 37.7 & 48.8 & 50.6 & 51.4 & 52.0 \\ \cdashline{2-9} 
 & MXINT2 QAT & 23.9 & \textbf{25.5} & 38.1 & 48.2 & 51.4 & 51.3 & 52.0 \\
 & MXINT4 QAT & 25.2 & 24.1 & \textbf{42.0} & 49.2 & \textbf{52.3} & 51.4 & 52.0 \\
 & MXINT6 QAT & 25.1 & 24.7 & 37.6 & 49.5 & 50.5 & \textbf{51.8} & \textbf{52.1} \\
 & MXINT8 QAT & \textbf{25.5} & 25.0 & 37.4 & \textbf{49.7} & 50.7 & 51.7 & 52.0 \\
\multirow{-6}{*}{Qwen3-0.6B} & \cellcolor[HTML]{CCFACC}Multi-format QAT & \cellcolor[HTML]{CCFACC}24.0 & \cellcolor[HTML]{CCFACC}23.4 & \cellcolor[HTML]{CCFACC}40.5 & \cellcolor[HTML]{CCFACC}49.2 & \cellcolor[HTML]{CCFACC}50.7 & \cellcolor[HTML]{CCFACC}51.5 & \cellcolor[HTML]{CCFACC}51.9 \\ \hline  
 & Full Precision FT & 24.3 & 25.7 & 54.5 & 61.5 & 63.3 & 63.4 & 63.5 \\ \cdashline{2-9} 
 & MXINT2 QAT & 24.5 & 25.8 & 54.6 & 61.6 & 63.0 & 63.6 & 64.0 \\
 & MXINT4 QAT & \textbf{24.8} & \textbf{27.4} & 55.1 & 61.4 & \textbf{63.8} & 64.0 & \textbf{64.2} \\
 & MXINT6 QAT & 24.1 & 27.0 & \textbf{55.7} & 62.0 & 62.9 & \textbf{64.1} & 64.0 \\
 & MXINT8 QAT & 24.5 & 26.1 & 54.7 & 61.8 & 63.4 & 64.0 & 64.0 \\
\multirow{-6}{*}{Qwen3-1.7B} & \cellcolor[HTML]{CCFACC}Multi-format QAT & \cellcolor[HTML]{CCFACC}24.4 & \cellcolor[HTML]{CCFACC}26.4 & \cellcolor[HTML]{CCFACC}55.5 & \cellcolor[HTML]{CCFACC}\textbf{62.7} & \cellcolor[HTML]{CCFACC}63.2 & \cellcolor[HTML]{CCFACC}63.0 & \cellcolor[HTML]{CCFACC}63.0 \\ \hline   
 & Full Precision FT & 24.9 & 26.6 & 69.9 & 73.0 & 73.8 & 74.4 & 74.6 \\ \cdashline{2-9} 
 & MXINT2 QAT & 23.9 & 26.5 & \textbf{70.1} & 72.7 & 74.1 & 74.0 & 74.0 \\
 & MXINT4 QAT & 24.2 & \textbf{28.5} & \textbf{70.1} & 73.2 & 74.0 & 74.5 & 74.7 \\
 & MXINT6 QAT & 24.5 & 28.0 & 69.8 & \textbf{73.3} & 73.9 & \textbf{74.6} & \textbf{74.8} \\
 & MXINT8 QAT & 23.3 & 26.9 & 69.9 & 73.2 & 74.1 & 74.5 & \textbf{74.8} \\
\multirow{-6}{*}{Qwen3-4B} & \cellcolor[HTML]{CCFACC}Multi-format QAT & \cellcolor[HTML]{CCFACC}\textbf{25.4} & \cellcolor[HTML]{CCFACC}28.3 & \cellcolor[HTML]{CCFACC}69.8 & \cellcolor[HTML]{CCFACC}72.8 & \cellcolor[HTML]{CCFACC}\textbf{74.2} & \cellcolor[HTML]{CCFACC}\textbf{74.6} & \cellcolor[HTML]{CCFACC}74.3 \\ \bottomrule 

\end{tabular}
}
\caption{MMLU accuracy~\citep{hendrycks_measuring_2021} for various models when quantized to the MXINT formats. Each row corresponds to a different QAT/FT training precision, and each column shows the PTQ evaluation precision. Columns marked with $^*$ denote precisions not seen during multi-format/QAT training. QAT, PTQ, and FT denote Quantization Aware Training, Post Training Quantization, and Finetuning, respectively. 
}
\label{tab:main_results_mxint_downstream_MMLU}
\end{table*}

\begin{table*}[t!]
\centering
\resizebox{0.97\textwidth}{!}{
\begin{tabular}{c|c|ccccccc}
\hline
 \multirow{2}{*}{\textbf{Model}} & \multirow{2}{*}{\textbf{QAT/FT Precision}} & \multicolumn{7}{c}{\textbf{PTQ Precision}}  \\ \cline{3-9} 
  &  & \textbf{MXINT2} & \textbf{MXINT3$^*$} & \textbf{MXINT4} & \textbf{MXINT5$^*$} & \textbf{MXINT6} & \textbf{MXINT7$^*$} & \textbf{MXINT8} \\ \hline 
 & Full Precision FT & 18.4 & 25.0 & 27.8 & 28.0 & \textbf{28.8} & 28.4 & \textbf{28.4} \\ \cdashline{2-9} 
 & MXINT2 QAT        & \textbf{20.6} & 26.4 & \textbf{28.0} & \textbf{29.0} & 28.2 & 28.4 & 27.8 \\
 & MXINT4 QAT        & 17.8 & 26.4 & 27.6 & 27.6 & 27.8 & \textbf{28.6} & 28.2 \\
 & MXINT6 QAT        & 15.0 & \textbf{26.8} & 26.8 & 27.4 & 27.2 & 28.4 & 28.0 \\
 & MXINT8 QAT        & 16.6 & 25.6 & 27.4 & 28.0 & 27.4 & \textbf{28.6} & 28.0 \\
\multirow{-6}{*}{Llama-2-7B} & \cellcolor[HTML]{CCFACC}Multi-format QAT & \cellcolor[HTML]{CCFACC}18.4 & \cellcolor[HTML]{CCFACC}24.8 & \cellcolor[HTML]{CCFACC}26.8 & \cellcolor[HTML]{CCFACC}27.4 & \cellcolor[HTML]{CCFACC}27.6 & \cellcolor[HTML]{CCFACC}28.4 & \cellcolor[HTML]{CCFACC}\textbf{28.4} \\ \hline 
 & Full Precision FT & 17.8 & 17.8 & 24.4 & 28.0 & 28.2 & \textbf{28.0} & 27.6 \\ \cdashline{2-9} 
 & MXINT2 QAT & 18.2 & 18.0 & 24.0 & 26.8 & 28.2 & 26.2 & 26.6 \\
 & MXINT4 QAT & 18.0 & 19.6 & \textbf{25.2} & 28.0 & \textbf{28.6} & 27.6 & \textbf{28.4} \\
 & MXINT6 QAT & 17.6 & 17.6 & 23.8 & \textbf{28.8} & 28.4 & 27.6 & 27.8 \\
 & MXINT8 QAT & 18.6 & 18.2 & 24.0 & \textbf{28.8} & 28.2 & \textbf{28.0} & 28.0 \\
\multirow{-6}{*}{Llama-3.2-1B} & \cellcolor[HTML]{CCFACC}Multi-format QAT & \cellcolor[HTML]{CCFACC}\textbf{20.0} & \cellcolor[HTML]{CCFACC}\textbf{21.2} & \cellcolor[HTML]{CCFACC}\textbf{25.2} & \cellcolor[HTML]{CCFACC}28.4 & \cellcolor[HTML]{CCFACC}\textbf{28.6} & \cellcolor[HTML]{CCFACC}27.4 & \cellcolor[HTML]{CCFACC}\textbf{28.4} \\ \hline 
 & Full Precision FT & \textbf{23.2} & 25.4 & \textbf{35.4} & 33.4 & 36.4 & 35.0 & 34.8 \\ \cdashline{2-9} 
 & MXINT2 QAT & 21.0 & 27.4 & 33.8 & 33.4 & 36.2 & \textbf{35.2} & \textbf{35.0} \\
 & MXINT4 QAT & 19.6 & 27.4 & 33.4 & 33.0 & 36.0 & 34.6 & 33.8 \\
 & MXINT6 QAT & 18.2 & 26.4 & 34.2 & \textbf{34.0} & \textbf{36.8} & 34.2 & 34.2 \\
 & MXINT8 QAT & 19.4 & 25.6 & 34.8 & 33.4 & 36.2 & 34.4 & 34.4 \\
\multirow{-6}{*}{Llama-3.2-3B} & \cellcolor[HTML]{CCFACC}Multi-format QAT & \cellcolor[HTML]{CCFACC}20.6 & \cellcolor[HTML]{CCFACC}\textbf{28.0} & \cellcolor[HTML]{CCFACC}35.0 & \cellcolor[HTML]{CCFACC}33.4 & \cellcolor[HTML]{CCFACC}35.8 & \cellcolor[HTML]{CCFACC}34.8 & \cellcolor[HTML]{CCFACC}34.6 \\ \hline 
 & Full Precision FT & 18.8 & 19.6 & 28.6 & \textbf{33.4} & 36.0 & 36.0 & \textbf{35.8} \\ \cdashline{2-9} 
 & MXINT2 QAT & 19.8 & 18.6 & 27.6 & 29.6 & 34.8 & 35.0 & 35.0 \\
 & MXINT4 QAT & \textbf{20.6} & 18.2 & 28.6 & 33.2 & \textbf{38.2} & 34.8 & 35.6 \\
 & MXINT6 QAT & 19.6 & 19.0 & 28.6 & 31.8 & 37.2 & \textbf{37.0} & \textbf{35.8} \\
 & MXINT8 QAT & 17.0 & 20.2 & 28.6 & 33.2 & 37.4 & \textbf{37.0} & \textbf{35.8} \\
\multirow{-6}{*}{Qwen3-0.6B} & \cellcolor[HTML]{CCFACC}Multi-format QAT & \cellcolor[HTML]{CCFACC}17.6 & \cellcolor[HTML]{CCFACC}\textbf{20.6} & \cellcolor[HTML]{CCFACC}\textbf{28.8} & \cellcolor[HTML]{CCFACC}32.0 & \cellcolor[HTML]{CCFACC}37.0 & \cellcolor[HTML]{CCFACC}36.6 & \cellcolor[HTML]{CCFACC}35.6 \\ \hline  
 & Full Precision FT & \textbf{19.8} & \textbf{20.4} & 36.8 & 40.6 & 40.8 & \textbf{41.2} & 42.6 \\ \cdashline{2-9} 
 & MXINT2 QAT & 18.6 & 19.6 & 37.2 & \textbf{41.6} & \textbf{42.8} & 41.0 & \textbf{43.4} \\
 & MXINT4 QAT & 15.4 & 19.8 & 35.8 & 40.4 & 42.4 & 41.0 & 42.8 \\
 & MXINT6 QAT & 19.0 & 19.2 & 36.2 & 40.6 & 41.8 & 41.0 & 42.4 \\
 & MXINT8 QAT & 16.8 & 19.4 & 36.8 & 40.2 & 41.4 & 40.6 & 42.4 \\
\multirow{-6}{*}{Qwen3-1.7B} & \cellcolor[HTML]{CCFACC}Multi-format QAT & \cellcolor[HTML]{CCFACC}18.4 & \cellcolor[HTML]{CCFACC}\textbf{20.4} & \cellcolor[HTML]{CCFACC}\textbf{37.6} & \cellcolor[HTML]{CCFACC}40.8 & \cellcolor[HTML]{CCFACC}40.8 & \cellcolor[HTML]{CCFACC}40.8 & \cellcolor[HTML]{CCFACC}43.0 \\ \hline   
 & Full Precision FT & \textbf{21.0} & 22.6 & 47.2 & 47.4 & 46.4 & 48.4 & 48.2 \\ \cdashline{2-9} 
 & MXINT2 QAT & 19.6 & 24.2 & 47.8 & \textbf{49.6} & 47.8 & 50.4 & \textbf{51.0} \\
 & MXINT4 QAT & 19.2 & \textbf{24.4} & \textbf{49.6} & 48.0 & 47.8 & 49.4 & 49.6 \\
 & MXINT6 QAT & 16.2 & 23.8 & 47.6 & 48.0 & \textbf{48.4} & \textbf{50.6} & \textbf{51.0} \\
 & MXINT8 QAT & 20.8 & 23.2 & 47.8 & 47.8 & 48.0 & 50.0 & \textbf{51.0} \\
\multirow{-6}{*}{Qwen3-4B} & \cellcolor[HTML]{CCFACC}Multi-format QAT & \cellcolor[HTML]{CCFACC}19.4 & \cellcolor[HTML]{CCFACC}22.8 & \cellcolor[HTML]{CCFACC}47.2 & \cellcolor[HTML]{CCFACC}47.8 & \cellcolor[HTML]{CCFACC}46.8 & \cellcolor[HTML]{CCFACC}48.6 & \cellcolor[HTML]{CCFACC}50.4 \\ \bottomrule 
\end{tabular}
}
\caption{MathQA normalized accuracy~\citep{amini_mathqa_2019} for various models when quantized to the MXINT formats. Each row corresponds to a different QAT/FT training precision, and each column shows the PTQ evaluation precision. Columns marked with $^*$ denote precisions not seen during multi-format/QAT training. QAT, PTQ, and FT denote Quantization Aware Training, Post Training Quantization, and Finetuning, respectively. 
}
\label{tab:main_results_mxint_downstream_MathQA}
\end{table*}

\begin{table*}[t!]
\centering
\resizebox{0.97\textwidth}{!}{
\begin{tabular}{c|c|ccccccc}
\hline
 \multirow{2}{*}{\textbf{Model}} & \multirow{2}{*}{\textbf{QAT/FT Precision}} & \multicolumn{7}{c}{\textbf{PTQ Precision}}  \\ \cline{3-9} 
  &  & \textbf{MXINT2} & \textbf{MXINT3$^*$} & \textbf{MXINT4} & \textbf{MXINT5$^*$} & \textbf{MXINT6} & \textbf{MXINT7$^*$} & \textbf{MXINT8} \\ \hline 
 & Full Precision FT & 24.8 & 59.4 & 66.2 & 65.0 & 66.2 & 67.2 & 66.8 \\ \cdashline{2-9} 
 & MXINT2 QAT        & \textbf{42.8} & \textbf{63.0} & 66.0 & 64.8 & 65.8 & 66.6 & 66.6 \\
 & MXINT4 QAT        & 25.6 & 61.2 & \textbf{66.4} & 65.0 & 66.2 & 66.6 & 67.0 \\
 & MXINT6 QAT        & 26.8 & 60.4 & 65.8 & 65.2 & \textbf{66.4} & \textbf{67.4} & \textbf{67.4} \\
 & MXINT8 QAT        & 24.8 & 60.4 & 66.2 & 65.6 & 66.0 & 67.0 & 67.2 \\
\multirow{-6}{*}{Llama-2-7B} & \cellcolor[HTML]{CCFACC}Multi-format QAT & \cellcolor[HTML]{CCFACC}27.0 & \cellcolor[HTML]{CCFACC}60.0 & \cellcolor[HTML]{CCFACC}65.8 & \cellcolor[HTML]{CCFACC}\textbf{65.8} & \cellcolor[HTML]{CCFACC}66.0 & \cellcolor[HTML]{CCFACC}67.0 & \cellcolor[HTML]{CCFACC}67.0 \\ \hline 
 & Full Precision FT & 24.0 & 35.6 & 51.6 & 57.2 & 56.4 & \textbf{56.4} & 56.4 \\ \cdashline{2-9} 
 & MXINT2 QAT & \textbf{26.8} & 36.8 & 52.0 & 56.4 & 56.2 & 56.2 & 56.0 \\
 & MXINT4 QAT & 26.4 & \textbf{40.8} & \textbf{54.6} & 56.6 & 56.4 & 56.0 & 56.6 \\
 & MXINT6 QAT & 26.2 & 36.0 & 53.4 & 57.4 & \textbf{56.6} & 55.6 & \textbf{57.0} \\
 & MXINT8 QAT & 25.4 & 35.0 & 52.6 & \textbf{58.0} & 56.2 & 56.0 & 56.4 \\
\multirow{-6}{*}{Llama-3.2-1B} & \cellcolor[HTML]{CCFACC}Multi-format QAT & \cellcolor[HTML]{CCFACC}25.4 & \cellcolor[HTML]{CCFACC}39.6 & \cellcolor[HTML]{CCFACC}52.6 & \cellcolor[HTML]{CCFACC}56.4 & \cellcolor[HTML]{CCFACC}55.2 & \cellcolor[HTML]{CCFACC}55.8 & \cellcolor[HTML]{CCFACC}56.0 \\ \hline 
 & Full Precision FT & 25.8 & 50.2 & 63.0 & 64.2 & 63.8 & 64.4 & 64.8 \\ \cdashline{2-9} 
 & MXINT2 QAT & \textbf{30.2} & \textbf{52.0} & 63.4 & 64.8 & 64.8 & \textbf{65.8} & 65.0 \\
 & MXINT4 QAT & 27.2 & 50.8 & 63.0 & \textbf{65.2} & 65.0 & 64.8 & 65.0 \\
 & MXINT6 QAT & 27.4 & 50.2 & 63.4 & \textbf{65.2} & 65.0 & 65.6 & 65.6 \\
 & MXINT8 QAT & 28.4 & 50.0 & 63.0 & \textbf{65.2} & \textbf{65.6} & 65.2 & 65.2 \\
\multirow{-6}{*}{Llama-3.2-3B} & \cellcolor[HTML]{CCFACC}Multi-format QAT & \cellcolor[HTML]{CCFACC}27.2 & \cellcolor[HTML]{CCFACC}50.8 & \cellcolor[HTML]{CCFACC}\textbf{63.6} & \cellcolor[HTML]{CCFACC}64.8 & \cellcolor[HTML]{CCFACC}65.0 & \cellcolor[HTML]{CCFACC}65.4 & \cellcolor[HTML]{CCFACC}\textbf{65.8} \\ \hline 
 & Full Precision FT & 27.8 & 30.0 & 49.0 & 48.4 & 51.0 & 52.6 & \textbf{53.0} \\ \cdashline{2-9}
 & MXINT2 QAT & \textbf{29.4} & 30.6 & 48.6 & 49.2 & 50.6 & 51.0 & 50.6 \\
 & MXINT4 QAT & 24.0 & \textbf{33.6} & 48.2 & \textbf{50.0} & 51.0 & 52.0 & 51.8 \\
 & MXINT6 QAT & 26.2 & 30.4 & 48.6 & \textbf{50.0} & \textbf{51.6} & \textbf{52.8} & 52.6 \\
 & MXINT8 QAT & 25.4 & 31.0 & 48.0 & 48.2 & 51.4 & 52.4 & 52.6 \\
\multirow{-6}{*}{Qwen3-0.6B} & \cellcolor[HTML]{CCFACC}Multi-format QAT & \cellcolor[HTML]{CCFACC}24.4 & \cellcolor[HTML]{CCFACC}31.6 & \cellcolor[HTML]{CCFACC}\textbf{49.6} & \cellcolor[HTML]{CCFACC}49.8 & \cellcolor[HTML]{CCFACC}50.6 & \cellcolor[HTML]{CCFACC}52.6 & \cellcolor[HTML]{CCFACC}51.2 \\ \hline  
 & Full Precision FT & 28.2 & \textbf{41.4} & 54.6 & 57.2 & 58.4 & 59.2 & 59.4 \\ \cdashline{2-9} 
 & MXINT2 QAT & 25.0 & 41.0 & 54.4 & 57.0 & 57.2 & 57.6 & 59.0 \\
 & MXINT4 QAT & 27.8 & \textbf{41.4} & 54.8 & 56.8 & 59.0 & 59.0 & 59.2 \\
 & MXINT6 QAT & 24.4 & 41.2 & 54.8 & \textbf{58.4} & 58.8 & 58.6 & 58.8 \\
 & MXINT8 QAT & \textbf{30.4} & 40.4 & 54.8 & 57.6 & 59.2 & 58.4 & 58.8 \\
\multirow{-6}{*}{Qwen3-1.7B} & \cellcolor[HTML]{CCFACC}Multi-format QAT & \cellcolor[HTML]{CCFACC}25.4 & \cellcolor[HTML]{CCFACC}41.2 & \cellcolor[HTML]{CCFACC}\textbf{55.0} & \cellcolor[HTML]{CCFACC}57.6 & \cellcolor[HTML]{CCFACC}\textbf{60.2} & \cellcolor[HTML]{CCFACC}\textbf{60.2} & \cellcolor[HTML]{CCFACC}\textbf{60.4} \\ \hline   
 & Full Precision FT & \textbf{27.8} & 44.2 & \textbf{63.2} & \textbf{61.2} & 62.6 & 62.4 & 62.2 \\ \cdashline{2-9} 
 & MXINT2 QAT & 24.6 & 44.0 & 60.2 & 58.4 & 60.8 & 61.2 & 61.0 \\
 & MXINT4 QAT & 26.4 & \textbf{45.6} & 62.6 & 60.2 & 63.0 & 62.6 & 61.6 \\
 & MXINT6 QAT & 26.0 & 43.8 & 62.8 & 60.4 & 63.0 & \textbf{62.8} & \textbf{62.8} \\
 & MXINT8 QAT & 24.8 & 44.2 & 62.6 & 60.6 & 62.8 & 62.6 & 62.6 \\
\multirow{-6}{*}{Qwen3-4B} & \cellcolor[HTML]{CCFACC}Multi-format QAT & \cellcolor[HTML]{CCFACC}26.8 & \cellcolor[HTML]{CCFACC}44.4 & \cellcolor[HTML]{CCFACC}63.0 & \cellcolor[HTML]{CCFACC}60.6 & \cellcolor[HTML]{CCFACC}\textbf{63.4} & \cellcolor[HTML]{CCFACC}62.6 & \cellcolor[HTML]{CCFACC}62.6 \\ \bottomrule 
\end{tabular}
}

\caption{HellaSwag normalized accuracy~\citep{zellers_hellaswag_2019} for various models when quantized to the MXINT formats. Each row corresponds to a different QAT/FT training precision, and each column shows the PTQ evaluation precision. Columns marked with $^*$ denote precisions not seen during multi-format/QAT training. QAT, PTQ, and FT denote Quantization Aware Training, Post Training Quantization, and Finetuning, respectively. 
}
\label{tab:main_results_mxint_downstream_HellaSwag}

\end{table*}

\begin{table*}[t!]
\centering
\resizebox{0.999\textwidth}{!}{
\begin{tabular}{c|ccc|ccc|ccc|ccc|ccc}
\hline
  \multirow{3}{*}{\textbf{QAT/FT Precision}} & \multicolumn{15}{c}{\textbf{PTQ Precision}}  \\ \cline{2-16} 
  &  \multicolumn{3}{c|}{MXFP4}  & \multicolumn{3}{c|}{MXFP5$^*$} & \multicolumn{3}{c|}{MXFP6} & \multicolumn{3}{c|}{MXFP7$^*$} & \multicolumn{3}{c}{MXFP8}\\ \cline{2-16}  
  &  \textbf{MMLU} & \textbf{Math} & \textbf{HellaS} & \textbf{MMLU} & \textbf{Math} & \textbf{HellaS} & \textbf{MMLU} & \textbf{Math} & \textbf{HellaS} & \textbf{MMLU} & \textbf{Math} & \textbf{HellaS} & \textbf{MMLU} & \textbf{Math} & \textbf{HellaS} \\ \hline \hline 
  \multicolumn{16}{c}{Llama-2-7B} \\ \hline 
 Full Prec. FT & 37.4 & 28.8 & 63.6 & 42.0 & \textbf{28.6} & 65.8 & 42.4 & \textbf{28.4} & 66.4 & 42.3 & 28.4 & \textbf{67.4} & \textbf{42.3} & \textbf{28.8} & \textbf{67.4} \\ \cdashline{2-16} 
 MXFP4 QAT        & \textbf{37.8} & \textbf{29.6} & 63.8 & \textbf{42.5} & 28.4 & \textbf{67.2} & 42.5 & 28.2 & \textbf{67.6} & \textbf{42.6} & 28.2 & 67.0 & 42.1 & 27.6 & 67.0 \\
 MXFP6 QAT        & 36.7 & 29.0 & 64.2 & 41.9 & \textbf{28.6} & 66.0 & 42.5 & \textbf{28.4} & 66.6 & 41.8 & 28.4 & 67.0 & 41.9 & 28.4 & 67.2 \\
 MXFP8 QAT        & 36.7 & \textbf{29.6} & \textbf{64.4} & 42.1 & 27.8 & 66.2 & 42.4 & 28.2 & 66.4 & 42.4 & \textbf{28.8} & 66.8 & 42.1 & 28.4 & 66.8 \\
\cellcolor[HTML]{CCFACC}Multi-format QAT & \cellcolor[HTML]{CCFACC}\textbf{37.8} & \cellcolor[HTML]{CCFACC}28.8 & \cellcolor[HTML]{CCFACC}64.0 & \cellcolor[HTML]{CCFACC}41.7 & \cellcolor[HTML]{CCFACC}28.4 & \cellcolor[HTML]{CCFACC}66.4 & \cellcolor[HTML]{CCFACC}\textbf{42.6} & \cellcolor[HTML]{CCFACC}\textbf{28.4} & \cellcolor[HTML]{CCFACC}66.4 & \cellcolor[HTML]{CCFACC}42.3 & \cellcolor[HTML]{CCFACC}28.4 & \cellcolor[HTML]{CCFACC}66.8 & \cellcolor[HTML]{CCFACC}42.1 & \cellcolor[HTML]{CCFACC}\textbf{28.8} & \cellcolor[HTML]{CCFACC}66.6 \\ \hline \hline
\multicolumn{16}{c}{Llama-3.2-1B} \\ \hline
Full Prec. FT & 30.6 & \textbf{26.6} & 55.0 & 39.2 & 25.8 & \textbf{58.4} & \textbf{40.2} & 26.0 & \textbf{58.4} & 39.4 & 27.6 & \textbf{57.6} & 39.7 & 27.6 & \textbf{57.8} \\ \cdashline{2-16} 
 MXFP4 QAT        & \textbf{31.9} & \textbf{26.6} & 56.2 & 40.2 & \textbf{27.0} & 57.0 & 38.9 & \textbf{26.8} & 57.8 & 40.1 & \textbf{28.8} & 57.0 & 39.9 & \textbf{29.0} & 56.4 \\
 MXFP6 QAT        & 31.1 & 24.8 & 56.0 & 40.2 & 25.0 & 58.2 & 39.5 & 26.4 & 57.4 & 40.2 & 28.0 & 57.2 & 39.8 & 27.4 & 57.0 \\
 MXFP8 QAT        & 31.0 & 25.6 & \textbf{56.8} & 40.4 & 24.8 & 57.8 & 39.4 & 26.0 & 57.8 & \textbf{40.3} & 28.2 & 56.8 & \textbf{40.1} & 28.0 & 57.2 \\
\cellcolor[HTML]{CCFACC}Multi-format QAT & \cellcolor[HTML]{CCFACC}31.3 & \cellcolor[HTML]{CCFACC}25.4 & \cellcolor[HTML]{CCFACC}56.2 & \cellcolor[HTML]{CCFACC}\textbf{40.8} & \cellcolor[HTML]{CCFACC}26.0 & \cellcolor[HTML]{CCFACC}57.8 & \cellcolor[HTML]{CCFACC}39.7 & \cellcolor[HTML]{CCFACC}26.4 & \cellcolor[HTML]{CCFACC}57.6 & \cellcolor[HTML]{CCFACC}39.6 & \cellcolor[HTML]{CCFACC}28.0 & \cellcolor[HTML]{CCFACC}56.6 & \cellcolor[HTML]{CCFACC}39.8 & \cellcolor[HTML]{CCFACC}28.4 & \cellcolor[HTML]{CCFACC}56.4 \\ \hline \hline
\multicolumn{16}{c}{Llama-3.2-3B} \\ \hline
 Full Prec. FT & 53.7 & \textbf{33.2} & \textbf{65.0} & 56.2 & \textbf{36.4} & 64.8 & 56.1 & 35.2 & 63.6 & 55.8 & \textbf{35.6} & 64.0 & 55.7 & \textbf{35.0} & 64.0 \\ \cdashline{2-16}
 MXFP4 QAT & \textbf{54.2} & 32.4 & 64.0 & 56.2 & 35.2 & 64.4 & 55.9 & 35.2 & 64.4 & \textbf{55.9} & 35.2 & 64.6 & 55.9 & 34.4 & 65.0 \\
 MXFP6 QAT & 53.7 & 32.0 & 64.2 & \textbf{56.5} & 35.2 & 65.2 & 55.7 & 35.6 & 64.4 & \textbf{55.9} & 34.6 & \textbf{65.6} & 56.0 & 33.8 & \textbf{65.2} \\
 MXFP8 QAT & 54.0 & 32.0 & 64.6 & 56.3 & \textbf{36.4} & 64.8 & 55.9 & \textbf{36.2} & \textbf{64.8} & 55.7 & 34.8 & 65.0 & \textbf{56.3} & 33.8 & 65.0 \\
 \cellcolor[HTML]{CCFACC}Multi-format QAT & \cellcolor[HTML]{CCFACC}54.0 & \cellcolor[HTML]{CCFACC}31.6 & \cellcolor[HTML]{CCFACC}64.2 & \cellcolor[HTML]{CCFACC}56.3 & \cellcolor[HTML]{CCFACC}35.4 & \cellcolor[HTML]{CCFACC}\textbf{65.8} & \cellcolor[HTML]{CCFACC}\textbf{56.3} & \cellcolor[HTML]{CCFACC}35.6 & \cellcolor[HTML]{CCFACC}\textbf{64.8} & \cellcolor[HTML]{CCFACC}55.7 & \cellcolor[HTML]{CCFACC}34.4 & \cellcolor[HTML]{CCFACC}65.2 & \cellcolor[HTML]{CCFACC}55.9 & \cellcolor[HTML]{CCFACC}34.0 & \cellcolor[HTML]{CCFACC}\textbf{65.2} \\ \hline \hline
 \multicolumn{16}{c}{Qwen3-0.6B} \\ \hline
 Full Prec. FT & \textbf{48.1} & \textbf{33.2} & 50.6 & 49.8 & 37.6 & 50.8 & 50.6 & 36.2 & 52.0 & 51.1 & 34.0 & \textbf{52.0} & 50.8 & 34.2 & 51.6 \\ \cdashline{2-16} 
 MXFP4 QAT & 47.6 & 32.6 & 51.0 & 50.1 & 36.6 & 51.0 & 51.2 & \textbf{36.6} & 52.2 & 51.4 & \textbf{35.0} & 51.8 & 51.2 & \textbf{35.2} & \textbf{52.2} \\
 MXFP6 QAT & 48.0 & 31.6 & \textbf{51.4} & \textbf{50.5} & 37.4 & \textbf{51.6} & 51.1 & 36.4 & \textbf{52.6} & \textbf{51.5} & 34.4 & 51.0 & 51.0 & 34.0 & 51.2 \\
 MXFP8 QAT & 47.9 & 32.2 & 50.8 & \textbf{50.5} & \textbf{38.0} & 51.0 & \textbf{51.3} & \textbf{36.6} & 52.4 & 51.3 & 33.6 & 51.8 & 51.1 & 33.0 & 51.4 \\
\cellcolor[HTML]{CCFACC}Multi-format QAT & \cellcolor[HTML]{CCFACC}47.6 & \cellcolor[HTML]{CCFACC}32.2 & \cellcolor[HTML]{CCFACC}50.6 & \cellcolor[HTML]{CCFACC}50.2 & \cellcolor[HTML]{CCFACC}36.8 & \cellcolor[HTML]{CCFACC}51.2 & \cellcolor[HTML]{CCFACC}\textbf{51.3} & \cellcolor[HTML]{CCFACC}36.2 & \cellcolor[HTML]{CCFACC}51.8 & \cellcolor[HTML]{CCFACC}51.4 & \cellcolor[HTML]{CCFACC}34.8 & \cellcolor[HTML]{CCFACC}50.6 & \cellcolor[HTML]{CCFACC}\textbf{51.3} & \cellcolor[HTML]{CCFACC}34.0 & \cellcolor[HTML]{CCFACC}50.8 \\ \hline \hline
\multicolumn{16}{c}{Qwen3-1.7B} \\ \hline
 Full Prec. FT & \textbf{60.3} & 38.8 & 56.4 & 61.8 & 40.4 & \textbf{59.0} & 62.5 & 39.2 & \textbf{59.4} & 62.5 & 40.6 & \textbf{59.8} & 62.6 & 41.4 & \textbf{60.0} \\ \cdashline{2-16} 
 MXFP4 QAT & 59.9 & \textbf{41.0} & 57.4 & 63.2 & 40.8 & 57.2 & 63.4 & 39.8 & 58.6 & 64.1 & \textbf{42.4} & 58.0 & \textbf{63.9} & \textbf{42.8} & 58.8 \\
 MXFP6 QAT & 59.8 & 38.0 & 57.2 & 63.4 & 40.4 & 58.2 & 63.6 & \textbf{40.2} & 59.0 & 64.0 & 40.6 & 58.2 & 63.7 & 41.6 & 58.8 \\
 MXFP8 QAT & 59.9 & 39.4 & \textbf{57.6} & 63.1 & \textbf{41.0} & 58.8 & 63.3 & 40.0 & 58.8 & 63.7 & 41.2 & 58.2 & 63.7 & 41.2 & 58.4 \\
\cellcolor[HTML]{CCFACC}Multi-format QAT & \cellcolor[HTML]{CCFACC}59.9 & \cellcolor[HTML]{CCFACC}40.0 & \cellcolor[HTML]{CCFACC}56.6 & \cellcolor[HTML]{CCFACC}\textbf{63.7} & \cellcolor[HTML]{CCFACC}40.2 & \cellcolor[HTML]{CCFACC}58.4 & \cellcolor[HTML]{CCFACC}\textbf{63.8} & \cellcolor[HTML]{CCFACC}\textbf{40.2} & \cellcolor[HTML]{CCFACC}58.8 & \cellcolor[HTML]{CCFACC}\textbf{64.2} & \cellcolor[HTML]{CCFACC}41.2 & \cellcolor[HTML]{CCFACC}59.0 & \cellcolor[HTML]{CCFACC}63.7 & \cellcolor[HTML]{CCFACC}41.6 & \cellcolor[HTML]{CCFACC}58.8 \\ \hline \hline
\multicolumn{16}{c}{Qwen3-4B} \\ \hline
 Full Prec. FT & 70.8 & 47.8 & \textbf{61.6} & 72.8 & 47.4 & 63.4 & 73.2 & 47.6 & 62.4 & 74.3 & 48.0 & 62.0 & 74.3 & 47.8 & 62.2 \\ \cdashline{2-16} 
 MXFP4 QAT & \textbf{72.2} & 47.4 & 61.2 & \textbf{73.8} & \textbf{48.8} & \textbf{64.0} & \textbf{74.1} & 48.6 & \textbf{63.0} & 74.4 & 49.6 & \textbf{63.0} & \textbf{74.5} & 49.8 & \textbf{63.0} \\
 MXFP6 QAT & 71.4 & \textbf{48.4} & 61.0 & 73.2 & 48.2 & 63.4 & 73.2 & 48.4 & 62.8 & \textbf{74.5} & 49.8 & 62.8 & \textbf{74.5} & 48.6 & 62.8 \\
 MXFP8 QAT & 71.3 & 47.8 & \textbf{61.6} & 73.3 & 48.6 & 63.8 & 73.7 & 49.6 & \textbf{63.0} & 74.0 & 49.8 & \textbf{63.0} & 74.3 & 49.4 & 62.8 \\
\cellcolor[HTML]{CCFACC}Multi-format QAT & \cellcolor[HTML]{CCFACC}72.0 & \cellcolor[HTML]{CCFACC}47.6 & \cellcolor[HTML]{CCFACC}61.4 & \cellcolor[HTML]{CCFACC}73.7 & \cellcolor[HTML]{CCFACC}\textbf{48.8} & \cellcolor[HTML]{CCFACC}63.2 & \cellcolor[HTML]{CCFACC}74.0 & \cellcolor[HTML]{CCFACC}\textbf{49.8} & \cellcolor[HTML]{CCFACC}62.8 & \cellcolor[HTML]{CCFACC}74.4 & \cellcolor[HTML]{CCFACC}\textbf{50.2} & \cellcolor[HTML]{CCFACC}62.6 & \cellcolor[HTML]{CCFACC}74.4 & \cellcolor[HTML]{CCFACC}\textbf{50.0} & \cellcolor[HTML]{CCFACC}62.4 \\ \bottomrule 

\end{tabular}
}

\caption{Comparison of MMLU~\citep{hendrycks_measuring_2021}, MathQA~\citep{amini_mathqa_2019}, and HellaSwag~\citep{zellers_hellaswag_2019} accuracies for various models when quantized to the MXFP formats. Each row corresponds to a different QAT/FT training precision, and each column shows the PTQ evaluation precision. Columns marked with $^*$ denote precisions not seen during multi-format/QAT training. QAT, PTQ, and FT denote Quantization Aware Training, Post Training Quantization, and Finetuning, respectively.
}
\label{tab:main_results_mxfp_downstream_detailed}
\end{table*}

\FloatBarrier
\section{Tensor-level reconstruction error for Slice-and-Scale}
\label{app:ss_tensor_level}

\begin{figure*}[th]
  \centering
  \includegraphics[width=0.49\textwidth]{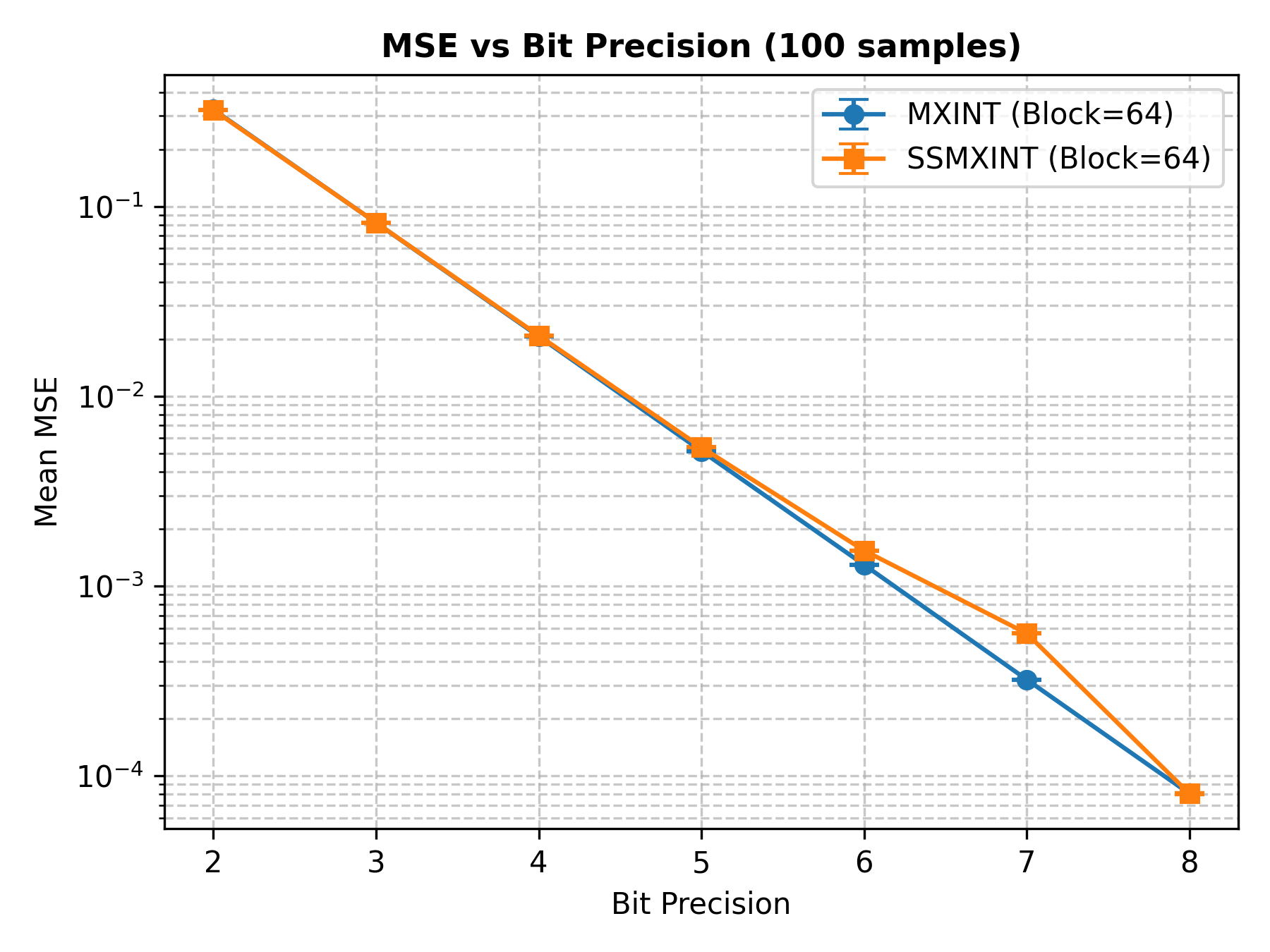}
  \hfill
  \includegraphics[width=0.49\textwidth]{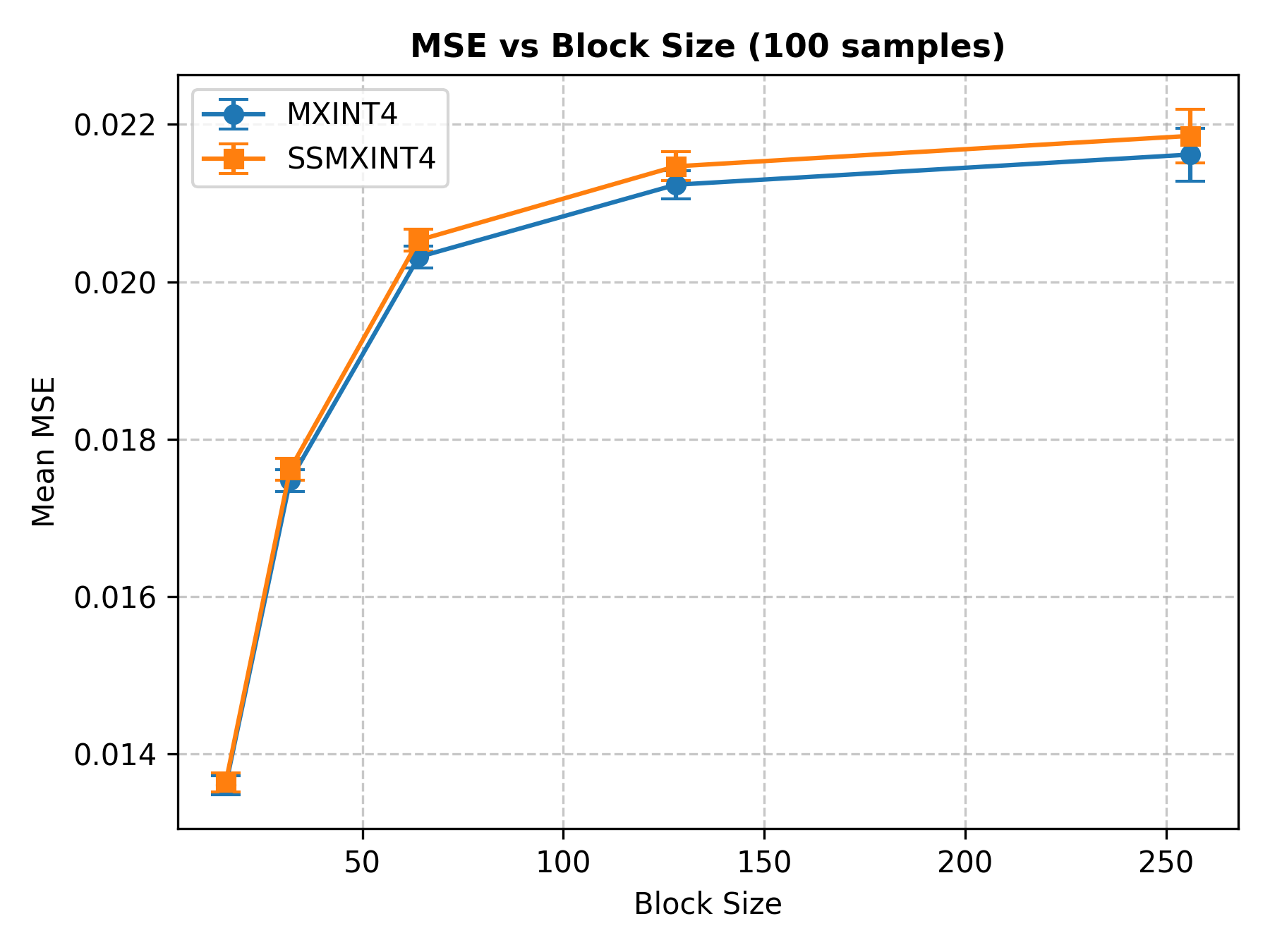}
 \caption{Tensor reconstruction MSE for direct MXINT quantization and Slice-and-Scale conversion (SSMXINT) on 100 random tensors. \textbf{Left:} Varying bit precision at block size 64. \textbf{Right:} Varying block size at 4-bit precision.}
  \label{fig:ssmxint_study_mse}
\end{figure*}

\begin{figure*}[th]
  \centering
  \includegraphics[width=0.49\textwidth]{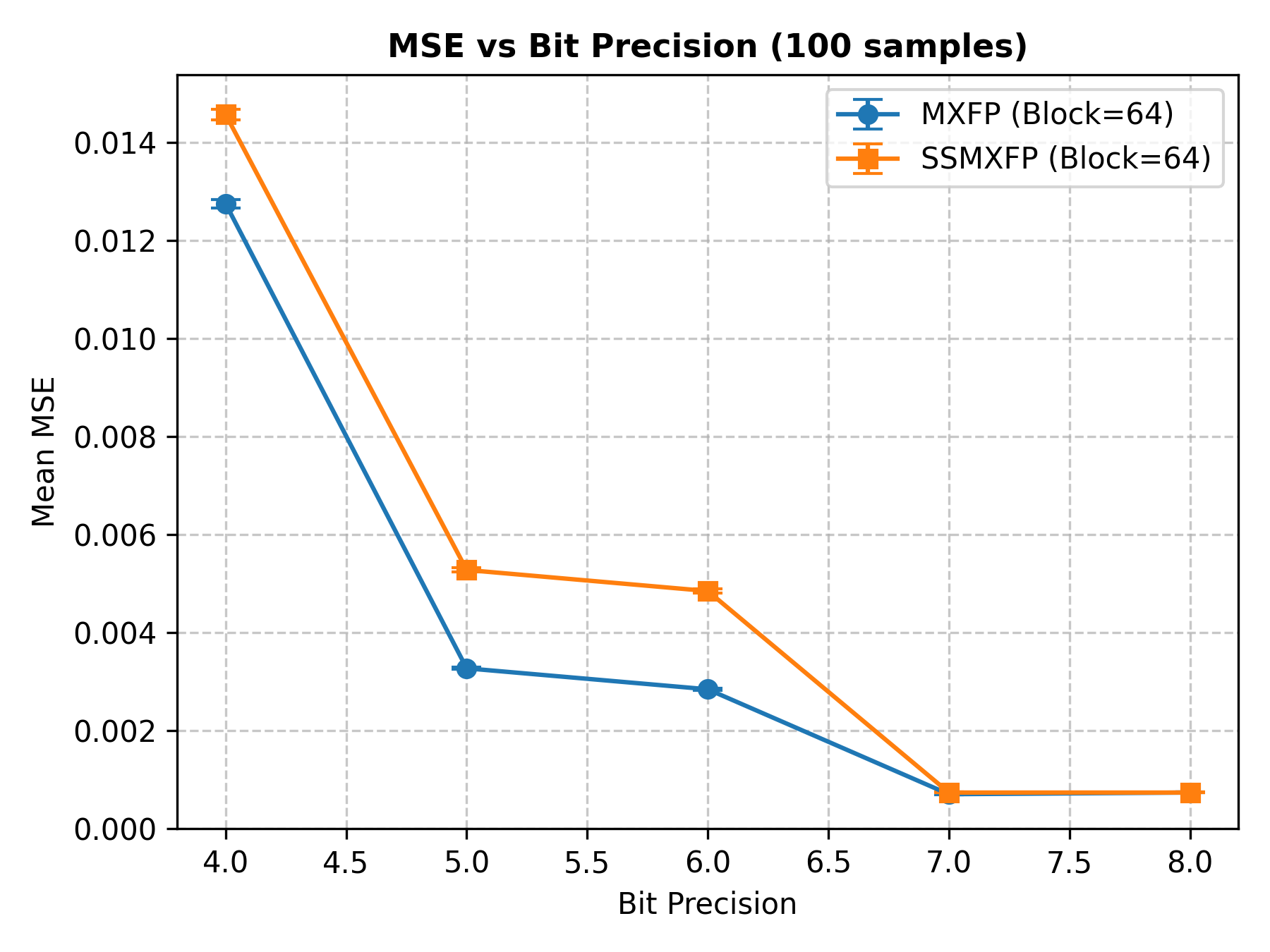}
  \hfill
  \includegraphics[width=0.49\textwidth]{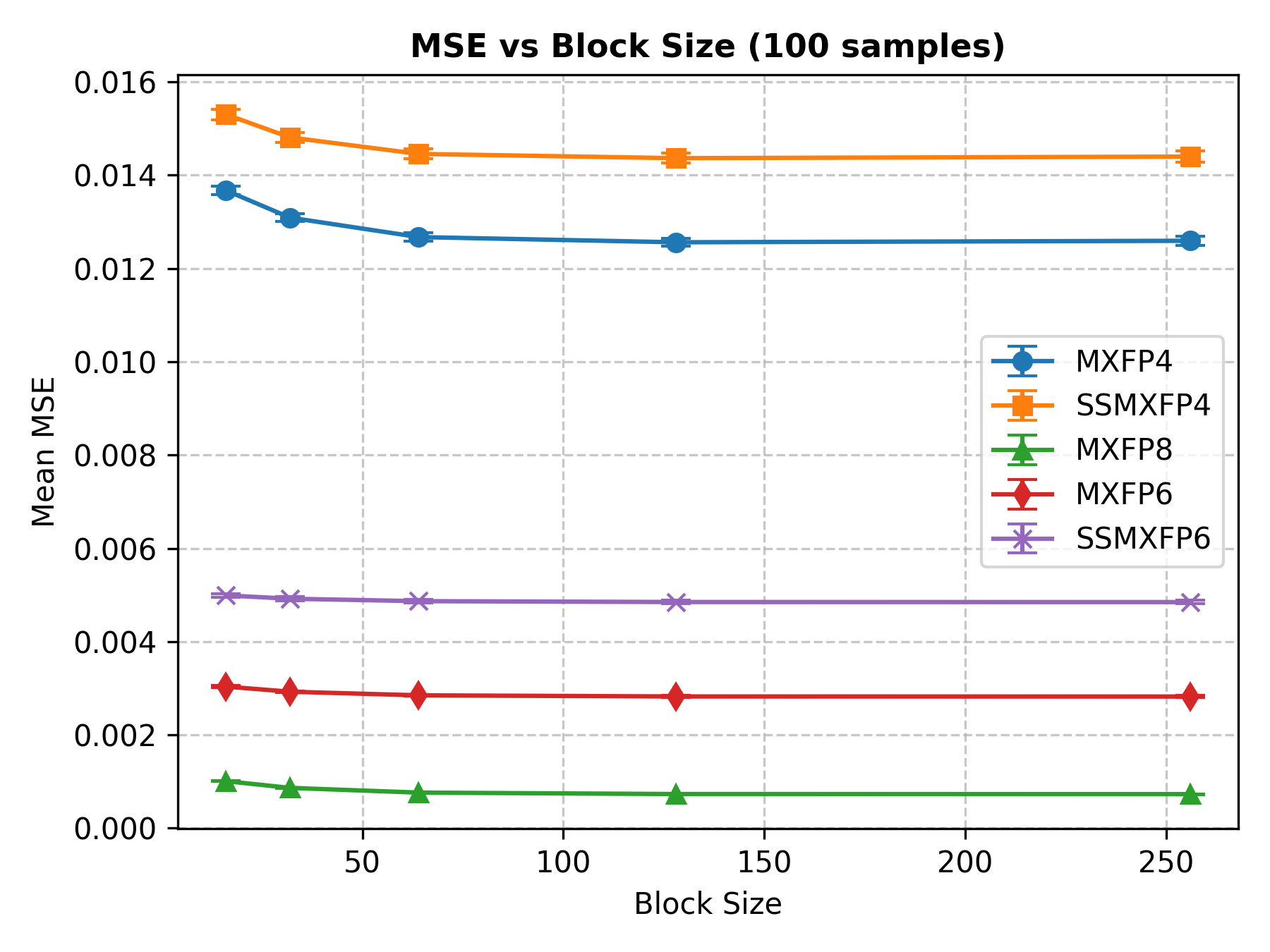}
  \caption{Tensor reconstruction MSE for direct MXFP quantization and Slice-and-Scale conversion (SSMXFP) on 100 random tensors. \textbf{Left:} Varying bit precision at block size 64. \textbf{Right:} Varying block size at 4-bit precision.}
  \label{fig:ssmxfp_study_mse}
\end{figure*}

Figures~\ref{fig:ssmxint_study_mse} and~\ref{fig:ssmxfp_study_mse} report average layer-wise MSE on 100 random tensors of shape $(1,1024)$ under two sweeps: varying bit precision at fixed block size 64, and varying block size at fixed 4-bit precision. As expected, reconstruction error decreases with higher precision and smaller block size. Across all settings, SS closely matches direct quantization for both MXINT and MXFP. Although SSMXFP exhibits a modestly larger relative gap at intermediate bit-widths, the absolute difference remains small, confirming that SS preserves the numerical behavior of the 8-bit anchor with high fidelity.

\end{document}